%% file: main.tex
\documentclass[10pt,twocolumn,letterpaper]{article}
\usepackage[pagenumbers]{cvpr}

\usepackage{iftex}
\ifXeTeX
  \usepackage{fontspec}
\fi

\usepackage{array,tabularx,multirow,pifont,colortbl}

\definecolor{TableInk}{HTML}{253D47}
\definecolor{TableRule}{HTML}{345B73}
\definecolor{TableSeparator}{HTML}{B6C9CF}
\definecolor{TableHeader}{HTML}{EEF4F6}
\definecolor{TableOverall}{HTML}{EEF4F5}
\definecolor{TableBest}{HTML}{C3E2DE}
\definecolor{TableSecond}{HTML}{E5EDE0}
\definecolor{TableOurs}{HTML}{E2EFED}
\definecolor{TableAccent}{HTML}{426D78}
\definecolor{TableGain}{HTML}{32814B}
\definecolor{TableLoss}{HTML}{B3433D}
\definecolor{TableMuted}{HTML}{ACB9BD}
\definecolor{TableHighlight}{HTML}{E2EEF0}
\definecolor{TableGroup}{HTML}{E3EEF1}

\makeatletter
\newcolumntype{D}[1]{>{\centering\arraybackslash}m{#1}}
\newcolumntype{d}[1]{>{\raggedright\arraybackslash}m{#1}}
\newcommand{\DataYes}{\textcolor{TableGain}{\ding{51}}}
\newcommand{\DataCross}{\textcolor{TableLoss}{\ding{55}}}

\newcolumntype{J}[1]{>{\raggedright\arraybackslash}m{#1}}
\newcolumntype{K}[1]{>{\centering\arraybackslash}m{#1}}
\newcommand{\MainHead}[2]{\parbox[c][29pt][c]{\linewidth}{\centering\bfseries #1\\#2}}
\newcommand{\MainSingleHead}[1]{\parbox[c][29pt][c]{\linewidth}{\centering\bfseries #1}}
\newcommand{\MainMethodHead}{\parbox[c][29pt][c]{\linewidth}{\raggedright\bfseries Method}}
\newcommand{\MainBest}[1]{\cellcolor{TableBest}#1}
\newcommand{\MainSecondBest}[1]{\cellcolor{TableSecond}#1}
\newcommand{\MainUp}[2]{\parbox[c][24pt][c]{\linewidth}{\centering #1\\[-1pt]{\footnotesize\textcolor{TableGain}{$(\uparrow\,#2)$}}}}

\newcommand{\MainSection}[1]{%
  \arrayrulecolor{TableSeparator}\specialrule{.45pt}{0pt}{0pt}%
  \rowcolor{TableGroup}%
  \multicolumn{9}{l}{\hspace*{-\tabcolsep}%
    \textcolor{TableAccent}{\rule[-\dp\@arstrutbox]{2pt}{\dimexpr\ht\@arstrutbox+\dp\@arstrutbox\relax}}%
    \hspace{7pt}\textcolor{TableAccent}{\itshape #1}}\\}
\newcommand{\MainSectionStrong}[1]{%
  \arrayrulecolor{TableRule}\specialrule{1pt}{0pt}{0pt}%
  \rowcolor{TableGroup}%
  \multicolumn{9}{l}{\hspace*{-\tabcolsep}%
    \textcolor{TableAccent}{\rule[-\dp\@arstrutbox]{2pt}{\dimexpr\ht\@arstrutbox+\dp\@arstrutbox\relax}}%
    \hspace{7pt}\textcolor{TableAccent}{\itshape #1}}\\}
\newcolumntype{A}[1]{>{\centering\arraybackslash}m{#1}}
\newcommand{\AblYes}{\textcolor{TableGain}{\ding{51}}}
\newcommand{\AblNo}{\textcolor{TableMuted}{\textendash}}
\newcommand{\AblHead}[1]{\parbox[c][34pt][c]{\linewidth}{\centering\bfseries #1}}
\newcommand{\AblScore}[2]{\parbox[c][28pt][c]{\linewidth}{\centering #1\\[-1pt]{\fontsize{9.5}{10}\selectfont #2}}}
\newcommand{\AblBase}[1]{\parbox[c][28pt][c]{\linewidth}{\centering #1}}
\newcommand{\AblSeparator}[1]{\ensuremath{\vcenter{\hbox{\rule{1pt}{#1}\kern2pt\rule{1pt}{#1}}}}}

\newcommand{\AblFirstCell}[1]{\cellcolor{TableBest}#1}
\newcommand{\AblSecondCell}[1]{\cellcolor{TableSecond}#1}
\makeatother

\definecolor{cvprblue}{rgb}{0.21,0.49,0.74}
\usepackage[pagebackref,breaklinks,colorlinks,allcolors=cvprblue]{hyperref}

\title{HyperGPT: Toward Hyperspectral Image Understanding via Multimodal Large Language Models}
\newcommand{\correspondingauthor}[1]{\textsuperscript{#1,\ding{41}}}

\author{%
  Mingzhen Xu\textsuperscript{1}, Chunyao Wu\textsuperscript{1},
  Liqing Yang\textsuperscript{1}, Dongchen Si\textsuperscript{1},
  Di Wang\correspondingauthor{1,2}, Haonan Guo\correspondingauthor{1,2},
  Bo Du\correspondingauthor{1,2}\\[2pt]
  \textsuperscript{1}Wuhan University\qquad
  \textsuperscript{2}Zhongguancun Academy\\[2pt]
  {\small\fontencoding{T1}\ttfamily
   mingzhenxu@whu.edu.cn, d\_wang@whu.edu.cn,
   haonan.guo@whu.edu.cn, dubo@whu.edu.cn}%
}

\hypersetup{
  pdfauthor={Mingzhen Xu, Chunyao Wu, Liqing Yang, Dongchen Si, Di Wang, Haonan Guo, Bo Du}
}

\begin{document}
\maketitle
\begin{abstract}
Multimodal large language models (MLLMs) have made substantial advances in visual understanding and natural language interaction, but they make limited use of the rich spectral information in hyperspectral imagery (HSI). Existing models cannot directly process HSI and typically rely on a three-channel input or a compressed spectral representation. Such adaptations weaken fine-grained spectral features and obscure their correspondence with spatial locations, thereby hindering accurate HSI interpretation. We introduce HyperGPT, an MLLM for HSI understanding, together with HyperInstruct (100,918 training QA pairs) and HyperBench (20,000 QA pairs). The dataset comprises 34 fine-grained task groups that assess models’ abilities to interpret spectral characteristics, reason about scene semantics and spatial relationships, and infer material composition and abundance. HyperGPT treats HSIs as video data varying along the spectral-channel dimension. It models spectral variation and spatial structure through wavelength-aware spectral positional encoding, spectral-difference residual embeddings, and cross-spectral attention that preserves spatial correspondence. Experimental results show that HyperGPT outperforms existing general-purpose and remote sensing MLLMs on both HyperBench and public benchmarks. Code will be released at https://github.com/Azhenmiddleblock/HyperGPT.git.
\end{abstract}

\section{Introduction}\label{sec:introduction}
\begin{figure}[t]
  \centering
  \includegraphics[width=\columnwidth]{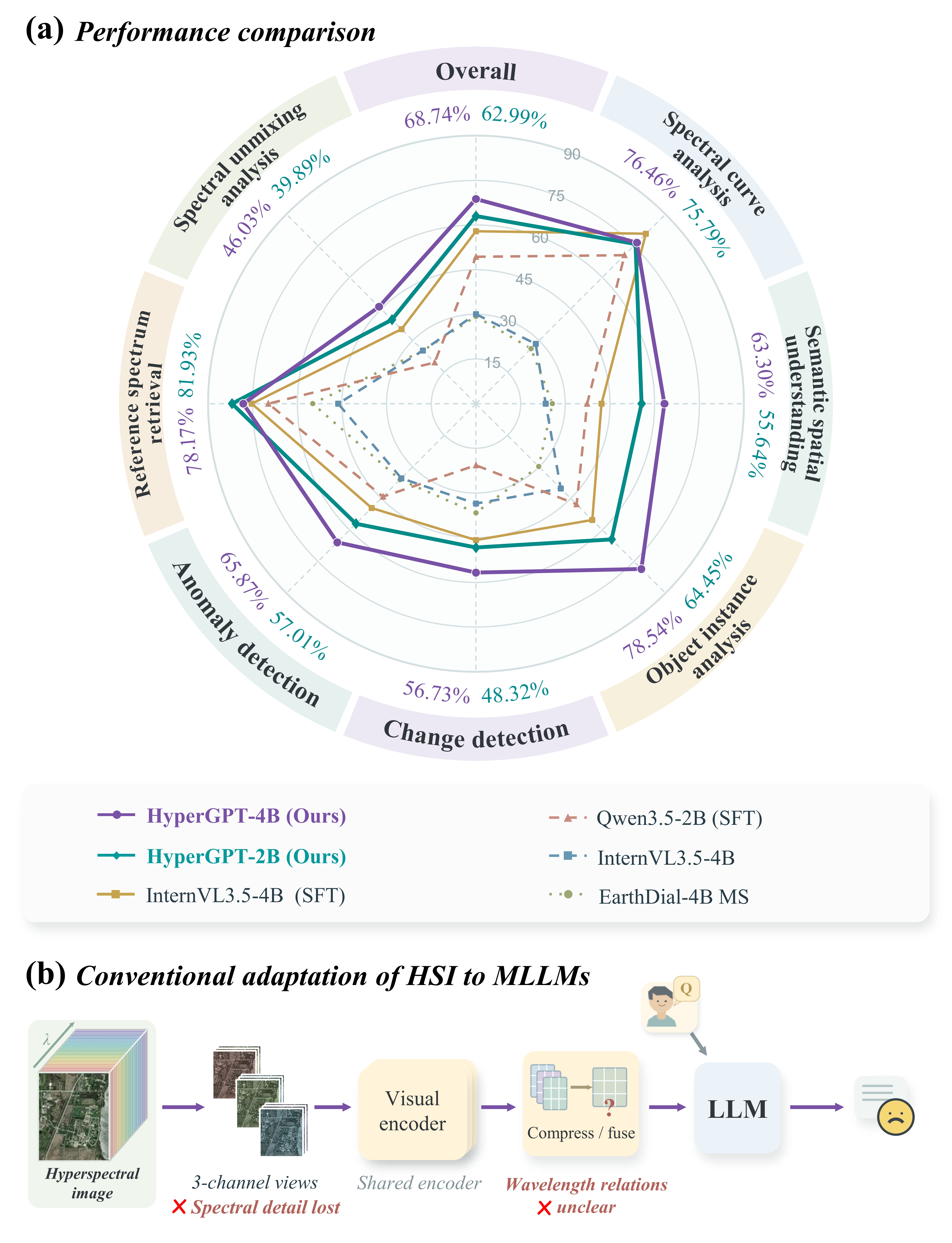}
  \caption{Limitations of MLLMs in HSI understanding. (a) Task-wise performance comparison on HyperBench. (b) Conventional HSI input processing limits spectral–spatial representation.}
  \label{fig:overview}
\end{figure}

Hyperspectral imagery (HSI) records densely sampled narrowband measurements at each spatial location. Its rich spectral information enables materials with similar visual appearances to be distinguished, supporting applications such as land-cover mapping~\cite{zhu2026knowledge,whuhi}, environmental monitoring~\cite{duren2019methane}, and material identification~\cite{goetz1985imaging,enmap2024minerals}. 

For a long time, most HSI research has focused on visual tasks such as classification, detection, and segmentation. Representative models, including SpectralFormer, SpectralGPT, and HyperSIGMA, have improved visual representations by modeling spectral sequences and spectral--spatial relationships \cite{spectralformer,spectralgpt,hypersigma}. However, the downstream applications of these models are largely defined by fixed task objectives and label spaces. This limits their ability to interpret HSI under different semantic conditions.

Recent advances in multimodal large language models (MLLMs) have enabled visual observations to be interpreted through natural-language interaction \cite{clip,flamingo,blip2,instructblip,llava,internvl,qwen25vl,qwen3vl}. Remote sensing models have further extended these capabilities to scene understanding, region-level reasoning, and multisensor analysis \cite{remoteclip,geochat,lhrsbot,earthgpt,earthdial}. Nevertheless, most of these models reuse visual encoders pretrained on three-channel imagery. Their input representations are poorly aligned with the dense spectral structure of HSI, limiting their performance on HSI understanding tasks, as shown in Fig.~\ref{fig:overview}(a). Existing adaptations typically convert HSI into RGB or PCA views, or divide its bands into three-channel groups for visual encoding~\cite{hmbench,earthdial}. Channel compression discards spectral information, while group-wise encoding alone does not explicitly capture wavelength-dependent relationships across groups. Consequently, it is difficult to ensure that spectral variations are correctly interpreted and associated with the correct spatial locations, as illustrated in Fig.~\ref{fig:overview}(b). Using MLLMs for these tasks therefore requires modeling mechanisms suited to the structure of HSIs. To this end, we propose HyperGPT, an MLLM designed for HSI understanding. We treat HSIs as video data varying along the spectral-channel dimension and jointly encode neighboring bands. Spectral Difference Projection Embedding (SDPE) is used to emphasize response changes between adjacent spectral groups. Cross-spectral position-wise attention (CSPA) modules at intervals in the feature extractor let distant spectral groups exchange information at the same spatial location. Spectral positional encodings (SPE) are computed from the representative wavelength of each spectral block. The encoded visual features are then mapped to the language model through a visual projection layer. Together, these components enable the model to relate spectral responses and spatial locations to natural language queries.

Learning to answer questions from spectral-spatial features also requires suitable supervision. Conventional HSI datasets provide annotations for predefined visual tasks, but rarely connect observations with natural-language questions and answers \cite{whuhi,hsidrive}. For example, SpectralEarth \cite{spectralearth} has a large number of samples but includes only two types of visual tasks. A few studies have begun to explore the construction of image-text data, which substantially expands the range of task types. Existing remote sensing language datasets mainly emphasize scene descriptions and spatial semantics \cite{rsicd,rsvqa,rs5m,vrsbench,xlrsbench}. HyperCap introduces pixel-level HSI text descriptions \cite{hypercap}, while HM-Bench provides HSI question answering (QA) and a training-free evidence organization framework \cite{hmbench}. However, acquiring textual annotations for such datasets is complex and costly, which generally constrains sample size, and training datasets that enable MLLMs to systematically acquire hyperspectral domain knowledge remain lacking. To this end, we construct HyperInstruct and HyperBench, an HSI QA training dataset and an evaluation benchmark for MLLMs, respectively. Together, they cover 34 fine-grained task groups across seven categories. Our contributions are as follows.

\begin{itemize}
\item
  We propose HyperGPT, the first MLLM designed for HSI understanding, and treat HSIs as video data varying along the spectral dimension. By introducing SDPE, CSPA, and SPE, HyperGPT achieves improved spectral–spatial representation. This design supports diverse tasks and accommodates HSI from different sensors and at varying scales, substantially expanding the range of tasks that MLLMs can address.
\item
  We construct HyperInstruct, a training dataset comprising 100,918 HSI QA pairs, and HyperBench, a benchmark containing 20,000 QA pairs. They cover 7 task categories and 34 fine-grained task groups. These complementary resources support supervised training and comprehensive evaluation of HSI understanding capabilities.
\item
  Extensive experiments demonstrate that HyperInstruct and HyperBench address gaps in existing datasets for HSI understanding and broadens the range of applications covered in evaluation, and show that HyperGPT performs well compared to existing general and remote sensing MLLMs on multiple HSI understanding tasks.
\end{itemize}

\section{HyperInstruct and HyperBench}\label{sec:dataset}

\begin{figure}[t]
\centering
\includegraphics[width=0.95\columnwidth]{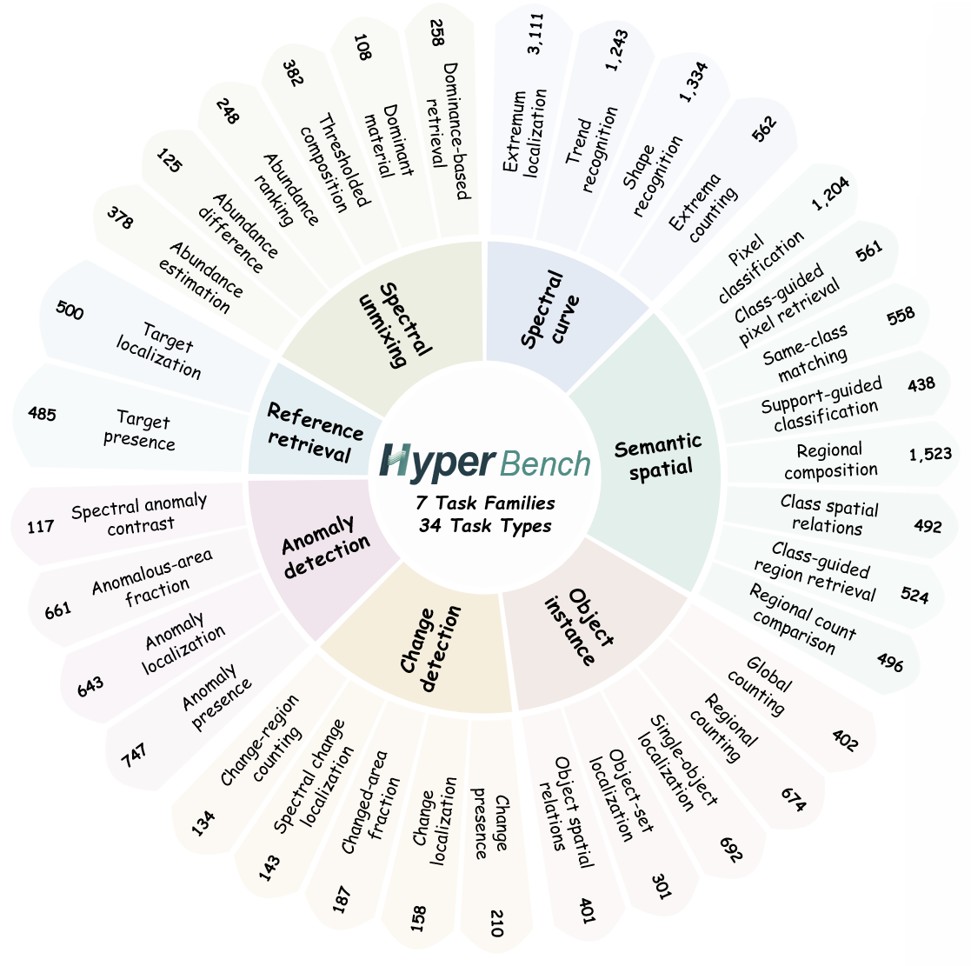}
\caption{Illustration of the 34 HSI task categories.}
\label{fig:task_categories}
\end{figure}

We construct HyperInstruct and HyperBench from 38 standardized source datasets comprising 4,754 HSIs acquired across different sensors, spatial resolutions, and spectral ranges~\cite{quh,whuhi,szutree,harvard,hyperspectralcity,cave,chikusei,muufl,houston2013,houston2018,xiongan,indianpines,szucar,hod3k,spod,change_original,change_data,abu,sandiego,unmixing_data,gff,xuzhou,washingtondc}. We first collect HSIs and associated metadata, including wavelengths, spatial annotations, temporal correspondences, reference spectra, and endmembers. Structured prompts and diverse sentence patterns and expressions were designed for different tasks. Task-specific questions and answer options were then generated with GPT-6.1 assistance. Correct answers were determined from source annotations or reproducible computational rules. Quality control includes automated consistency checks, statistical audits, and manual sampling inspections involving ten invited domain experts. The resulting datasets contain 100,918 training QA pairs (HyperInstruct) and 20,000 benchmark QA pairs (HyperBench). Further details of data collection, QA generation, and quality control are provided in the Appendix.

HyperBench organizes 34 fine-grained task groups into 7 categories, as shown in Fig.~\ref{fig:task_categories}. Spectral curve analysis examines spectral signatures and their variations, while spatial semantic understanding focuses on scene content and spatial relationships. Object instance analysis addresses the properties of individual objects. Change detection compares bi-temporal observations, whereas anomaly detection focuses on identifying unusual spectral or spatial responses. Reference spectrum retrieval matches observed spectra with reference signatures, and spectral unmixing analysis examines material composition using endmember information. These tasks share an evidence-grounded choice format but differ in their required spectral, spatial, temporal, and material-related information. Together, they provide a unified framework for evaluating different aspects of HSI understanding. A comparison with existing datasets is shown in Table~\ref{tab:dataset_comparison}. Our dataset is distinguished by its comprehensive task coverage, containing nearly three times as many tasks as the most task-diverse existing dataset. It includes HSI knowledge in both visual and textual forms, and its number of samples is approximately six times that of other image-text datasets. More importantly, it is the first dataset to support both HSI question-answering training and evaluation.

\begin{table}[t]
\centering
\captionsetup{skip=4pt}
\caption{Comparison of different HSI datasets.}
\label{tab:dataset_comparison}
\begingroup
\newcommand{\DatasetHead}[1]{\parbox[c][18pt][c]{\linewidth}{\centering\bfseries \mbox{#1}}}
\newcommand{\DatasetBase}[1]{\parbox[c][12pt][c]{\linewidth}{\centering #1}}
\color{TableInk}
\setlength{\tabcolsep}{1.5pt}
\renewcommand{\arraystretch}{1.00}
\setlength{\extrarowheight}{0pt}
\fontsize{8.465}{9.31}\selectfont
\renewcommand{\tabularxcolumn}[1]{m{#1}}
\begin{tabularx}{\columnwidth}{d{\dimexpr\columnwidth-129pt\relax}D{27pt}D{29pt}D{23pt}D{35pt}}
\arrayrulecolor{TableRule}\specialrule{1pt}{0pt}{0pt}
\rowcolor{TableHeader}
\DatasetHead{Dataset} & \DatasetHead{Tasks} & \DatasetHead{Visual} & \DatasetHead{Text} & \DatasetHead{Samples} \\
\arrayrulecolor{TableRule}\specialrule{1pt}{0pt}{0pt}
Hyper-Seg~\cite{hyperfree} & \DatasetBase{1} & \DataYes & \DataCross & \mbox{41,900} \\
SpectralEarth \cite{spectralearth} & \DatasetBase{2} & \DataYes & \DataCross & \mbox{538,974} \\
WHU-OHS \cite{whuohs} & \DatasetBase{1} & \DataYes & \DataCross & \mbox{7,795} \\
HSOD-BIT-V2 \cite{hsodbitv2} & \DatasetBase{1} & \DataYes & \DataCross & \mbox{500} \\
HyperCOD \cite{hypercod} & \DatasetBase{1} & \DataYes & \DataCross & \mbox{350} \\
HM-Bench~\cite{hmbench} & \DatasetBase{13} & \DataYes & \DataYes & \mbox{19,337} \\
HyperCap~\cite{hypercap} & \DatasetBase{2} & \DataYes & \DataYes & \mbox{21,237} \\
HyperImageNet~\cite{hyperimagenet} & \DatasetBase{2} & \DataYes & \DataCross & \mbox{26,084} \\
\mbox{NLCD-hypercaption}~\cite{dofaclip} & \DatasetBase{2} & \DataYes & \DataYes & \mbox{15,000} \\
\arrayrulecolor{TableSeparator}\specialrule{.4pt}{0pt}{0pt}
\rowcolor{TableOurs}
\textbf{Ours} & \DatasetBase{34} & \DataYes & \DataYes & \mbox{120,918} \\
\arrayrulecolor{TableRule}\specialrule{1pt}{0pt}{0pt}
\end{tabularx}
\endgroup
\end{table}

\begin{figure*}[t]
\centering
\includegraphics[width=\textwidth]{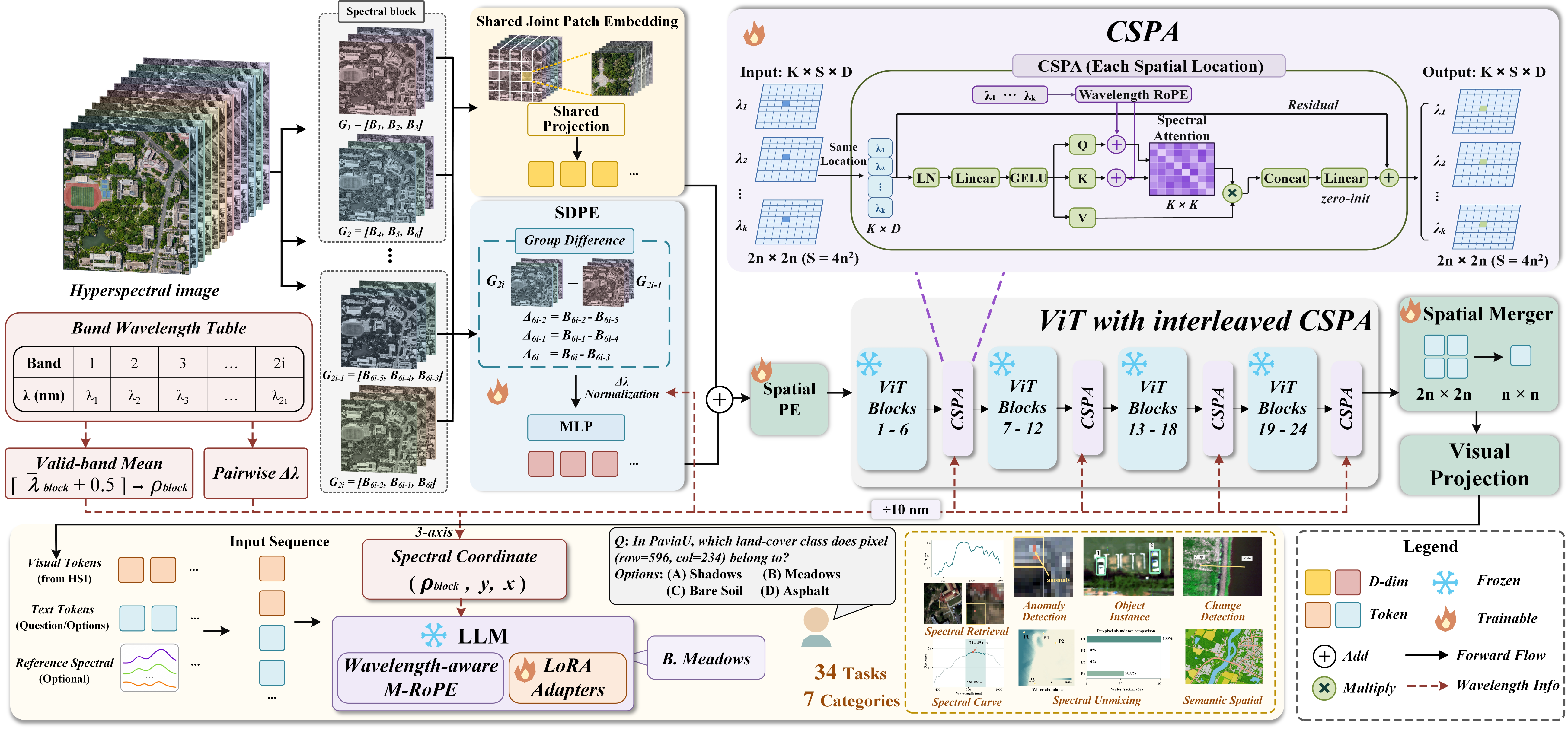}
\caption{Overall architecture of HyperGPT. The HSI is organized into three-band pseudo-frames and paired spectral blocks. Shared patch embedding and SDPE initialize the visual tokens; interleaved CSPA layers exchange information across spectral blocks at matching spatial positions. Wavelength-derived spectral positions are passed to cross-spectral attention and multimodal positional encoding.}
\label{fig:hypergpt_architecture}
\end{figure*}

\section{Method}\label{sec:method}
HyperGPT aims to predict answers based on hyperspectral observations, natural-language questions, and task-specific auxiliary information. Built upon the pretrained Qwen3.5 MLLM \cite{qwen35}, HyperGPT follows a unified pipeline from observation processing to answer prediction. Inputs are organized according to the query, with Spectral Difference Projection Embedding (SDPE) enhancing their initial representations. Cross-Spectral Position-wise Attention (CSPA) is interleaved within the vision Transformer to progressively extract spectral-spatial features while preserving spatial correspondence. Wavelength-aware positional information is incorporated into both visual feature interaction and language reasoning. Finally, the visual features, auxiliary spectral information, and question text are assembled into a unified sequence, and the model is jointly fine-tuned using answer supervision, as shown in Fig.~\ref{fig:hypergpt_architecture}.

\subsection{Input Organization and SDPE}\label{sec:input_sdpe}

Let \(X \in \mathbb{R}^{H \times W \times B}\) denote an HSI with spatial dimensions \(H \times W\) and \(B\) spectral bands. To preserve the original spectral measurements while reusing the pretrained visual encoder, we organize the HSI as a sequence of pseudo-frames along the spectral dimension. Following the original band order, every three consecutive bands form a channel pseudo-frame. Two adjacent pseudo-frames are then paired into a spectral block and jointly processed by the two-frame patch embedding. Each spectral block thus contains up to six original bands, yielding \(G = \left\lceil B/6 \right\rceil\) spectral blocks. This organization reuses the pretrained visual input interface without compressing the spectra into three PCA channels or performing frame subsampling or spectral resampling. Missing channels or pseudo-frames in the final spectral block are padded by replication to match the required input dimensions. A separate band-validity mask distinguishes original spectral values from replicated values.

The pretrained patch projection aggregates local observations but does not explicitly model spectral differences. SDPE therefore adds a residual derived from local spectral variations. Let \(P^{(0)}_{g,c,s}\) and \(P^{(1)}_{g,c,s}\) denote the corresponding spatial patches in the two pseudo-frames of spectral block \(g\), after the backbone's channel-wise normalization. Here, \(c\in\{1,2,3\}\) indexes the channels, \(s\) indexes spatial patches, and \(\sigma_c\) is the normalization standard deviation of channel \(c\). The spectral difference is computed as

\begin{equation}\label{eq:sdpe_difference}
\delta_{g,c,s}=
\begin{cases}
\displaystyle
\frac{\sigma_c
\left(P^{(1)}_{g,c,s}-P^{(0)}_{g,c,s}\right)}
{d_{g,c}},
& m_{g,c}=1,\\
0, & m_{g,c}=0.
\end{cases}
\end{equation}
Here, \(m_{g,c}=1\) only when both channels correspond to valid, non-padded bands. For physical wavelength coordinates, their wavelength difference must also be positive. The normalization factor is defined as

\begin{equation}\label{eq:sdpe_norm}
\begin{aligned}
d_{g,c}
&=
\begin{cases}
\displaystyle
\frac{\max(\Delta\lambda_{g,c},1\,\mathrm{nm})}
{10\,\mathrm{nm}},
& \text{physical wavelengths},\\
1, & \text{band indices only},
\end{cases}\\
\Delta\lambda_{g,c}
&=\lambda^{(1)}_{g,c}-\lambda^{(0)}_{g,c}.
\end{aligned}
\end{equation}

The wavelength differences are calculated from the original, unrounded metadata. Corresponding channels in the two pseudo-frames are separated by three positions in the original band sequence. Multiplication by \(\sigma_c\) restores the response-difference scale before channel-wise normalization. For index-only inputs, the response differences are used without wavelength normalization.

Let \(P_{g,s}\) stack the two normalized three-channel patches,
and let \(\delta_{g,s}\) stack their three channel-wise difference
patches. To account for incomplete spectral blocks, we define
\(\rho_g=\frac{1}{3}\sum_{c=1}^{3}m_{g,c}\) as the fraction of
valid channel pairs and use it to scale the SDPE residual.
The initial visual embedding is

\begin{equation}\label{eq:sdpe_embed}
\begin{aligned}
h^{(0)}_{g,s}
={}&E_{\mathrm{patch}}(P_{g,s})+e^{\mathrm{sp}}_s\\
&+\rho_g W_D^{\mathrm{up}}
\mathrm{GELU}\!\left(
\mathrm{LN}\!\left(
W_D^{\mathrm{down}}
\mathrm{vec}(\delta_{g,s})
\right)\right).
\end{aligned}
\end{equation}
Here, \(E_{\mathrm{patch}}\) is the pretrained patch embedding layer, and \(e^{\mathrm{sp}}_s\) is the spatial positional embedding inherited from the backbone. The difference patches are flattened and passed through a bottleneck projection, LayerNorm, and GELU before being projected back to the visual hidden dimension. \(W_D^{\mathrm{up}}\) is initialized to zero, so SDPE initially preserves the pretrained feature pathway. Its contribution is subsequently learned through question-answering supervision.

\subsection{Visual Encoding with CSPA}\label{sec:cspa}

SDPE captures local spectral variations, but distinguishing different objects also requires modeling response relationships across more distant wavelengths. To establish these relationships progressively during visual feature extraction, we insert CSPA after selected layers of the pretrained vision transformer~\cite{vit}. Let \(T_{\ell}\) denote the \(\ell\)-th original vision transformer layer and \(C_{\ell}\) the subsequent spectral adaptation module. This visual encoding process is formulated as

\begin{equation}\label{eq:interleaved_vit}
H^{(\ell)}=C_{\ell}\!\left(T_{\ell}\!\left(H^{(\ell-1)}\right)\right).
\end{equation}

The original vision transformer layers retain the backbone\textquotesingle s spatial modeling and positional encoding mechanisms, while CSPA enables interactions across spectral blocks at corresponding spatial positions. These operations alternately update the same feature grid, progressively integrating local spectral variations with broader spectral-spatial relationships. The local spectral information introduced by SDPE can therefore propagate through the visual encoder, while CSPA establishes interactions across a broader wavelength range at different feature depths.

For a fixed spatial position \(s\), we collect the features of all \(G\) spectral blocks into \(H_{s} = \left\lbrack h_{1,s},\ldots,h_{G,s} \right\rbrack^{\top}\). CSPA first normalizes and projects these features into a lower-dimensional space, then generates the queries, keys, and values for multi-head attention:

\begin{equation}\label{eq:cspa_qkv}
\begin{aligned}
u_{g,s} &= \mathrm{GELU}\!\left(W_C^{\mathrm{down}}\mathrm{LN}(h_{g,s})\right),\\
[q_{g,s},k_{g,s},v_{g,s}] &= W_{\mathrm{qkv}}u_{g,s}.
\end{aligned}
\end{equation}

To account for wavelength separations between spectral blocks in attention computation, we apply spectral rotary positional encoding to the queries and keys. Let \(\xi_g\) denote the spectral coordinate of block \(g\), and let \(R(\xi_g)\) denote its rotation operator, as defined in Sec.~\ref{sec:positional}. The projected queries, keys, and values are split into \(n_h\) attention heads. For head \(a\), the attention weights are computed as

\begin{equation}\label{eq:cspa_attention}
\begin{aligned}
\widetilde q_{i,s}^{(a)}
&=R(\xi_i)q_{i,s}^{(a)},\\
\widetilde k_{j,s}^{(a)}
&=R(\xi_j)k_{j,s}^{(a)},\\
A_{ij}^{(s,a)}
&=\operatorname{softmax}_{j}
\left(
\frac{
\left(\widetilde q_{i,s}^{(a)}\right)^\top
\widetilde k_{j,s}^{(a)}
}{\sqrt{d_h}}
\right).
\end{aligned}
\end{equation}
Here, \(i,j\in\{1,\ldots,G\}\) index the query and key spectral blocks, respectively, and \(d_h\) is the dimension of each attention head. At each spatial position \(s\), each spectral block attends to all \(G\) spectral blocks. The head outputs are concatenated, projected back to the visual hidden dimension, and added through a residual connection:

\begin{equation}\label{eq:cspa_output}
h'_{i,s}
={}h_{i,s}+W_C^{\mathrm{up}} \left[
\operatorname{Concat}_{a=1}^{n_h}
\left(
\sum_{j=1}^{G}
A_{ij}^{(s,a)}v_{j,s}^{(a)}
\right)\right].
\end{equation}

This operation enables feature interactions across different wavelength ranges at the same spatial position, while preserving the spatial association and spectral block order of each output token.

\subsection{Spectral Positional Encoding}\label{sec:positional}

Spectral band indices indicate sampling order but do not provide consistent physical wavelength information across sensors. The same band index may correspond to different wavelengths, while variations in sampling intervals also affect spectral distances. To address these differences, we construct spectral coordinates from original wavelength metadata and incorporate them into both cross-spectral attention in the visual encoder and multimodal positional encoding in the language model. Although both mechanisms derive positions from the same valid spectral bands, they serve different purposes: the former incorporates wavelength differences between spectral blocks into cross-spectral attention, whereas the latter specifies the spectral positions of visual evidence passed to the language model.

Let \(B_g\) denote the band slots of spectral block \(g\), and let \(v_b\) indicate whether a slot contains a valid, non-padded measurement. Let \(b\) denote its one-based native band index. When physical wavelength metadata are available, \(\lambda_b\) denotes the corresponding wavelength. Let \(\bar{\lambda}_g\) and \(\bar{b}_g\) denote the mean wavelength and mean native band index over valid measurements in block \(g\), respectively. The integer spectral position \(p_g\) is computed as follows:

\begin{equation}\label{eq:spectral_wavelength}
\begin{aligned}
\bar{\lambda}_g
&=\frac{\sum_{b\in B_g}v_b\lambda_b}
        {\sum_{b\in B_g}v_b},\\
\bar{b}_g
&=\frac{\sum_{b\in B_g}v_b b}
        {\sum_{b\in B_g}v_b},\\
p_g
&=
\begin{cases}
\displaystyle
\left\lfloor
\frac{\bar{\lambda}_g}{1\,\mathrm{nm}}+0.5
\right\rfloor,
& \text{physical wavelengths},\\
\kappa+\left\lfloor\bar{b}_g+0.5\right\rfloor,
& \text{band indices only},
\end{cases}\\
\kappa&=16384.
\end{aligned}
\end{equation}

Here, \(\kappa=16384\) is a fixed offset applied to index-based
positions to distinguish them from wavelength-based positions. The mean wavelength is evaluated only when physical wavelength metadata are available. Replicated channels, replicated pseudo-frames, and invalid bands are excluded from both means. All physical wavelength positions use the same 1 nm scale, without normalization by each data source's wavelength range. Integer quantization affects only the positional coordinates. SDPE continues to use the original, unrounded wavelengths. The offset \(\kappa\) separates index-based positions from physical wavelength positions; it does not represent a physical wavelength.

For CSPA, the spectral phase coordinate is defined as
\begin{equation}
\xi_g=
\begin{cases}
p_g/10, & \text{physical wavelengths},\\
p_g-\kappa, & \text{band indices only}.
\end{cases}
\end{equation}

Spectral rotary positional encoding (RoPE)~\cite{roformer} rotates adjacent pairs of query and key dimensions. For an attention head of dimension \(d_h\), the rotation operator is
\begin{equation}
\begin{aligned}
R(\xi)
&=\operatorname{diag}
\left(R_0(\xi),\ldots,R_{d_h/2-1}(\xi)\right),\\
R_r(\xi)
&=
\begin{bmatrix}
\cos(\xi\omega_r)&-\sin(\xi\omega_r)\\
\sin(\xi\omega_r)&\cos(\xi\omega_r)
\end{bmatrix},\\
\omega_r&=\theta^{-2r/d_h},
\qquad \theta=10000.
\end{aligned}
\end{equation}

The value vectors remain unchanged. The attention scores in Eq.~\eqref{eq:cspa_attention} therefore depend on both spectral content and coordinate differences.

On the language side, we reuse the backbone's multimodal rotary positional encoding interface~\cite{qwen25vl,qwen3vl,qwen35}. Only the original temporal coordinate is replaced with \(p_g\). The height and width coordinates generated by the backbone are retained. After spatial merging, the three-axis position of a visual token is

\begin{equation}\label{eq:mrope_coordinates}
\mathbf{p}_{g,s}^{\mathrm{LLM}}
=
\left(
p_g,\,
p_{g,s}^{h,\mathrm{base}},\,
p_{g,s}^{w,\mathrm{base}}
\right).
\end{equation}
Here, \(s\) indexes the spatially merged visual tokens, and the superscript \(\mathrm{base}\) denotes positions inherited from the backbone. Spatial merging preserves the spectral block identity of each token.

\begin{table*}[t]
\centering
\captionsetup{skip=4pt}
\caption{Overall and task-wise accuracy (\%) on HyperBench with different models.}
\label{tab:main_comparison}
\begingroup
\centering
\color{TableInk}
\setlength{\tabcolsep}{4pt}
\renewcommand{\arraystretch}{1.20}
\setlength{\extrarowheight}{0pt}
\fontsize{10}{11}\selectfont
\resizebox{\linewidth}{!}{%
\begin{tabular}{J{5.9cm}K{1.35cm}*{7}{K{1.55cm}}}
\arrayrulecolor{TableRule}\specialrule{1pt}{0pt}{0pt}
\rowcolor{TableHeader}
\MainMethodHead & \cellcolor{TableHeader}\MainSingleHead{Overall}
& \MainHead{Spectral}{curve} & \MainHead{Semantic}{space}
& \MainHead{Object}{instance} & \MainHead{Change}{detection}
& \MainHead{Anomaly}{detection} & \MainHead{Reference}{retrieval}
& \MainHead{Spectral}{unmixing} \\
\MainSectionStrong{MLLMs (Direct Evaluation)}
Qwen3.5-2B & 26.56 & 25.89 & 24.26 & 23.68 & 20.67 & 40.82 & 29.24 & 23.88 \\
Qwen3.5-4B & 28.29 & 29.18 & 26.00 & 27.85 & 28.00 & 31.83 & 32.69 & 26.22 \\
Qwen3-VL-2B & 29.97 & 26.85 & 26.57 & 38.79 & 31.85 & 32.66 & 44.47 & 27.08 \\
InternVL3.5-4B & 30.05 & 28.40 & 23.41 & 40.28 & 33.53 & 35.47 & 46.29 & 25.22 \\
MiniCPM-V-4.7-1B & 26.07 & 24.74 & 24.60 & 26.60 & 22.96 & 32.24 & 32.79 & 24.82 \\
SmolVLM2-2.2B & 14.44 & 12.90 & 11.37 & 27.61 & 0.00 & 34.18 & 0.00 & 0.00 \\
GeoChat-7B & 15.35 & 14.00 & 10.70 & 31.66 & 0.00 & 36.53 & 0.00 & 0.00 \\
EarthDial-4B-MS & 29.31 & 26.13 & 25.67 & 29.80 & 36.66 & 36.07 & 54.82 & 25.15 \\
\MainSection{MLLMs (Supervised Fine-tuning)}
Qwen3.5-2B & 49.31 & 70.54 & 37.30 & 47.77 & 20.67 & 44.14 & 69.75 & 19.61 \\
InternVL3.5-4B & 57.89 & \MainBest{80.67} & 42.17 & 55.22 & 45.79 & 49.54 & 75.33 & 35.36 \\
\MainSection{HyperGPT}
HyperGPT-2B
& \MainSecondBest{\MainUp{62.99}{13.68}}
& \MainUp{75.79}{5.25}
& \MainSecondBest{\MainUp{55.64}{18.34}}
& \MainSecondBest{\MainUp{64.45}{16.68}}
& \MainSecondBest{\MainUp{48.32}{27.65}}
& \MainSecondBest{\MainUp{57.01}{12.87}}
& \MainBest{\MainUp{81.93}{12.18}}
& \MainSecondBest{\MainUp{39.89}{20.28}} \\
HyperGPT-4B
& \MainBest{\MainUp{68.74}{19.43}}
& \MainSecondBest{\MainUp{76.46}{5.92}}
& \MainBest{\MainUp{63.30}{26.00}}
& \MainBest{\MainUp{78.54}{30.77}}
& \MainBest{\MainUp{56.73}{36.06}}
& \MainBest{\MainUp{65.87}{21.73}}
& \MainSecondBest{\MainUp{78.17}{8.42}}
& \MainBest{\MainUp{46.03}{26.42}} \\
\arrayrulecolor{TableRule}\specialrule{1pt}{0pt}{0pt}
\end{tabular}%
}
\par\vspace{2pt}
{\footnotesize\raggedright\color{TableAccent}
\textcolor{TableBest}{\rule{8pt}{7pt}}\ Highest score\qquad
\textcolor{TableSecond}{\rule{8pt}{7pt}}\ Second-highest score\qquad
\textcolor{TableGain}{$\uparrow$} / \textcolor{TableLoss}{$\downarrow$}:
score-point gain / drop relative to the Qwen3.5-2B with supervised fine-tuning.\par}
\endgroup
\end{table*}

\subsection{Sequence Construction and QA Training}\label{sec:multimodal_training}

After SDPE and the interleaved visual and spectral encoding stages, the final visual features \(H^{(L)}\)are processed by the backbone\textquotesingle s spatial merger \(M\). This module aggregates neighboring spatial patches and projects them into the language model\textquotesingle s hidden dimension while preserving the spectral block order:

\begin{equation}\label{eq:spatial_merge}
Z=M\!\left(H^{(L)}\right)\in\mathbb{R}^{N_v\times d_{\mathrm{LLM}}}.
\end{equation}
Here, \(N_{v}\) is the number of visual tokens after spatial merging, and \(d_{LLM}\) denotes the hidden dimension of the language model. The question, all answer options, publicly available query conditions, and necessary geometric information are processed by the text embedding layer. The visual features are inserted into their corresponding visual placeholder positions.

For a batch of \(N\) examples, let \(C_n\) and \(y_n\) denote the multimodal context and target answer sequence of example \(n\), and \(A_n\) its valid answer-token positions, including the end-of-sequence marker. The autoregressive cross-entropy loss is normalized by the total number of supervised tokens:

\begin{equation}\label{eq:qa_loss}
\mathcal{L}_{\mathrm{QA}}
={}-\frac{1}{\sum_{n=1}^{N}|A_n|} \sum_{n=1}^{N}\sum_{t\in A_n}
\log p_{\theta}
\left(y_{n,t}\mid C_n,y_{n,<t}\right).
\end{equation}

Only answer tokens and end-of-sequence markers are included in the loss computation and normalization. For multiple-choice question answering, the target sequence consists of the correct option identifier and its associated text, allowing supervision to constrain both option selection and answer semantics.

\section{Experiments}\label{sec:experiments}

Experiments are conducted to address three research questions: (1) Can the proposed HyperInstruct and HyperBench provide a domain-specific knowledge foundation for HSI understanding? (2) Does HyperGPT outperform existing MLLMs in HSI understanding? (3) Do the proposed strategies and architectural components improve the model\textquotesingle s understanding of HSI? Consistent input protocols and evaluation criteria are first established to compare different models on HyperBench, with supervision for domain-specific training provided by HyperInstruct. We then conduct ablation studies to assess the contributions of individual components. In addition, we evaluate HyperGPT on an existing public HSI benchmark to examine its cross-benchmark performance. Finally, we present representative QA examples to illustrate the behavior of HyperGPT.

\subsection{Experimental Settings}\label{sec:settings}

\paragraph{Model Implementation and Training Configuration} We develop HyperGPT in two variants, HyperGPT-2B and HyperGPT-4B, based on the pretrained Qwen3.5. The CSPA modules are inserted after the 6th, 12th, 18th, and 24th layers of the 24-layer visual encoder. The original weights of both the visual and language backbones are frozen, while SDPE, CSPA, the visual merger, and the necessary numerical input interface parameters are jointly optimized. LoRA~\cite{lora} is applied to the language layers for parameter-efficient adaptation. LoRA is configured with a rank of 16, a scaling parameter of 32, and a dropout rate of 0.05. Optimization uses AdamW~\cite{adamw} with a warmup phase followed by cosine learning rate decay.

\paragraph{Comparison Methods and Input Protocols} HyperGPT is compared with general-purpose and remote sensing MLLMs. The general-purpose models include Qwen3.5-2B and Qwen3.5-4B~\cite{qwen35}, Qwen3-VL-2B~\cite{qwen3vl}, InternVL3.5-4B-HF~\cite{internvl35}, MiniCPM-V-4.7-1B~\cite{minicpm47}, and SmolVLM2-2.2B~\cite{smolvlm2}. The remote sensing MLLMs are GeoChat-7B~\cite{geochat} and EarthDial-4B-MS~\cite{earthdial}. For direct evaluation of frozen models, only deterministic input transformations and message-format adaptations are applied. Models supporting video or multiple images receive ordered three-band groups. GeoChat receives three-channel PCA images to fit its single-image interface, while EarthDial-4B-MS follows its original three-band group encoding and average-fusion configuration. In addition to frozen-model comparisons, we include two supervised fine-tuning methods built upon Qwen3.5-2B and InternVL3.5-4B, respectively, both adopting EarthDial-style grouped feature fusion~\cite{earthdial}. Samples exceeding a model's context are skipped and are not counted as correct predictions.

\paragraph{Evaluation Metrics} We use choice accuracy over all test questions as the primary evaluation metric. Let \(N\) denote the total number of test questions, \(y_i\) the ground-truth answer option, and \(\hat y_i\) the predicted option parsed according to a unified evaluation protocol. Accuracy is calculated as
\begin{equation}\label{eq:accuracy}
\mathrm{Acc}_{\mathrm{all}}
= \frac{100}{N}\sum_{i=1}^{N}
\mathbf{1}\!\left[\hat{y}_i = y_i\right].
\end{equation}
Here, $\mathbf{1}[\cdot]$ is the indicator function, which equals
1 if the condition holds and 0 otherwise.

\subsection{Comparative Experiments}\label{sec:comparisons}

We evaluate 8 comparison methods, including 6 general-purpose and 2 remote sensing MLLMs. Table~\ref{tab:main_comparison} shows that most frozen MLLMs achieve approximately 30\% accuracy. SmolVLM2 and GeoChat cannot fully process some inputs because of context-length constraints, further reducing their overall accuracy. Fine-tuning Qwen3.5-2B and InternVL3.5-4B with grouped feature fusion on HyperInstruct improves accuracy by 22.75 and 27.84 percentage points, respectively, indicating the value of HyperInstruct.

After supervised fine-tuning, HyperGPT-2B and HyperGPT-4B achieve accuracies of 62.99\% and 68.74\%, respectively. These represent improvements of 36.43 and 40.45 percentage points over their corresponding pretrained Qwen3.5 baselines. Compared with the fine-tuned Qwen3.5-2B baseline using grouped feature fusion, HyperGPT-2B further improves overall accuracy by 13.68 percentage points and outperforms it across all seven task categories. HyperGPT-4B also exceeds the fine-tuned InternVL3.5-4B by 10.85 percentage points overall. These results demonstrate the effectiveness of explicit spectral-spatial modeling for HSI understanding tasks over grouped feature aggregation.

\subsection{Ablation Studies}\label{sec:ablation}

We conduct ablation studies on HyperGPT-2B to evaluate the contributions of its hyperspectral image input organization strategy, spectral positional encoding (SPE), SDPE, and CSPA, as shown in Table~\ref{tab:ablation}. Compared with independent three-channel encoding and representation fusing, video-like input organization improves overall accuracy from 49.31\% to 57.93\%, demonstrating the benefit of joint spectral sequence modeling. Incorporating SPE further increases accuracy to 60.30\%, while adding SDPE or CSPA yields 61.47\% and 61.57\%, respectively. The full model achieves 62.99\%, outperforming the video-like input baseline by 5.06 percentage points. These results demonstrate the complementary contributions of the proposed components. More detailed analyses are provided in the Appendix.

\begin{table}[t]
\centering
\captionsetup{skip=4pt}
\caption{Ablation results (accuracy, \%) on HyperBench.}
\label{tab:ablation}
\begingroup
\renewcommand{\AblBase}[1]{\parbox[c][16pt][c]{\linewidth}{\centering #1}}

\newcommand{\AblScoreInline}[2]{\parbox[c][16pt][c]{\linewidth}{\centering #1\hspace{0.5pt}{\fontsize{8.5}{9}\selectfont #2}}}
\renewcommand{\AblScore}[2]{\parbox[c][22pt][c]{\linewidth}{\centering #1\\[-2pt]{\fontsize{8.5}{9}\selectfont #2}}}
\renewcommand{\AblHead}[1]{\parbox[c][22pt][c]{\linewidth}{\centering\bfseries #1}}
\color{TableInk}
\setlength{\tabcolsep}{4pt}
\renewcommand{\arraystretch}{1.20}
\setlength{\extrarowheight}{0pt}
\fontsize{9.123}{10.035}\selectfont
\resizebox{\linewidth}{!}{%

\begin{tabular}{J{2.8cm}A{20pt}A{28pt}A{28pt}|A{57pt}}

\arrayrulecolor{TableRule}\specialrule{1pt}{0pt}{0pt}
\rowcolor{TableHeader}
\AblHead{Method} & \AblHead{SPE} & \AblHead{SDPE} & \AblHead{CSPA} & \AblHead{Overall} \\
\arrayrulecolor{TableRule}\specialrule{1pt}{0pt}{0pt}
\AblBase{Qwen3.5-2B} & \AblBase{\AblNo} & \AblBase{\AblNo} & \AblBase{\AblNo} & \AblBase{49.31} \\
\AblBase{HyperGPT-2B} & \AblBase{\AblNo} & \AblBase{\AblNo} & \AblBase{\AblNo}
& \AblScoreInline{57.93}{\textcolor{TableGain}{(+8.62)}} \\
\AblBase{HyperGPT-2B} & \AblBase{\AblYes} & \AblBase{\AblNo} & \AblBase{\AblNo} & \AblScoreInline{60.30}{\textcolor{TableGain}{(+10.99)}} \\
\AblBase{HyperGPT-2B} & \AblBase{\AblYes} & \AblBase{\AblYes} & \AblBase{\AblNo} & \AblScoreInline{61.47}{\textcolor{TableGain}{(+12.16)}} \\
\AblBase{HyperGPT-2B} & \AblBase{\AblYes} & \AblBase{\AblNo} & \AblBase{\AblYes} & \AblScoreInline{61.57}{\textcolor{TableGain}{(+12.26)}} \\
\AblBase{HyperGPT-2B} & \AblBase{\AblNo} & \AblBase{\AblYes} & \AblBase{\AblYes} & \AblScoreInline{61.71}{\textcolor{TableGain}{(+12.40)}} \\
\AblBase{HyperGPT-2B} & \AblBase{\AblYes} & \AblBase{\AblYes} & \AblBase{\AblYes} & \AblScoreInline{\textbf{62.99}}{\textcolor{TableGain}{(+13.68)}} \\
\arrayrulecolor{TableRule}\specialrule{1pt}{0pt}{0pt}
\end{tabular}
}
\endgroup
\end{table}

\begin{figure*}[t]
\centering
\includegraphics[width=\textwidth]{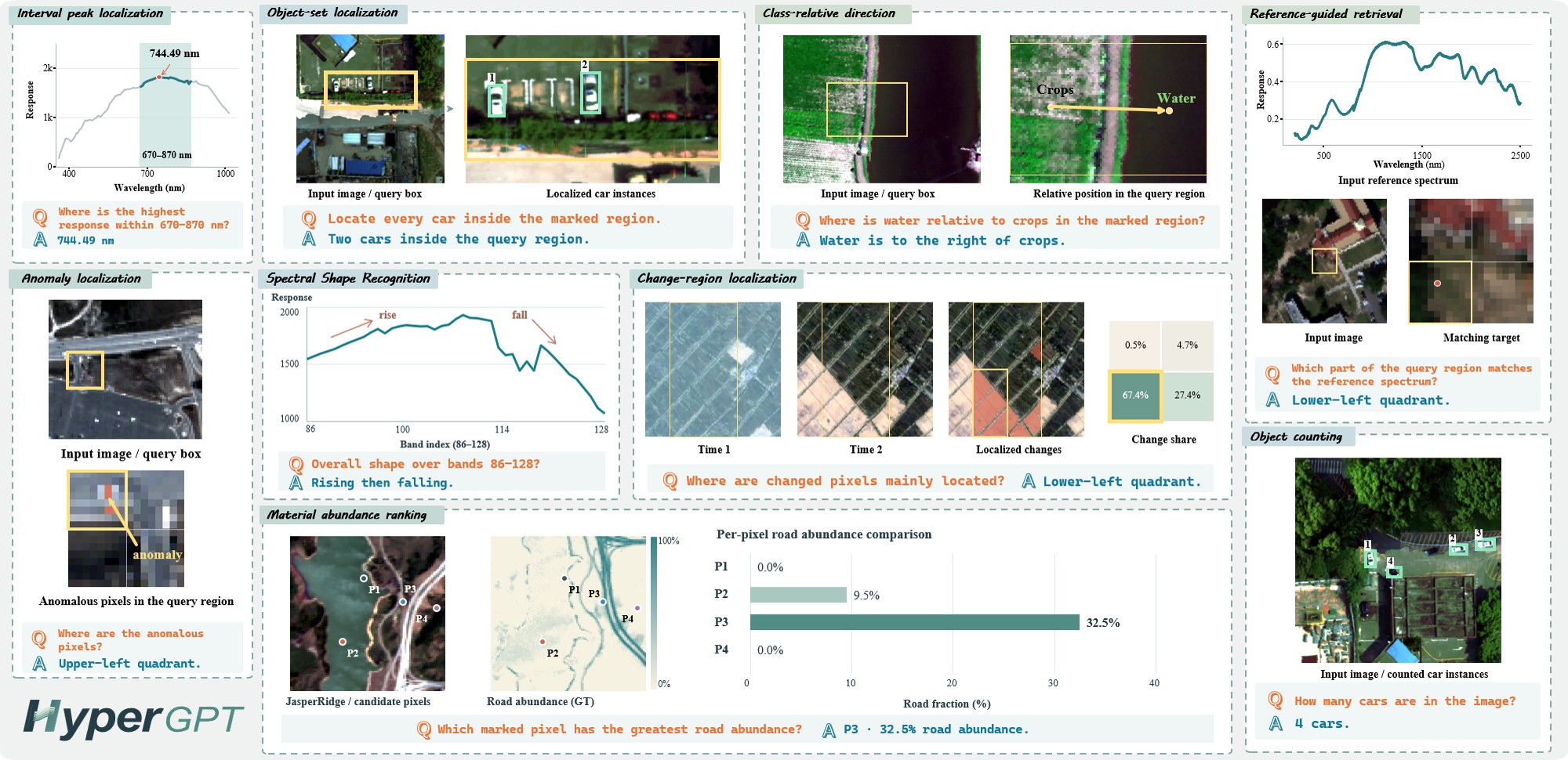}
\caption{Qualitative results of HyperGPT on representative HSI understanding tasks.}
\label{fig:qualitative_analysis}
\end{figure*}

\begin{figure}[t]
\centering
\includegraphics[width=\columnwidth]{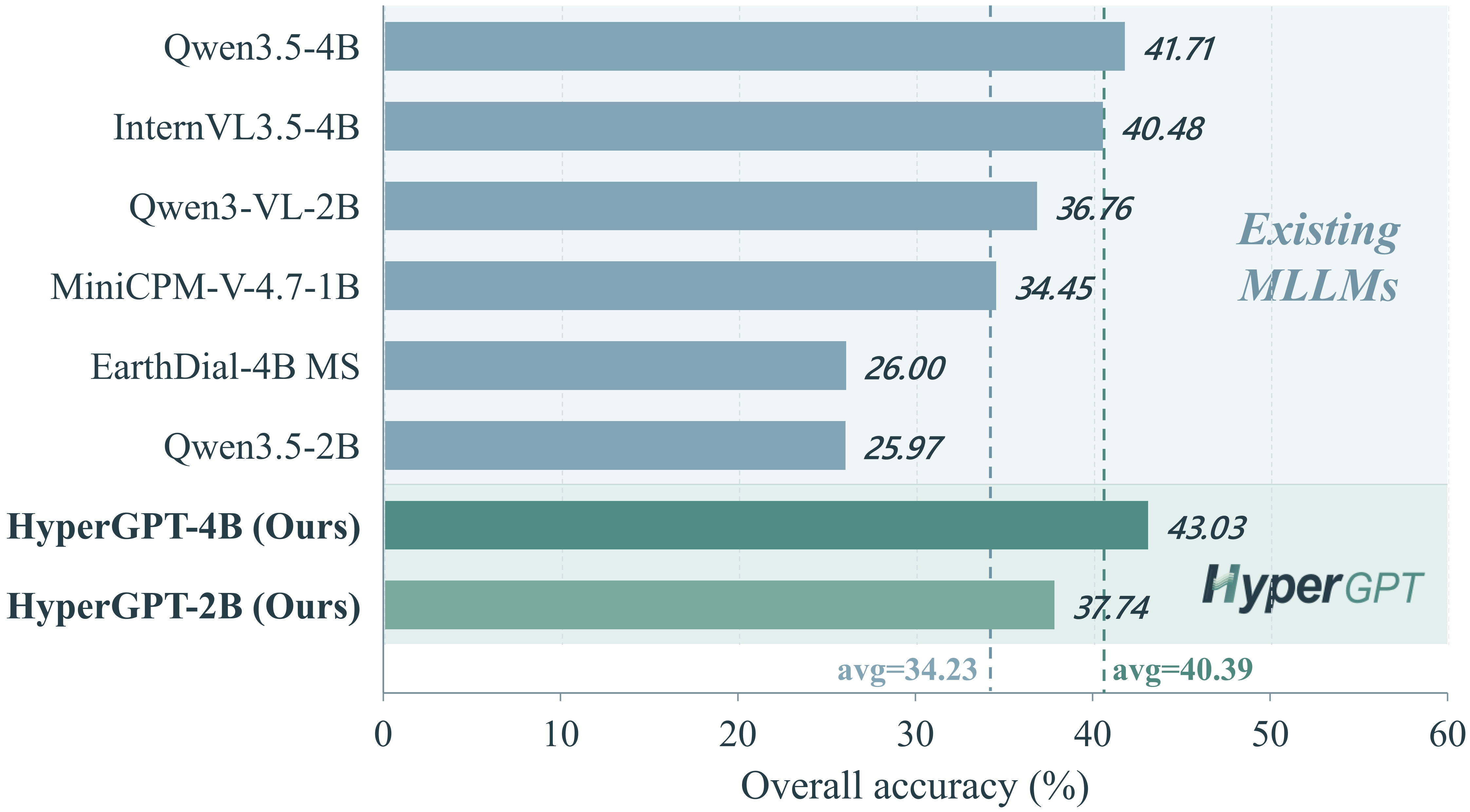}
\caption{Overall choice accuracy (\%) results on HM-Bench.}
\label{fig:hm_bench}
\end{figure}

\subsection{Cross-Benchmark Evaluation on HM-Bench}\label{sec:hm_bench}

To evaluate zero-shot generalization, we further assess HyperGPT against competitive methods from Table \ref{tab:main_comparison} on the external HM-Bench benchmark \cite{hmbench}, which differs from HyperBench in task definitions, image patch scales, and evidence organization. As shown in Fig.~\ref{fig:hm_bench}, HyperGPT-4B achieves the highest overall accuracy of 43.03\%, correctly answering approximately 255 and 493 more questions than Qwen3.5-4B and InternVL3.5-4B, respectively. Notably, HyperGPT-2B achieves 37.74\% accuracy, surpassing its pretrained backbone Qwen3.5-2B by 11.77 percentage points, as well as Qwen3-VL-2B and MiniCPM-V-4.7-1B by 0.98 and 3.29 percentage points, respectively. These results demonstrate the strong generalization of HyperGPT to unseen HSI datasets.

\subsection{Qualitative Analysis of HSI Understanding}\label{sec:qualitative_analysis}

Representative tasks from the seven major task families are visualized in Fig.~\ref{fig:qualitative_analysis} to illustrate HyperGPT’s HSI understanding capabilities. In the spectral analysis examples, HyperGPT locates the maximum response within a specified wavelength interval and recognizes the overall shape of a spectral curve despite local fluctuations. In the spatial understanding examples, it distinguishes anomalous pixels from the background, localizes and counts target objects, and determines the relative direction between land-cover classes. It also identifies regions where changes occur between paired observations. Given a reference spectrum, HyperGPT identifies the matching region within the query area. For material composition analysis, it ranks candidate locations by the abundance of the target material in mixed spectra. These examples illustrate how HyperGPT uses spectral and spatial evidence to answer questions under different semantic conditions. Additional visualizations are provided in the Appendix.

\section{Conclusion}\label{sec:conclusion}

We present HyperInstruct, the first QA training dataset for hyperspectral image understanding, together with HyperBench, the hyperspectral QA benchmark with the most comprehensive task coverage to date. They provide 120,918 QA pairs and cover 7 task categories and 34 fine-grained task groups, enabling models to learn and evaluate diverse HSI understanding capabilities within a unified question answering framework. Building upon these resources, we introduce HyperGPT, the first multimodal large language model designed for hyperspectral image understanding, to address the challenges of modeling hyperspectral observations for natural-language question answering. Comparative experiments, ablation studies and zero-shot evaluations demonstrate the effectiveness of HyperGPT from multiple perspectives. Based on these experimental findings, we draw two main conclusions: (1) HyperInstruct and HyperBench provide complementary resources for acquiring and evaluating domain-specific knowledge of hyperspectral images, supporting multimodal large language models in HSI understanding. (2) By jointly modeling spectral-spatial representations, HyperGPT learns hyperspectral-specific features and generally outperforms existing multimodal large language models across diverse hyperspectral understanding tasks. Overall, our approach provide a strong data and modeling foundation for natural-language question answering with hyperspectral obserbatioins.

{\small
\bibliographystyle{ieeenat_fullname}
\bibliography{references}
}
\clearpage
\appendix
\input{appendix/appendix}

\end{document}

%% file: appendix/appendix.tex
\makeatletter
\renewcommand{\paragraph}{%
  \@startsection{paragraph}{4}{\z@}%
    {1.0ex \@plus .2ex \@minus .1ex}%
    {-1em}%
    {\normalfont\normalsize\bfseries}%
}
\makeatother

\newcolumntype{Y}{>{\centering\arraybackslash}X}
\setcounter{figure}{0}
\setcounter{table}{0}
\renewcommand{\thefigure}{S\arabic{figure}}
\renewcommand{\thetable}{S\arabic{table}}
\renewcommand{\theHfigure}{supp.\arabic{figure}}
\renewcommand{\theHtable}{supp.\arabic{table}}

\maketitlesupplementary

\section{Detailed Dataset Construction}\label{app:sec:dataset}

\subsection{Source Dataset} \label{app:sec:source_dataset_coverage}

We collect hyperspectral data from 38 source datasets ~\cite{quh,whuhi,szutree,harvard,hyperspectralcity,cave,chikusei,muufl,houston2013,houston2018,xiongan,indianpines,szucar,hod3k,spod,change_original,change_data,abu,sandiego,unmixing_data,gff,xuzhou,washingtondc} and establish a unified data inventory and loading protocol to standardize these heterogeneous data. The standardized data encompass hyperspectral image cubes, wavelength metadata, spatial coordinates, class or object annotations, temporal correspondences, and auxiliary information such as reference spectra and endmembers, providing the foundation for subsequent question-answer pair construction. We introduce the source datasets used to construct HyperInstruct and HyperBench below. Where verified, band counts and image dimensions are reported based on local data files, while approximate spectral ranges indicate nominal wavelength coverage.

\noindent\textbf{Houston 2013.}
The Houston 2013 GRSS benchmark uses CASI hyperspectral observations of an urban
area with 144 bands covering approximately 364--1046\,nm and about 2.5\,m
spatial resolution. The project uses it for semantic-spatial evaluation.\par

\noindent\textbf{Houston 2018.}
The Houston 2018 multimodal classification benchmark provides urban land-cover
observations from CASI. The project file contains $601\times2384\times50$
spectral data, and this 50-band version supplies semantic-spatial training
questions.\par

\noindent\textbf{Botswana.}
Botswana is an EO-1 Hyperion wetland scene with vegetation, water, and wetland
classes. The project retains 145 bands from the approximately 400--2500\,nm
sensor range and uses the $1476\times256$ scene for spectral and semantic
evaluation.\par

\noindent\textbf{Indian Pines.}
Indian Pines is an AVIRIS agricultural scene with 16 conventional land-cover
classes. The project uses a $149\times149$ input with 200 retained bands and
tests crop-related spectral discrimination and curve understanding.\par

\noindent\textbf{KSC.}
KSC is an AVIRIS wetland and vegetation scene. The project uses the
$512\times614$ processing version with 176 effective bands for spectral and
spatial evaluation.\par

\noindent\textbf{Pavia Centre.}
Pavia Centre is a ROSIS urban scene containing roads, buildings, vegetation, and
water. The project uses $1096\times715$ pixels and 102 bands from the
approximately 430--860\,nm range for semantic-spatial training.\par

\noindent\textbf{Pavia University.}
Pavia University is a ROSIS urban scene with 103 bands, $610\times340$ pixels,
and approximately 430--860\,nm coverage. It provides the held-out urban
semantic-spatial scene.\par

\noindent\textbf{Salinas.}
Salinas is an AVIRIS agricultural scene with approximately 3.7\,m spatial
resolution. The project uses 204 retained bands from a $512\times217$ image and
constructs semantic questions from three explicit training windows.\par

\noindent\textbf{WHU-Hi HanChuan.}
WHU-Hi HanChuan is a UAV Headwall Nano-Hyperspec agricultural scene with
$1217\times303$ pixels, 274 bands, approximately 400--1000\,nm coverage, and
about 0.109\,m spatial resolution. It provides training annotations.\par

\noindent\textbf{WHU-Hi HongHu.}
WHU-Hi HongHu is a UAV Headwall Nano-Hyperspec agricultural scene with
$940\times475$ pixels, 270 bands, 400--1000\,nm coverage, and approximately
0.043\,m spatial resolution. It provides training annotations.\par

\noindent\textbf{WHU-Hi LongKou.}
WHU-Hi LongKou is a UAV agricultural scene with $550\times400$ pixels, 270
bands, approximately 400--1000\,nm coverage, and nine labeled classes. It is
used for agricultural semantic-spatial evaluation.\par

\noindent\textbf{QUH-Pingan.}
QUH-Pingan is a UAV hyperspectral scene containing water, roads, trees,
buildings, ships, and vehicles. The project file has $1230\times1000$ pixels and
176 bands over approximately 400--1000\,nm, and supplies training questions.\par

\noindent\textbf{QUH-Qingyun.}
QUH-Qingyun is a 176-band urban scene with $880\times1360$ pixels and
approximately 400--1000\,nm coverage. It supplies evaluation questions under
shadow and illumination variation.\par

\noindent\textbf{QUH-Tangdaowan.}
QUH-Tangdaowan is a 176-band UAV scene with $1740\times860$ pixels. Its
fine-scale vegetation and pavement categories provide complementary spectral and
spatial training evidence.\par

\noindent\textbf{GFF.}
GFF is a labeled hyperspectral source with a $572\times906$ image and 125 bands
in the project files. It contributes semantic-spatial evaluation questions.
The original release and acquisition metadata remain unverified.

\noindent\textbf{SZUTree R1.}
SZUTree R1 provides tree-species and surface labels with a
$2405\times3085$ image and 98 bands. The inventory records an approximate
399--924\,nm range and an annotation-resolution alignment step. It is used for
evaluation.
 
\noindent\textbf{SZUTree R2.}
SZUTree R2 is the companion tree-species region with a $2444\times4040$ image
and 98 bands. It provides training questions, while R1 evaluates transfer
across the two regions.

\noindent\textbf{Xiongan.}
Xiongan covers vegetation and agricultural land around Matu Bay in Xiong'an.
The project input is $1580\times3750\times256$, while the official description
mentions 250 bands and approximately 400--1000\,nm coverage. The benchmark
retains the local 256-band version for training.\par

\noindent\textbf{Xuzhou.}
The Xuzhou source used here is a HySpex version with $500\times260$ pixels and
436 bands. Its registered coverage is approximately 415--2508\,nm, and it is
kept separate from other Xuzhou versions with different sensors or band counts.

\noindent\textbf{Chikusei.}
Chikusei was acquired by a Headwall Hyperspec-VNIR-C in a Japanese agricultural
and urban region. The project uses $2517\times2335$ pixels, 128 bands,
approximately 363--1018\,nm coverage, and 2.5\,m ground sampling distance.\par

\noindent\textbf{HyperspectralCity V2.}
HyperspectralCity V2 provides 128-channel urban road scenes and 19 semantic
classes. The project uses approximately 450--950\,nm coverage and separate
official train and test file lists. The validation files are a coarser-label
subset of test and are not treated as an independent split.\par

\noindent\textbf{Washington DC Mall.}
Washington DC Mall is a HYDICE urban scene. The project uses a
$1280\times307$ image with 191 registered bands spanning approximately
401.3--2473.2\,nm, mainly for point-spectrum curve evaluation.\par

\noindent\textbf{Harvard Hyperspectral.}
The Harvard collection contains natural indoor and outdoor scenes. The project
uses 77 MAT files with $1040\times1392$ pixels and 31 bands over 420--720\,nm at
10\,nm intervals, divided into 60 training files and 17 evaluation files.\par

\noindent\textbf{CAVE Multispectral.}
CAVE contains laboratory scenes of objects, materials, food, and real/fake
object pairs. The project uses 32 files with $512\times512$ pixels and 31 bands
over 400--700\,nm at 10\,nm intervals, divided into 24 training files and eight
evaluation files.\par

\noindent\textbf{Airport--Beach--Urban (ABU).}
ABU contributes 13 anomaly-detection scenes with binary anomaly masks. The
project uses airport1--3, beach1--3, and urban1--3 for training and airport4,
beach4, and urban4--5 for evaluation. Band counts vary from 102 to 207 across
scenes, so the scenes do not share one forced wavelength axis.\par

\noindent\textbf{HYDICE Urban.}
The anomaly-detection version of HYDICE Urban is an $80\times100$ crop with
175 retained bands and vehicle anomaly masks. It is distinct from the 162-band
Urban unmixing version and is used only for evaluation.\par

\noindent\textbf{Bay Area.}
Bay Area provides AVIRIS bi-temporal observations from 2013 and 2015 with
$600\times500$ pixels and 224 bands. The complete pair supplies change-detection
training questions, with unlabeled pixels excluded from change fractions.\par

\noindent\textbf{Santa Barbara.}
Santa Barbara contains registered AVIRIS observations from 2013 and 2014 with
$984\times740$ pixels and 224 bands. It supplies bi-temporal change-detection
training questions.\par

\noindent\textbf{Hermiston.}
Hermiston is an EO-1 Hyperion agricultural change scene. The project uses the
$307\times241$ local processing version with 154 retained bands and a complete
bi-temporal pair for training.\par

\noindent\textbf{Farmland.}
Farmland is an EO-1 Hyperion agricultural change scene with a
$450\times140$ processing version, 155 retained bands, and change ground truth.
It is reserved for evaluation of temporal presence, localization, area, spectral
difference, and connected-region questions.\par

\noindent\textbf{MUUFL Gulfport.}
MUUFL Gulfport provides campus hyperspectral observations, target locations,
and laboratory reference spectra. The project uses 72 bands over approximately
367.7--1043.4\,nm, with flights 1 and 3 for training and flight 4 for
evaluation.\par

\noindent\textbf{San Diego.}
San Diego contains aircraft targets in a $100\times100$ AVIRIS parent image with
189 retained bands. Training uses rows $[0,50)$ and columns $[0,100)$, while
evaluation uses rows $[50,100)$ and columns $[0,50)$ after conversion to local
coordinates.\par

\noindent\textbf{SZU-CAR.}
SZU-CAR is a UAV Specim FX10 vehicle-detection source with 605
$640\times640$ images, 112 bands, and approximately 400--1000\,nm coverage at
0.1\,m spatial resolution. The source train and validation roles are used for
the project training and evaluation sides.\par

\noindent\textbf{HOD3K.}
HOD3K is a snapshot-camera hyperspectral object-detection source. The project
uses the local $256\times512\times16$ decoding of a $4\times4$ mosaic with people, bike,
worker, and car labels, while the original paper describes three classes. This
version distinction is retained in the benchmark description.\par

\noindent\textbf{SPOD.}
SPOD contains synthetic point and subpixel targets with target mixtures,
abundances, and spatial annotations. The project uses the 30-band experimental
version rather than the original 150-band release and therefore treats its
spectral axis as band indices without assigning uncalibrated physical
wavelengths.\par

\noindent\textbf{Urban.}
Urban is a HYDICE city scene used for spectral unmixing. The project uses a
$307\times307$ cube with 162 retained bands and four-, five-, and six-endmember
reference configurations. These configurations come from one cube and are
counted as one standardized source.\par

\noindent\textbf{Samson.}
Samson is a 95$\times$95 scene with 156 bands over approximately 401--889\,nm.
Its soil, tree, and water reference materials support abundance, composition,
dominance, and candidate-comparison evaluation questions.\par

\noindent\textbf{Jasper Ridge.}
Jasper Ridge is an AVIRIS unmixing scene with $100\times100$ pixels and 198
effective bands. The project uses road, soil, water, and tree reference
materials with associated endmember and abundance data for evaluation.\par

\paragraph{Scope and version notes.}
The source count follows the project normalization rules. The 13 ABU scenes are
grouped as one source, and the four-, five-, and six-endmember Urban
configurations are grouped as one source. The project retains local-version discrepancies 
such as the 50-band Houston 2018 file, the 256-band Xiongan file, the 149-by-149 Indian
Pines file, and the four-class HOD3K annotation version. Physical wavelength
metadata have different confidence levels across sources, so approximate
coverage ranges are reported as context rather than as exact per-band
calibration.

\subsection{Dataset Composition and Split}

From the 38 source datasets, we collect 4,754 hyperspectral images spanning diverse sensor types, spatial resolutions, and spectral ranges. To ensure a sufficiently large number of question--answer pairs while limiting the storage overhead of hyperspectral imagery, we retain the original images rather than physically cropping them into smaller patches. Instead, we define spatial windows within each image for question--answer pair generation, allowing multiple pairs to be derived from a single image, with each pair associated with a specific spatial window. This window-based strategy also enables explicit spatial separation between training and evaluation samples, preventing spatial overlap between their corresponding regions.

The resulting HyperInstruct comprises 100,918 training question--answer pairs from 22 source datasets, while the final HyperBench evaluation set contains 20,000 questions from 25 source datasets. Among the 38 source datasets, 13 contribute exclusively to training, 16 exclusively to evaluation, and 9 to both. Table~\ref*{app:tab:source_dataset_split} details the question counts and data partitioning rules for each source dataset.

\begin{table*}[!tp]
\centering
\captionsetup{skip=4pt}
\caption{Source-level allocation for HyperInstruct and HyperBench}
\label{app:tab:source_dataset_split}
\begingroup
\color{TableInk}
\setlength{\tabcolsep}{4pt}
\renewcommand{\arraystretch}{1.06}
\small
\begin{tabularx}{\textwidth}{@{}>{\raggedright\arraybackslash}p{4.05cm}
>{\centering\arraybackslash}p{1.45cm}
>{\centering\arraybackslash}p{1.45cm}
>{\centering\arraybackslash}p{1.45cm}
>{\raggedright\arraybackslash}X@{}}
\arrayrulecolor{TableRule}\specialrule{1pt}{0pt}{0pt}
\rowcolor{TableHeader}
\textbf{Source dataset} & \textbf{Scenes} & \textbf{Train QA} & \textbf{Test QA} &
\textbf{Split rule}\\
\arrayrulecolor{TableSeparator}\specialrule{.45pt}{0pt}{0pt}
Houston 2013 & 1 & 0 & 573 & Evaluation scene.\\
Houston 2018 & 1 & 1,771 & 0 & Training scene only.\\
Botswana & 1 & 0 & 585 & Evaluation scene only.\\
Indian Pines & 1 & 0 & 644 & Evaluation scene only.\\
KSC & 1 & 0 & 465 & Evaluation scene only.\\
Pavia Centre & 1 & 1,776 & 0 & Training scene.\\
Pavia University & 1 & 0 & 655 & Evaluation scene only.\\
Salinas & 3 & 1,243 & 0 & Training questions use three windows.\\
WHU-Hi HanChuan & 1 & 1,065 & 0 & Training scene.\\
WHU-Hi HongHu & 1 & 709 & 0 & Training scene.\\
WHU-Hi LongKou & 1 & 0 & 703 & Evaluation scene.\\
QUH-Pingan & 1 & 3,640 & 0 & Training scene.\\
QUH-Qingyun & 1 & 0 & 1,025 & Evaluation scene.\\
QUH-Tangdaowan & 1 & 3,115 & 0 & Training scene.\\
GFF & 1 & 0 & 143 & Evaluation scene only.\\
SZUTree R1 & 1 & 0 & 1,937 & Evaluation region.\\
SZUTree R2 & 1 & 6,439 & 0 & Training region.\\
Xiongan & 1 & 356 & 0 & Training source only.\\
Xuzhou & 1 & 0 & 438 & Evaluation source only.\\
Chikusei & 1 & 0 & 1,103 & Evaluation scene only.\\
HyperspectralCity V2 & 1,306 & 18,289 & 2,742 & Official train and test lists (1,030 and 276 scenes).\\
Washington DC Mall & 1 & 0 & 229 & Evaluation scene only.\\
Harvard Hyperspectral & 77 & 2,919 & 221 & 60 files for training and 17 files for evaluation.\\
CAVE Multispectral & 32 & 2,842 & 237 & 24 files for training and 8 files for evaluation.\\
Airport--Beach--Urban & 13 & 12,000 & 1,631 & Train on Airport1--3, Beach1--3, Urban1--3. Test on Airport4, Beach4, Urban4--5.\\
HYDICE Urban & 1 & 0 & 537 & Evaluation only (175-band anomaly-detection version).\\
Bay Area & 1 & 2,670 & 0 & Complete bi-temporal pair for training.\\
Santa Barbara & 1 & 2,666 & 0 & Complete bi-temporal pair for training.\\
Hermiston & 2 & 2,664 & 0 & Complete local bi-temporal processing version for training.\\
Farmland & 2 & 0 & 832 & Complete bi-temporal pair for evaluation.\\
MUUFL Gulfport & 3 & 8,182 & 1,053 & Flights 1 and 3 for training. Flight 4 for evaluation.\\
San Diego & 2 & 1,386 & 123 & Upper parent-image crop train; lower-left crop test.\\
SZU-CAR & 596 & 7,182 & 979 & Source training list for training and validation list for evaluation.\\
HOD3K & 2,320 & 6,003 & 1,272 & Official source train/test split.\\
SPOD & 372 & 2,001 & 374 & 30-band source train/test version.\\
Urban & 1 & 12,000 & 0 & Training scene only.\\
Samson & 1 & 0 & 604 & Evaluation scene only.\\
Jasper Ridge & 1 & 0 & 895 & Evaluation scene only.\\
\arrayrulecolor{TableRule}\specialrule{1pt}{0pt}{0pt}
\rowcolor{TableHeader}
\textbf{Total (38 entries)} & \textbf{4,754} & \textbf{100,918} & \textbf{20,000} & \\
\arrayrulecolor{TableRule}\specialrule{1pt}{0pt}{0pt}
\end{tabularx}
\endgroup
\end{table*}

\subsection{Task Design}\label{app:sec:dataset_comparison}

Based on questions whose answers can be directly determined from these data and annotations, we organize 34 fine-grained question-answering task groups into seven categories: spectral curve analysis, spatial semantic understanding, object instance analysis, change detection, anomaly detection, reference spectrum retrieval, and spectral unmixing analysis. 

\subsubsection{Spectral Curve Analysis}
This group focuses on interpreting spectral responses along the wavelength axis, from individual extrema to interval-level trends and overall curve shapes. It assesses whether a model can associate response values with spectral positions and distinguish local patterns from global behavior.
\begin{itemize}
\item \noindent\textbf{Extremum localization.} Given a pixel spectrum and a wavelength or band interval, the model identifies the location of its maximum or minimum response. This tests whether it can relate spectral values to the requested spectral coordinate. \par
\item \noindent\textbf{Trend recognition.} The model determines whether a spectrum rises, falls, remains approximately flat, or changes direction over a specified interval. It must characterize the interval-level trend while accounting for local fluctuations. \par
\item \noindent\textbf{Shape recognition.} The model selects the overall shape of a pixel spectrum over a specified range, such as rising then falling or falling then rising. This evaluates global curve interpretation beyond a single extremum or endpoint comparison. \par
    \item \noindent\textbf{Extrema counting.} The model counts strict local maxima or minima within a spectral interval under explicit endpoint rules. This tests detection of local spectral structure and consistent application of the counting definition. \par
 
\end{itemize}

\subsubsection{Spatial Semantic Understanding}
This group examines the semantic content and spatial organization of pixels and regions in hyperspectral images. The questions assess whether a model can use spectral evidence and spatial context to identify classes, compare regions, and reason about relationships between land-cover types.
\begin{itemize}
\item \noindent\textbf{Pixel classification.} The model assigns a land-cover or material class to a queried pixel. It must use the pixel's spectral response and its spatial context as evidence for the class decision. \par
\item \noindent\textbf{Class-guided pixel retrieval.} Given a target class and several candidate coordinates, the model selects a pixel belonging to that class. This combines class recognition with spatially grounded candidate selection. \par
\item \noindent\textbf{Same-class matching.} Given a reference pixel and candidate pixels, the model identifies the candidate that shares the reference class. This tests comparative spectral-semantic matching across spatial locations. \par
\item \noindent\textbf{Support-guided classification.} The model classifies a query pixel using labeled support pixels supplied in the question. This evaluates few-shot transfer from spectral examples to an unseen query location. \par
\item \noindent\textbf{Regional composition.} The model identifies the dominant class, estimates class prevalence, or rejects a single-class description for a queried region. This tests whether it can summarize heterogeneous label distributions rather than treating a mixed region as uniform. \par
\item \noindent\textbf{Class spatial relations.} Given two classes within a region, the model determines their relative spatial direction. It must combine class-conditioned localization with coordinate-based relational reasoning. \par
\item \noindent\textbf{Class-guided region retrieval.} Given a target class and candidate regions, the model selects a region that contains the class. This tests semantic grounding over multiple spatial extents. \par
\item \noindent\textbf{Regional count comparison.} The model compares candidate subregions by the number of pixels assigned to a specified class. This evaluates localized counting and comparison over spatial partitions. \par
\end{itemize}

\subsubsection{Object Instance Analysis}
This group focuses on discrete object instances and their spatial annotations. It evaluates whether a model can count and localize objects, identify sets of instances, and reason about their spatial relationships at the image or region level.
\begin{itemize}
\item \noindent\textbf{Global counting.} The model counts instances of a specified object category across the full image. This tests image-level object recognition and enumeration. \par
\item \noindent\textbf{Regional counting.} The model counts objects whose centers fall within a specified region. It must apply the region's coordinate bounds consistently when assigning instances. \par
\item \noindent\textbf{Single-object localization.} Given a target category and candidate boxes, the model selects the box that encloses the queried instance. This evaluates category-conditioned spatial localization. \par
\item \noindent\textbf{Object-set localization.} The model selects a candidate box set that covers all target instances in a region. This requires joint recognition, instance enumeration, and set-level localization. \par
\item \noindent\textbf{Object spatial relations.} Given a marked instance or two target categories, the model determines the relative direction between objects. This tests instance-level grounding and spatial relation reasoning. \par
\end{itemize}

\subsubsection{Change Detection}
This group evaluates comparisons between co-registered hyperspectral observations acquired at different times. The questions assess whether a model can interpret temporal differences in both spatial extent and spectral response while respecting the specified query region.
\begin{itemize}
\item \noindent\textbf{Change presence.} The model determines whether labeled pixels changed between two observations within a queried region. This evaluates temporal comparison under the provided spatial scope. \par
\item \noindent\textbf{Change localization.} The model identifies the part of a region where changed pixels are concentrated. It must ground temporal differences to the corresponding image coordinates. \par
\item \noindent\textbf{Changed-area fraction.} The model estimates the proportion of valid labeled pixels that changed within a region and selects the appropriate range. This tests quantitative interpretation of spatial change extent. \par
\item \noindent\textbf{Spectral change localization.} The model selects the band interval with the largest response difference between two observations. This requires comparing temporal spectra while preserving their band correspondence. \par
\item  \noindent\textbf{Change-region counting.} The model counts connected changed components under a specified neighborhood rule and minimum-size threshold. This tests spatial topology reasoning over a binary change pattern. \par

\end{itemize}

\subsubsection{Anomaly Detection}
This group focuses on identifying anomalous pixels that differ from their surrounding background in spectral or spatial evidence. It evaluates anomaly recognition, localization, extent estimation, and comparison of spectral responses between anomalous and background regions.
\begin{itemize}
\item \noindent\textbf{Anomaly presence.} The model determines whether anomalous pixels occur in a queried region. This evaluates discrimination between anomalous responses and the surrounding background. \par
\item \noindent\textbf{Anomaly localization.} Given candidate subregions or boxes, the model identifies the one containing anomalous pixels. It must connect anomaly recognition to fine-grained spatial localization. \par
\item \noindent\textbf{Anomalous-area fraction.} The model estimates the fraction of a region occupied by anomalous pixels. This tests quantitative interpretation of anomaly extent. \par
\item \noindent\textbf{Spectral anomaly contrast.} The model selects the spectral interval with the largest response difference between anomaly and nearby background. This evaluates comparative spectral analysis conditioned on spatially identified regions. \par
\end{itemize}

\subsubsection{Reference Spectrum Retrieval}
This group evaluates whether a model can use a provided reference spectrum to search hyperspectral observations for a matching target. It tests spectral-reference matching together with spatial grounding, including determining whether a target is present and where it occurs.
\begin{itemize}
\item \noindent\textbf{Target presence.} Given a reference spectrum, the model determines whether a matching target occurs within a specified region. This tests comparison between a spectral reference and image observations. \par
\item \noindent\textbf{Target localization.} Given a reference spectrum and candidate regions, the model identifies the region containing the matching target. This combines spectral matching with spatial retrieval. \par
\end{itemize}

\subsubsection{Spectral Unmixing Analysis}
This group focuses on interpreting mixed pixels through the abundances of their constituent materials. It assesses whether a model can estimate and compare material proportions, identify dominant components, and reason about material composition under specified thresholds.
\begin{itemize}
\item \noindent\textbf{Abundance estimation.} The model selects the abundance interval for a material in a pixel or region. This evaluates interpretation of continuous material proportions from mixed spectra. \par
\item \noindent\textbf{Abundance difference.} Given two locations and a reference material, the model determines the direction and magnitude range of their abundance difference. This tests quantitative comparison of material proportions. \par
\item \noindent\textbf{Abundance ranking.} The model ranks candidate pixels or regions by the abundance of a specified material. It compare material-specific proportions across locations. \par
\item \noindent\textbf{Thresholded composition.} Given a minimum abundance threshold, the model identifies the materials that satisfy it in a pixel or region. This tests multi-material set reasoning from abundance values. \par
\item \noindent\textbf{Dominant material.} The model identifies the material with the largest abundance at a queried location or region. This evaluates selection of the leading component in a mixed spectrum. \par
\item \noindent\textbf{Dominance-based retrieval.} Given a material and candidate locations, the model finds where that material is dominant among the components. This combines abundance comparison with spatial retrieval. \par
\end{itemize}

\subsection{Question-Answer Generation}\label{app:sec:qa_generation}
\begin{figure*}[t]
\centering
\includegraphics[width=\textwidth]{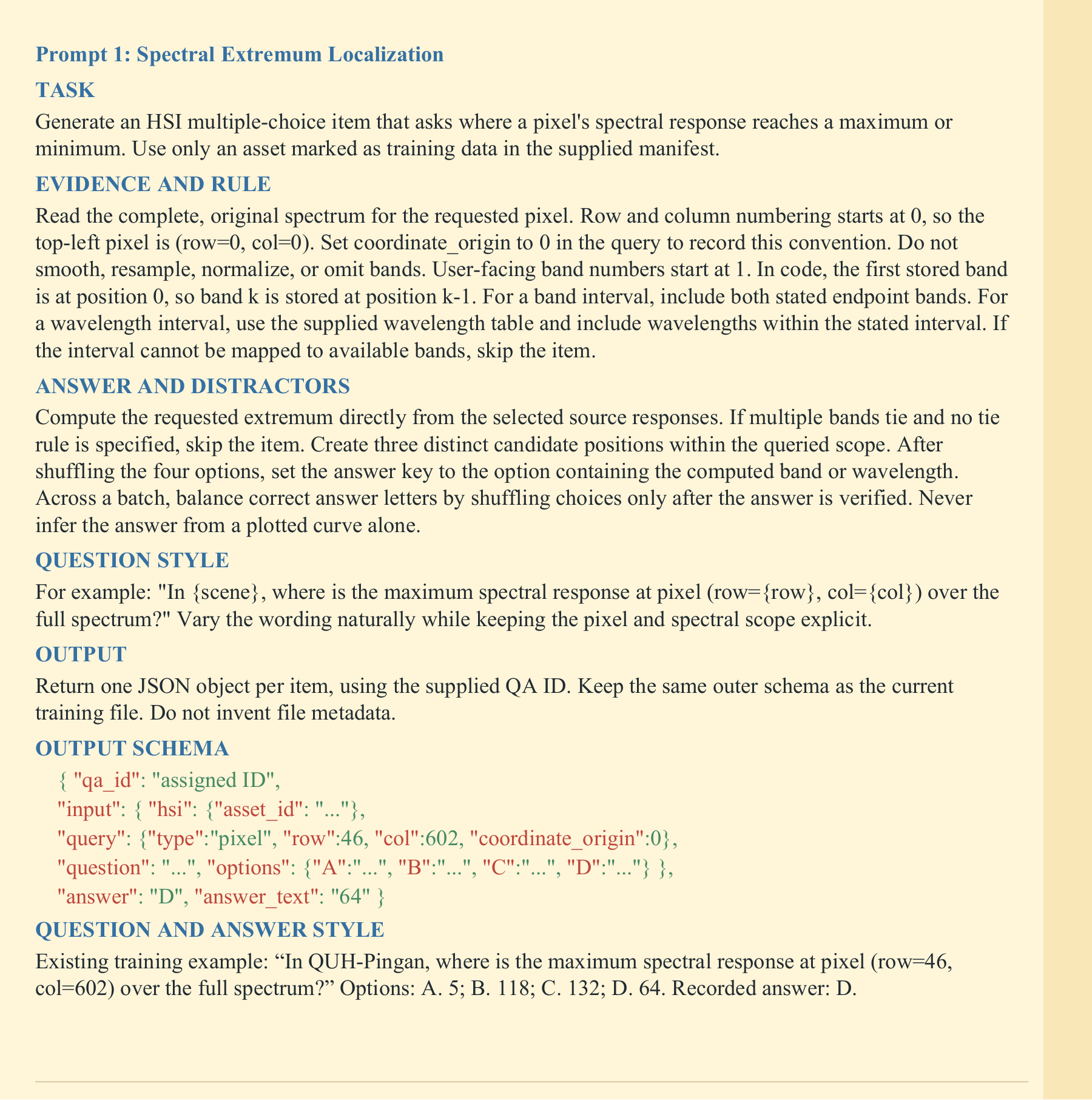}
\caption{Structured prompt for spectral extremum localization.}
\label{app:fig:01_extremum_localization}
\end{figure*}

\begin{figure*}[t]
\centering
\includegraphics[width=\textwidth]{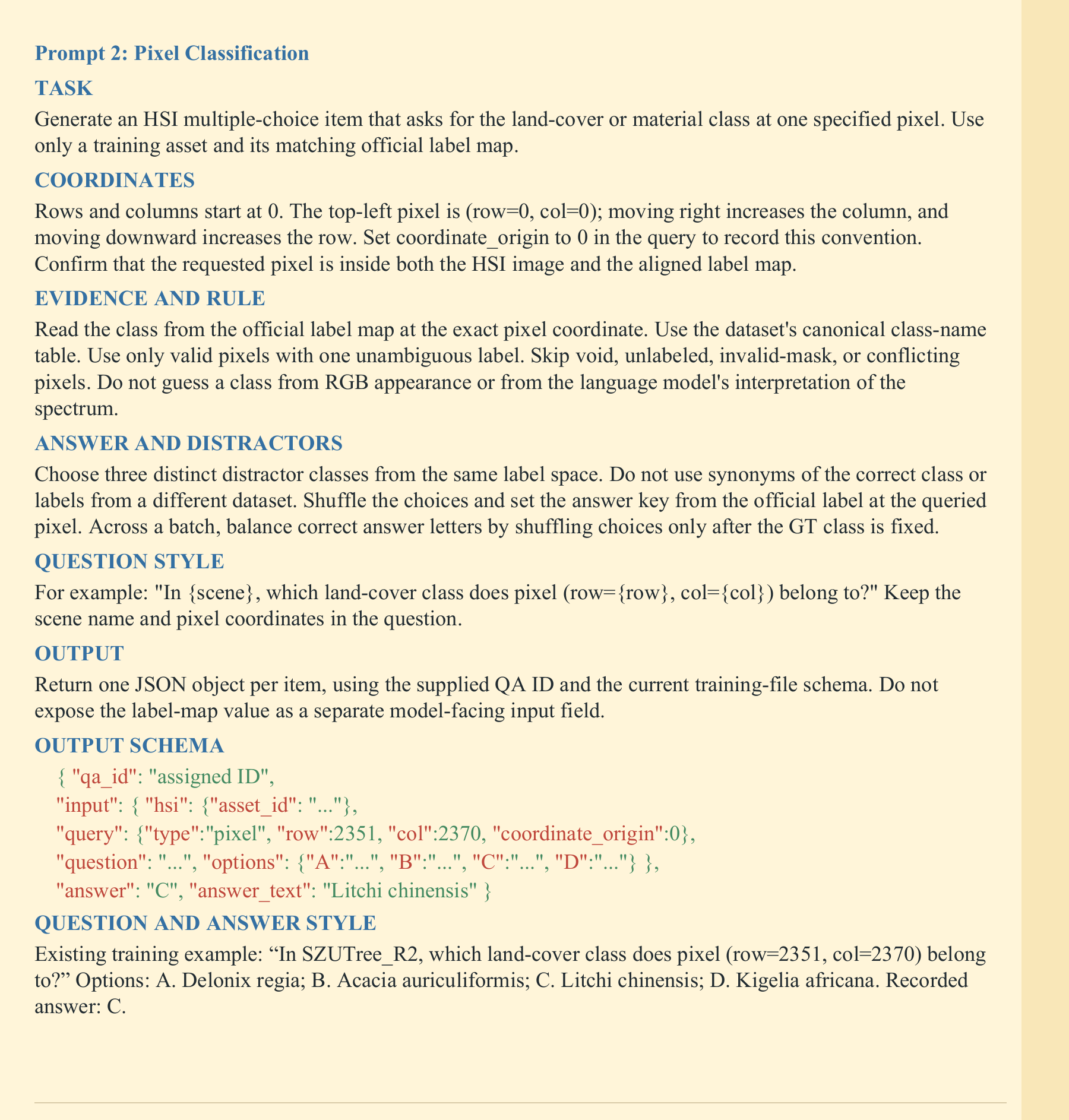}
\caption{Structured prompt for HSI pixel classification.}
\label{app:fig:02_pixel_classification}
\end{figure*}

\begin{figure*}[t]
\centering
\includegraphics[width=\textwidth]{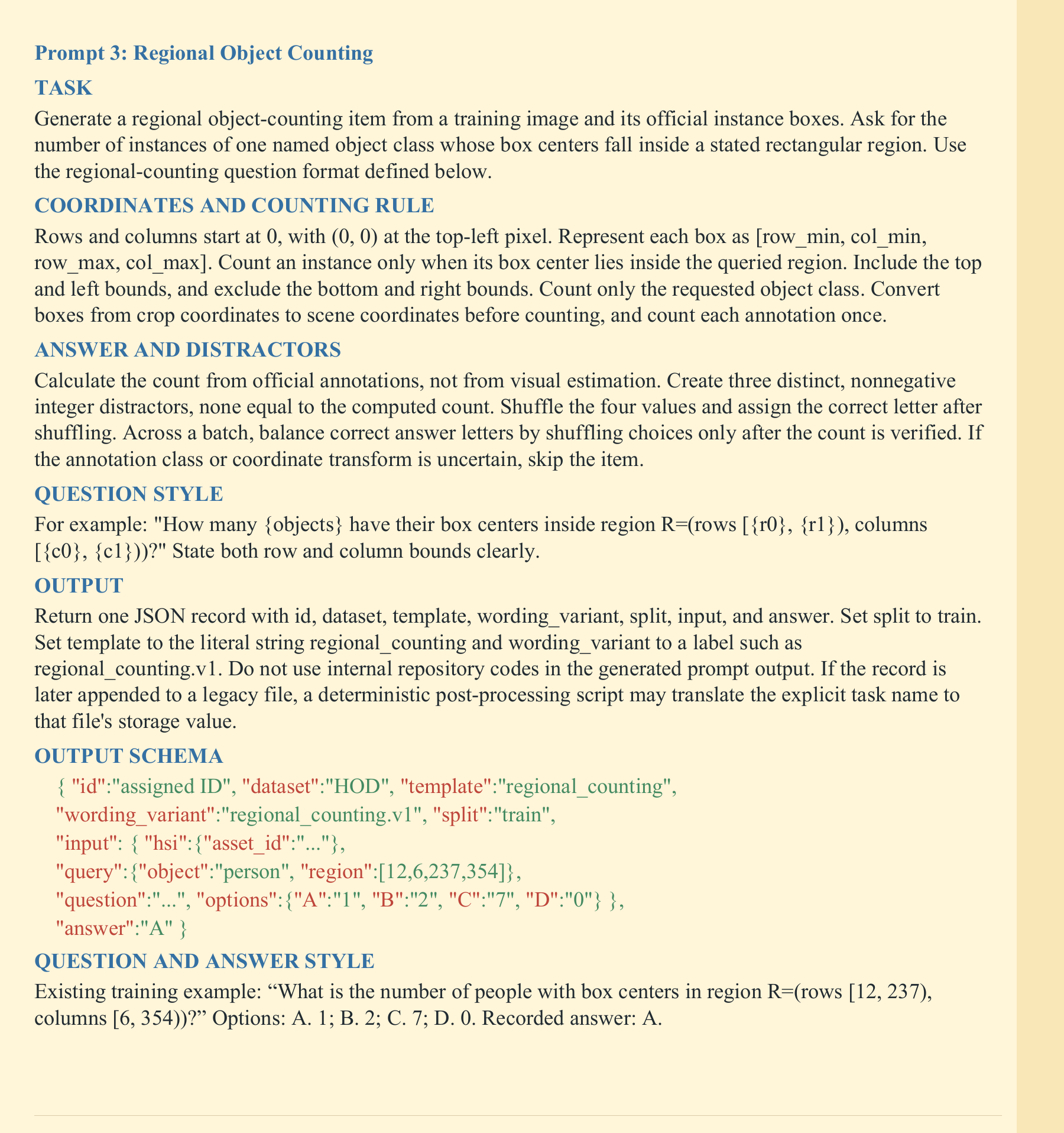}
\caption{Structured prompt for regional object counting using annotated box centers.}
\label{app:fig:03_regional_object_counting}
\end{figure*}

For each fine-grained task, we specify the question objective, observation scope (i.e., window), candidate answer generation rules, and criteria for determining the correct answer. Inputs involving single regions, multiple regions, bi-temporal observations, reference spectra, and endmembers are organized into a unified multimodal multiple-choice question-answering format. GPT-6.1-sol \cite{openai2026gpt61sol} serves as an auxiliary model for data construction and analysis, assisting in generating natural-language questions, answer options, and explanatory descriptions from original hyperspectral images, corresponding labels, and task templates. The answer keys and the semantic content of the options are determined from the original observations, annotations, or reproducible computational rules. We retain the source dataset, scene, query region, and evidence indices for subsequent verification. Each question can therefore be traced to specific spectral observations, spatial information, or annotation records rather than subjective descriptions. We provide three examples of structured prompts for task construction in Figs.~\ref{app:fig:01_extremum_localization}-\ref{app:fig:03_regional_object_counting}. 

\subsection{Quality Control}\label{app:sec:quality_control}
\begin{figure*}[t]
\centering
\includegraphics[width=\textwidth]{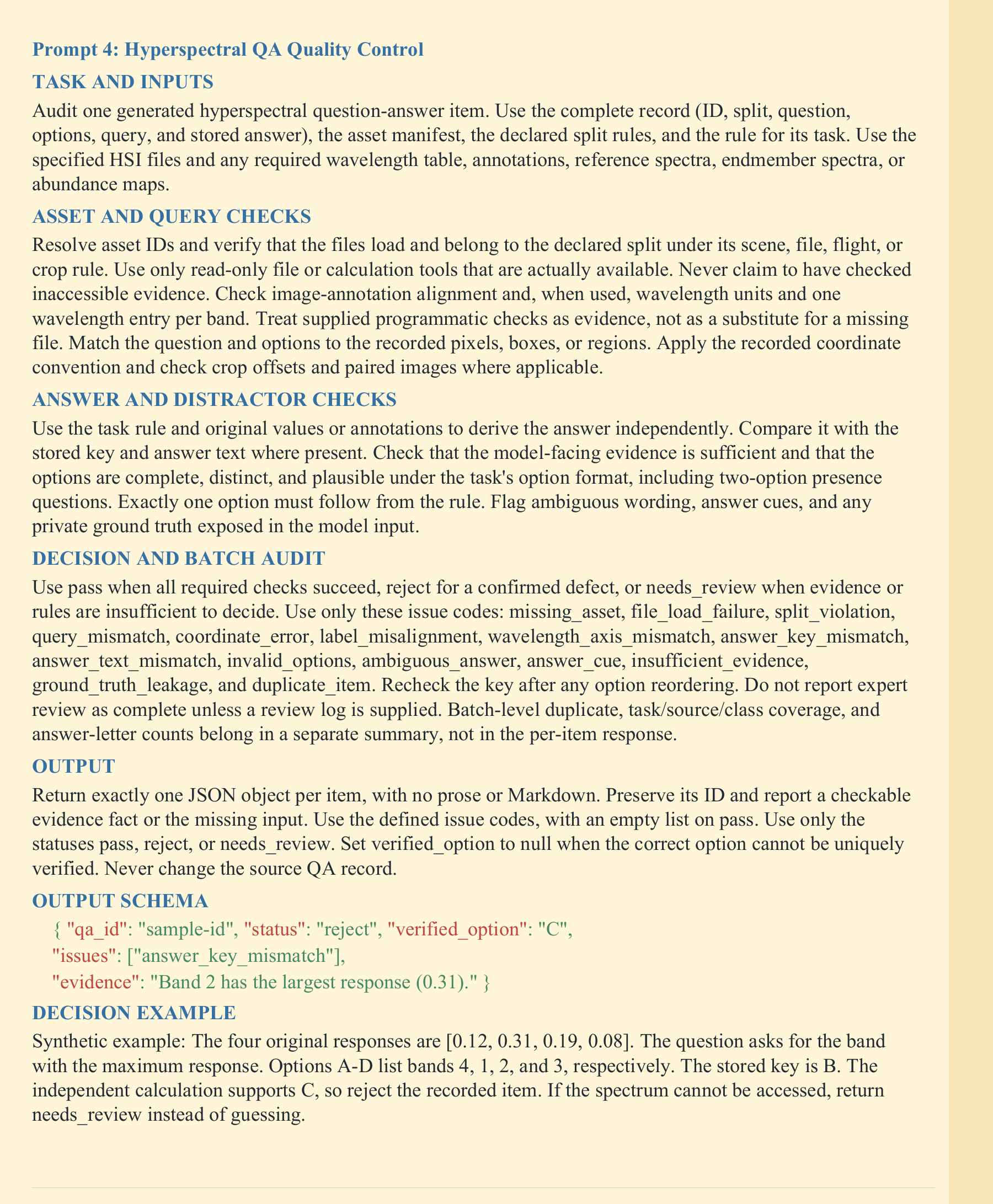}
\caption{Structured prompt for hyperspectral question-answer quality control.}
\label{app:fig:qa_quality_control_prompt}
\end{figure*}

 Quality control consists of automated consistency checks, statistical audits, and manual sampling inspections. We use GPT-6.1 to assist with the automated checks, which cover the validity of data paths and successful data loading, spatial alignment between images and annotations, agreement between the number of wavelength entries and spectral bands, correspondence between question descriptions and query regions, completeness of answer options, and consistency of answer keys with the underlying evidence. Records containing missing, conflicting, or irreproducible information are identified and excluded from training and evaluation. We then examine coverage and duplication across task categories, fine-grained tasks, data sources, classes or materials, and input formats. Underrepresented spectral patterns and answer semantics are supplemented where necessary. Ten experts in hyperspectral imaging or remote sensing are invited to manually inspect sampled questions, focusing on question clarity, sufficiency of the provided evidence, consistency between reference answers and source data, and potential ambiguity or superficial clues in distractors. Answer options are randomly reordered under controlled constraints, and their positional distributions are examined across task groups to maintain approximately balanced answer positions. The structured prompts for quality control is shown in Fig.~\ref{app:fig:qa_quality_control_prompt}.
 
\section{Detailed Experimental Results}\label{app:sec:experiments}

\subsection{Comparative Experiments}\label{app:sec:comparisons}

Tables~\ref{app:tab:fine_spectral_curve}--\ref{app:tab:fine_spectral_unmixing}
report the accuracy of each model on the 34 fine-grained HyperBench tasks
across seven task groups. HyperGPT ranks in the top two on all 34 fine-grained tasks and achieves the highest accuracy on 31 tasks. These first-place results cover every fine-grained task in six of the seven task families. In spectral curve analysis, HyperGPT leads on extrema counting and ranks second on the other three tasks, where fine-tuned InternVL3.5-4B \cite{internvl35} leads. One possible explanation is that spectral curve analysis questions provide reference spectra, allowing some questions about extrema and trends to be answered directly from the provided curves without requiring a full interpretation of the raw hyperspectral images. The remaining six task categories evaluate broader hyperspectral understanding capabilities. HyperGPT achieves higher accuracy on these tasks, possibly benefiting from its explicit modeling of spectral-spatial relationships. At the task group level, HyperGPT-4B achieves the highest overall accuracy in five families, while HyperGPT-2B leads in reference spectral retrieval. This indicates that the benefits of model scaling are not uniform across tasks. We speculate that increasing model capacity helps enhance general representation learning but may weaken the learning performance for specific land-cover categories in the reference retrieval task.

\newcommand{\FineSection}[2]{%
  \arrayrulecolor{TableSeparator}\specialrule{.45pt}{0pt}{0pt}%
  \rowcolor{TableGroup}%
  \multicolumn{#1}{l}{\hspace*{-\tabcolsep}\textcolor{TableAccent}{\itshape #2}}\\}
\newcommand{\FineHead}[2]{\parbox[c][32pt][c]{\linewidth}{\centering\fontsize{7.5}{8}\selectfont\bfseries #1\\#2}}
\newcommand{\FineHeadNarrow}[2]{\parbox[c][32pt][c]{\linewidth}{\centering\fontsize{7}{7.5}\selectfont\bfseries #1\\#2}}
\newcommand{\FineHeadSingle}[1]{\parbox[c][29pt][c]{\linewidth}{\centering\fontsize{7.5}{8}\selectfont\bfseries #1}}

\smallskip
\begin{table*}[t]
\centering
\captionsetup{skip=4pt}
\captionof{table}{Fine-grained accuracy (\%) on spectral curve analysis.}
\label{app:tab:fine_spectral_curve}
\begingroup
\centering
\color{TableInk}
\setlength{\tabcolsep}{4pt}
\renewcommand{\arraystretch}{1.20}
\setlength{\extrarowheight}{0pt}
\renewcommand{\tabularxcolumn}[1]{m{#1}}
\fontsize{8.5}{9.5}\selectfont
\begin{tabularx}{\textwidth}{J{4.8cm}*{5}{Y}}
\arrayrulecolor{TableRule}\specialrule{1pt}{0pt}{0pt}
\rowcolor{TableHeader}
\MainMethodHead & \FineHeadSingle{Extremum localization} & \FineHeadSingle{Trend recognition} & \FineHeadSingle{Shape recognition} & \FineHeadSingle{Extrema counting} & \cellcolor{TableHeader}\MainSingleHead{Overall} \\
\FineSection{6}{MLLMs (Direct Evaluation)}
Qwen3.5-2B~\cite{qwen35} & 25.52 & 27.92 & 25.56 & 24.20 & 25.89 \\
Qwen3.5-4B~\cite{qwen35} & 25.01 & 36.28 & 34.03 & 25.09 & 29.18 \\
Qwen3-VL-2B~\cite{qwen3vl} & 27.32 & 31.86 & 20.84 & 27.40 & 26.85 \\
InternVL3.5-4B~\cite{internvl35} & 29.48 & 12.39 & 31.78 & 49.82 & 28.40 \\
MiniCPM-V-4.7-1B~\cite{minicpm47} & 24.85 & 25.91 & 23.61 & 24.20 & 24.74 \\
SmolVLM2-2.2B~\cite{smolvlm2} & 13.56 & 13.76 & 12.44 & 8.36 & 12.90 \\
GeoChat-7B~\cite{geochat} & 13.02 & 16.81 & 14.02 & 13.17 & 14.00 \\
EarthDial-4B-MS~\cite{earthdial} & 26.29 & 24.86 & 24.36 & 32.21 & 26.13 \\
\FineSection{6}{MLLMs (Supervised Fine-tuning)}
Qwen3.5-2B~\cite{qwen35} & 68.47 & 72.24 & 75.64 & 66.19 & 70.54 \\
InternVL3.5-4B~\cite{internvl35} & \MainBest{77.79} & \MainBest{87.37} & \MainBest{86.28} & 68.51 & \MainBest{80.67} \\
\FineSection{6}{HyperGPT}
HyperGPT-2B & 75.09 & \MainSecondBest{78.76} & \MainSecondBest{77.44} & \MainSecondBest{69.22} & 75.79 \\
HyperGPT-4B & \MainSecondBest{77.08} & 78.68 & 75.19 & \MainBest{71.17} & \MainSecondBest{76.46} \\
\arrayrulecolor{TableRule}\specialrule{1pt}{0pt}{0pt}
\end{tabularx}
\endgroup
\end{table*}

\begin{table*}[t]
\centering
\captionsetup{skip=4pt}
\caption{Fine-grained accuracy (\%) on spatial semantic understanding.}
\label{app:tab:fine_spatial_semantic}
\begingroup
\centering
\color{TableInk}
\setlength{\tabcolsep}{4pt}
\renewcommand{\arraystretch}{1.20}
\setlength{\extrarowheight}{0pt}
\renewcommand{\tabularxcolumn}[1]{m{#1}}
\fontsize{8.5}{9.5}\selectfont
\begin{tabularx}{\textwidth}{J{4.4cm}*{9}{Y}}
\arrayrulecolor{TableRule}\specialrule{1pt}{0pt}{0pt}
\rowcolor{TableHeader}
\MainMethodHead & \FineHeadNarrow{Pixel}{classifi-\\cation} & \FineHeadNarrow{Class-\\guided}{pixel\\retrieval} & \FineHeadNarrow{Same-class}{matching} & \FineHeadNarrow{Support-\\guided}{classifi-\\cation} & \FineHeadNarrow{Regional}{compo-\\sition} & \FineHeadNarrow{Class}{spatial\\relations} & \FineHeadNarrow{Class-\\guided}{region\\retrieval} & \FineHeadNarrow{Regional}{count\\compari-\\son} & \cellcolor{TableHeader}\MainSingleHead{Overall} \\
\FineSection{10}{MLLMs (Direct Evaluation)}
Qwen3.5-2B & 24.00 & 21.39 & 24.37 & 23.74 & 23.90 & 27.24 & 25.19 & 25.60 & 24.26 \\
Qwen3.5-4B & 25.50 & 26.56 & 28.32 & 24.20 & 25.80 & 22.56 & 31.11 & 24.19 & 26.00 \\
Qwen3-VL-2B & 31.40 & 19.96 & 26.16 & 28.08 & 25.34 & 26.83 & 24.62 & 27.02 & 26.57 \\
InternVL3.5-4B & 27.24 & 26.38 & 24.91 & 5.02 & 23.77 & 15.24 & 27.48 & 28.02 & 23.41 \\
MiniCPM-V-4.7-1B & 24.67 & 25.85 & 26.16 & 23.52 & 24.10 & 23.58 & 24.81 & 24.60 & 24.60 \\
SmolVLM2-2.2B & 15.53 & 0.00 & 0.00 & 0.00 & 17.20 & 20.12 & 0.00 & 22.38 & 11.37 \\
GeoChat-7B & 12.54 & 0.00 & 0.00 & 0.00 & 18.98 & 17.28 & 3.44 & 15.52 & 10.70 \\
EarthDial-4B-MS & 28.07 & 22.99 & 22.40 & 26.26 & 26.40 & 25.20 & 25.76 & 24.19 & 25.67 \\
\FineSection{10}{MLLMs (Supervised Fine-tuning)}
Qwen3.5-2B & 32.97 & 26.74 & 39.25 & 39.95 & 41.76 & 61.99 & 38.17 & 16.13 & 37.30 \\
InternVL3.5-4B & 39.78 & \MainSecondBest{44.03} & 45.70 & 10.05 & 47.60 & 64.02 & \MainSecondBest{48.66} & 25.00 & 42.17 \\
\FineSection{10}{HyperGPT}
HyperGPT-2B & \MainSecondBest{57.23} & 39.04 & \MainSecondBest{52.69} & \MainSecondBest{65.98} & \MainSecondBest{58.90} & \MainBest{77.24} & 48.09 & \MainSecondBest{41.33} & \MainSecondBest{55.64} \\
HyperGPT-4B & \MainBest{60.22} & \MainBest{63.81} & \MainBest{54.30} & \MainBest{75.34} & \MainBest{64.15} & \MainSecondBest{76.42} & \MainBest{69.85} & \MainBest{47.18} & \MainBest{63.30} \\
\arrayrulecolor{TableRule}\specialrule{1pt}{0pt}{0pt}
\end{tabularx}
\endgroup
\end{table*}

\begin{table*}[t]
\centering
\captionsetup{skip=4pt}
\caption{Fine-grained accuracy (\%) on object instance analysis.}
\label{app:tab:fine_object_instance}
\begingroup
\centering
\color{TableInk}
\setlength{\tabcolsep}{4pt}
\renewcommand{\arraystretch}{1.20}
\setlength{\extrarowheight}{0pt}
\renewcommand{\tabularxcolumn}[1]{m{#1}}
\fontsize{8.5}{9.5}\selectfont
\begin{tabularx}{\textwidth}{J{4.8cm}*{6}{Y}}
\arrayrulecolor{TableRule}\specialrule{1pt}{0pt}{0pt}
\rowcolor{TableHeader}
\MainMethodHead & \FineHeadSingle{Global counting} & \FineHeadSingle{Regional counting} & \FineHead{Single-object}{localization} & \FineHead{Object-set}{localization} & \FineHead{Object}{spatial relations} & \cellcolor{TableHeader}\MainSingleHead{Overall} \\
\FineSection{7}{MLLMs (Direct Evaluation)}
Qwen3.5-2B & 24.13 & 25.67 & 20.81 & 24.25 & 24.44 & 23.68 \\
Qwen3.5-4B & 42.04 & 29.97 & 20.52 & 24.92 & 24.94 & 27.85 \\
Qwen3-VL-2B & 57.46 & 48.37 & 30.20 & 24.58 & 29.43 & 38.79 \\
InternVL3.5-4B & 60.20 & 52.82 & 26.16 & 30.56 & 30.92 & 40.28 \\
MiniCPM-V-4.7-1B & 25.37 & 24.78 & 27.46 & 28.90 & 27.68 & 26.60 \\
SmolVLM2-2.2B & 39.80 & 30.56 & 21.53 & 23.92 & 23.69 & 27.61 \\
GeoChat-7B & 32.84 & 40.65 & 29.77 & 24.25 & 24.19 & 31.66 \\
EarthDial-4B-MS & 39.55 & 34.12 & 24.86 & 25.25 & 24.69 & 29.80 \\
\FineSection{7}{MLLMs (Supervised Fine-tuning)}
Qwen3.5-2B & 52.24 & 64.69 & 36.56 & 34.22 & 44.39 & 47.77 \\
InternVL3.5-4B\ & 58.96 & 66.17 & 47.11 & \MainSecondBest{50.83} & \MainSecondBest{50.37} & 55.22 \\
\FineSection{7}{HyperGPT}
HyperGPT-2B & \MainSecondBest{76.87} & \MainSecondBest{80.12} & \MainSecondBest{51.01} & 43.52 & \MainBest{64.59} & \MainSecondBest{64.45} \\
HyperGPT-4B & \MainBest{80.10} & \MainBest{81.45} & \MainBest{86.13} & \MainBest{71.10} & \MainBest{64.59} & \MainBest{78.54} \\
\arrayrulecolor{TableRule}\specialrule{1pt}{0pt}{0pt}
\end{tabularx}
\endgroup
\end{table*}

\begin{table*}[t]
\centering
\captionsetup{skip=4pt}
\caption{Fine-grained accuracy (\%) on change detection.}
\label{app:tab:fine_change_detection}
\begingroup
\centering
\color{TableInk}
\setlength{\tabcolsep}{4pt}
\renewcommand{\arraystretch}{1.20}
\setlength{\extrarowheight}{0pt}
\renewcommand{\tabularxcolumn}[1]{m{#1}}
\fontsize{8.5}{9.5}\selectfont
\begin{tabularx}{\textwidth}{J{4.8cm}*{6}{Y}}
\arrayrulecolor{TableRule}\specialrule{1pt}{0pt}{0pt}
\rowcolor{TableHeader}
\MainMethodHead & \FineHeadSingle{Change presence} & \FineHead{Change}{localization} & \FineHead{Changed-area}{fraction} & \FineHead{Spectral change}{localization} & \FineHead{Change-region}{counting} & \cellcolor{TableHeader}\MainSingleHead{Overall} \\
\FineSection{7}{MLLMs (Direct Evaluation)}
Qwen3.5-2B & 4.29 & 23.42 & 24.06 & 23.78 & \MainBest{35.07} & 20.67 \\
Qwen3.5-4B & 35.24 & 31.65 & 25.67 & 24.48 & 19.40 & 28.00 \\
Qwen3-VL-2B & 42.86 & 33.54 & 23.53 & 24.48 & 32.09 & 31.85 \\
InternVL3.5-4B & 65.71 & 23.42 & 26.74 & 22.38 & 16.42 & 33.53 \\
MiniCPM-V-4.7-1B & 35.24 & 21.52 & 24.60 & 18.88 & 7.46 & 22.96 \\
SmolVLM2-2.2B & 0.00 & 0.00 & 0.00 & 0.00 & 0.00 & 0.00 \\
GeoChat-7B & 0.00 & 0.00 & 0.00 & 0.00 & 0.00 & 0.00 \\
EarthDial-4B-MS & 64.76 & 25.32 & 24.60 & 26.57 & \MainSecondBest{33.58} & 36.66 \\
\FineSection{7}{MLLMs (Supervised Fine-tuning)}
Qwen3.5-2B & 24.76 & 14.56 & 3.74 & 37.76 & 26.87 & 20.67 \\
InternVL3.5-4B & \MainSecondBest{82.38} & 28.48 & \MainSecondBest{28.88} & \MainSecondBest{52.45} & 25.37 & 45.79 \\
\FineSection{7}{HyperGPT}
HyperGPT-2B & 79.05 & \MainSecondBest{50.63} & 25.13 & 43.36 & \MainBest{35.07} & \MainSecondBest{48.32} \\
HyperGPT-4B & \MainBest{92.38} & \MainBest{62.03} & \MainBest{31.55} & \MainBest{53.15} & \MainSecondBest{33.58} & \MainBest{56.73} \\
\arrayrulecolor{TableRule}\specialrule{1pt}{0pt}{0pt}
\end{tabularx}
\endgroup
\end{table*}

\begin{table*}[t]
\centering
\captionsetup{skip=4pt}
\caption{Fine-grained accuracy (\%) on anomaly detection.}
\label{app:tab:fine_anomaly_detection}
\begingroup
\centering
\color{TableInk}
\setlength{\tabcolsep}{4pt}
\renewcommand{\arraystretch}{1.20}
\setlength{\extrarowheight}{0pt}
\renewcommand{\tabularxcolumn}[1]{m{#1}}
\fontsize{8.5}{9.5}\selectfont
\begin{tabularx}{\textwidth}{J{4.8cm}*{5}{Y}}
\arrayrulecolor{TableRule}\specialrule{1pt}{0pt}{0pt}
\rowcolor{TableHeader}
\MainMethodHead & \FineHeadSingle{Anomaly presence} & \FineHeadSingle{Anomaly localization} & \FineHead{Anomalous-area}{fraction} & \FineHead{Spectral anomaly}{contrast} & \cellcolor{TableHeader}\MainSingleHead{Overall} \\
\FineSection{6}{MLLMs (Direct Evaluation)}
Qwen3.5-2B & 70.28 & 26.44 & 24.66 & 23.08 & 40.82 \\
Qwen3.5-4B & 44.18 & 25.19 & 23.00 & 39.32 & 31.83 \\
Qwen3-VL-2B & 50.74 & 24.73 & 24.36 & 7.69 & 32.66 \\
InternVL3.5-4B & 52.07 & 23.64 & 26.17 & \MainSecondBest{47.01} & 35.47 \\
MiniCPM-V-4.7-1B & 49.80 & 24.42 & 21.33 & 24.79 & 32.24 \\
SmolVLM2-2.2B & 50.20 & 23.48 & \MainSecondBest{31.47} & 5.98 & 34.18 \\
GeoChat-7B & 71.49 & 12.13 & 22.84 & 24.79 & 36.53 \\
EarthDial-4B-MS & 50.47 & 27.22 & 30.11 & 26.50 & 36.07 \\
\FineSection{6}{MLLMs (Supervised Fine-tuning)}
Qwen3.5-2B & \MainSecondBest{97.59} & 15.40 & 17.85 & 9.40 & 44.14 \\
InternVL3.5-4B\ & \MainBest{100.00} & 17.57 & 25.72 & 37.61 & 49.54 \\
\FineSection{6}{HyperGPT}
HyperGPT-2B\ & \MainBest{100.00} & \MainSecondBest{35.46} & 31.32 & 46.15 & \MainSecondBest{57.01} \\
HyperGPT-4B & \MainBest{100.00} & \MainBest{48.99} & \MainBest{41.45} & \MainBest{78.63} & \MainBest{65.87} \\
\arrayrulecolor{TableRule}\specialrule{1pt}{0pt}{0pt}
\end{tabularx}
\endgroup
\end{table*}

\begin{table*}[t]
\centering
\captionsetup{skip=4pt}
\caption{Fine-grained accuracy (\%) on reference spectrum retrieval.}
\label{app:tab:fine_reference_retrieval}
\begingroup
\centering
\color{TableInk}
\setlength{\tabcolsep}{4pt}
\renewcommand{\arraystretch}{1.20}
\setlength{\extrarowheight}{0pt}
\renewcommand{\tabularxcolumn}[1]{m{#1}}
\fontsize{8.5}{9.5}\selectfont
\begin{tabularx}{\textwidth}{J{4.8cm}*{3}{Y}}
\arrayrulecolor{TableRule}\specialrule{1pt}{0pt}{0pt}
\rowcolor{TableHeader}
\MainMethodHead & \FineHeadSingle{Target presence} & \FineHeadSingle{Target localization} & \cellcolor{TableHeader}\MainSingleHead{Overall} \\
\FineSection{4}{MLLMs (Direct Evaluation)}
Qwen3.5-2B & 42.89 & 16.00 & 29.24 \\
Qwen3.5-4B & 41.03 & 24.60 & 32.69 \\
Qwen3-VL-2B & 49.28 & 39.80 & 44.47 \\
InternVL3.5-4B & 46.80 & 45.80 & 46.29 \\
MiniCPM-V-4.7-1B & 40.82 & 25.00 & 32.79 \\
SmolVLM2-2.2B & 0.00 & 0.00 & 0.00 \\
GeoChat-7B & 0.00 & 0.00 & 0.00 \\
EarthDial-4B-MS & 50.52 & \MainSecondBest{59.00} & 54.82 \\
\FineSection{4}{MLLMs (Supervised Fine-tuning)}
Qwen3.5-2B & 38.56 & \MainBest{100.00} & 69.75 \\
InternVL3.5-4B & 49.90 & \MainBest{100.00} & 75.33 \\
\FineSection{4}{HyperGPT}
HyperGPT-2B & \MainBest{63.30} & \MainBest{100.00} & \MainBest{81.93} \\
HyperGPT-4B & \MainSecondBest{55.67} & \MainBest{100.00} & \MainSecondBest{78.17} \\
\arrayrulecolor{TableRule}\specialrule{1pt}{0pt}{0pt}
\end{tabularx}
\endgroup
\end{table*}

\begin{table*}[t]
\centering
\captionsetup{skip=4pt}
\caption{Fine-grained accuracy (\%) on spectral unmixing analysis.}
\label{app:tab:fine_spectral_unmixing}
\begingroup
\centering
\color{TableInk}
\setlength{\tabcolsep}{4pt}
\renewcommand{\arraystretch}{1.20}
\setlength{\extrarowheight}{0pt}
\renewcommand{\tabularxcolumn}[1]{m{#1}}
\fontsize{8.5}{9.5}\selectfont
\begin{tabularx}{\textwidth}{J{4.8cm}*{7}{Y}}
\arrayrulecolor{TableRule}\specialrule{1pt}{0pt}{0pt}
\rowcolor{TableHeader}
\MainMethodHead & \FineHead{Abundance}{estimation} & \FineHead{Abundance}{difference} & \FineHead{Abundance}{ranking} & \FineHead{Thresh-\\olded}{composition} & \FineHead{Dominant}{material} & \FineHead{Dominance-\\based}{retrieval} & \cellcolor{TableHeader}\MainSingleHead{Overall} \\
\FineSection{8}{MLLMs (Direct Evaluation)}
Qwen3.5-2B & 24.60 & 23.20 & 23.79 & 25.92 & 17.59 & 22.87 & 23.88 \\
Qwen3.5-4B & 22.49 & 28.00 & 26.61 & 26.96 & 31.48 & 27.13 & 26.22 \\
Qwen3-VL-2B & 25.40 & 19.20 & 25.81 & 31.68 & 28.70 & 27.13 & 27.08 \\
InternVL3.5-4B & 23.81 & 24.00 & 26.61 & 28.01 & 17.59 & 25.58 & 25.22 \\
MiniCPM-V-4.7-1B & 25.13 & 28.00 & 25.40 & 24.87 & 28.70 & 20.54 & 24.82 \\
SmolVLM2-2.2B & 0.00 & 0.00 & 0.00 & 0.00 & 0.00 & 0.00 & 0.00 \\
GeoChat-7B & 0.00 & 0.00 & 0.00 & 0.00 & 0.00 & 0.00 & 0.00 \\
EarthDial-4B-MS & 23.28 & 22.40 & 24.19 & 27.23 & 33.33 & 23.64 & 25.15 \\
\FineSection{8}{MLLMs (Supervised Fine-tuning)}
Qwen3.5-2B & 18.25 & 15.20 & 20.56 & 19.90 & 25.93 & 19.77 & 19.61 \\
InternVL3.5-4B & 27.51 & \MainSecondBest{33.60} & 35.48 & 38.74 & 39.81 & \MainSecondBest{40.70} & 35.36 \\
\FineSection{8}{HyperGPT}
HyperGPT-2B & \MainSecondBest{39.42} & 32.00 & \MainSecondBest{37.10} & \MainSecondBest{43.72} & \MainBest{47.22} & 38.37 & \MainSecondBest{39.89} \\
HyperGPT-4B & \MainBest{41.53} & \MainBest{40.80} & \MainBest{52.02} & \MainBest{48.17} & \MainSecondBest{43.52} & \MainBest{47.29} & \MainBest{46.03} \\
\arrayrulecolor{TableRule}\specialrule{1pt}{0pt}{0pt}
\end{tabularx}
\endgroup
\end{table*}

\subsection{Ablation Studies}\label{app:sec:ablation}

Table~\ref{app:tab:ablation} presents the task-level performance variations in the component ablation experiments reported in the main paper, where the performance of each task group is measured by the overall accuracy across all questions in its constituent subtasks. Adding SPE to the joint-encoding variant improves change detection by 11.90 percentage points, suggesting that wavelength-aware positions help the model compare spectral responses across observations. With SPE enabled, adding SDPE improves object instance analysis by 5.99 points, indicating that response differences between adjacent spectral groups can provide additional cues for distinguishing objects. Adding CSPA to the SPE-only variant improves anomaly detection and spectral unmixing by 5.12 and 8.54 points, respectively. These gains are consistent with the design of CSPA, which relates responses across spectral blocks at the same spatial location. The full model improves accuracy across all seven task groups over joint encoding alone and leads in spectral curve analysis, object instance analysis, and reference spectrum retrieval. 

\begin{table*}[t]
\centering
\caption{Ablation results (accuracy, \%) for HyperGPT-2B on HyperBench.}
\label{app:tab:ablation}
\begingroup
\color{TableInk}
\setlength{\tabcolsep}{4.1pt}
\renewcommand{\arraystretch}{1}
\fontsize{11}{12}\selectfont
\setlength{\arrayrulewidth}{0.5pt}
\setlength{\doublerulesep}{2pt}
\resizebox{\linewidth}{!}{%
\begin{tabular}{
  J{105pt}A{21pt}A{28pt}A{28pt}
  ||
  >{\columncolor{TableOverall}}A{50pt}
  *{7}{A{50pt}}
}
\arrayrulecolor{TableRule}\specialrule{1pt}{0pt}{0pt}
\rowcolor{TableHeader}
\textbf{Method} & \AblHead{SPE} & \AblHead{SDPE} & \AblHead{CSPA}
& \AblHead{Overall} & \AblHead{Spectral\\curve}
& \AblHead{Semantic\\spatial} & \AblHead{Object\\instance}
& \AblHead{Change\\detection} & \AblHead{Anomaly\\detection}
& \AblHead{Reference\\retrieval} & \AblHead{Spectral\\unmixing} \\
\arrayrulecolor{TableRule}\specialrule{1pt}{0pt}{0pt}

Qwen3.5-2B
& \AblBase{\AblNo} & \AblBase{\AblNo} & \AblBase{\AblNo}
& \AblBase{49.31} & \AblBase{70.54} & \AblBase{37.30} & \AblBase{47.77}
& \AblBase{20.67} & \AblBase{44.14} & \AblBase{69.75} & \AblBase{19.61} \\

HyperGPT-2B
& \AblBase{\AblNo} & \AblBase{\AblNo} & \AblBase{\AblNo}
& \AblBase{57.93} & \AblBase{72.86} & \AblBase{49.03} & \AblBase{56.23}
& \AblBase{43.03} & \AblBase{54.98} & \AblBase{76.35} & \AblBase{33.29} \\

HyperGPT-2B
& \AblBase{\AblYes} & \AblBase{\AblNo} & \AblBase{\AblNo}
& \AblScore{60.30}{\textcolor{TableGain}{(+2.37)}}
& \AblScore{73.38}{\textcolor{TableGain}{(+0.52)}}
& \AblScore{53.55}{\textcolor{TableGain}{(+4.52)}}
& \AblScore{58.14}{\textcolor{TableGain}{(+1.91)}}
& \AblFirstCell{\AblScore{\textbf{54.93}}{\textcolor{TableGain}{(+11.90)}}}
& \AblScore{53.23}{\textcolor{TableLoss}{(-1.75)}}
& \AblScore{78.88}{\textcolor{TableGain}{(+2.53)}}
& \AblScore{36.36}{\textcolor{TableGain}{(+3.07)}} \\

HyperGPT-2B
& \AblBase{\AblYes} & \AblBase{\AblYes} & \AblBase{\AblNo}
& \AblScore{61.47}{\textcolor{TableGain}{(+3.54)}}
& \AblScore{73.10}{\textcolor{TableGain}{(+0.24)}}
& \AblScore{55.12}{\textcolor{TableGain}{(+6.09)}}
& \AblSecondCell{\AblScore{64.13}{\textcolor{TableGain}{(+7.90)}}}
& \AblScore{45.43}{\textcolor{TableGain}{(+2.40)}}
& \AblScore{54.89}{\textcolor{TableLoss}{(-0.09)}}
& \AblScore{78.68}{\textcolor{TableGain}{(+2.33)}}
& \AblScore{40.23}{\textcolor{TableGain}{(+6.94)}} \\

HyperGPT-2B
& \AblBase{\AblYes} & \AblBase{\AblNo} & \AblBase{\AblYes}
& \AblScore{61.57}{\textcolor{TableGain}{(+3.64)}}
& \AblScore{71.90}{\textcolor{TableLoss}{(-0.96)}}
& \AblSecondCell{\AblScore{56.23}{\textcolor{TableGain}{(+7.20)}}}
& \AblScore{58.79}{\textcolor{TableGain}{(+2.56)}}
& \AblScore{44.71}{\textcolor{TableGain}{(+1.68)}}
& \AblFirstCell{\AblScore{\textbf{58.35}}{\textcolor{TableGain}{(+3.37)}}}
& \AblSecondCell{\AblScore{81.12}{\textcolor{TableGain}{(+4.77)}}}
& \AblFirstCell{\AblScore{\textbf{44.90}}{\textcolor{TableGain}{(+11.61)}}} \\

HyperGPT-2B
& \AblBase{\AblNo} & \AblBase{\AblYes} & \AblBase{\AblYes}
& \AblSecondCell{\AblScore{61.71}{\textcolor{TableGain}{(+3.78)}}}
& \AblSecondCell{\AblScore{73.41}{\textcolor{TableGain}{(+0.55)}}}
& \AblFirstCell{\AblScore{\textbf{57.87}}{\textcolor{TableGain}{(+8.84)}}}
& \AblScore{55.43}{\textcolor{TableLoss}{(-0.80)}}
& \AblScore{45.31}{\textcolor{TableGain}{(+2.28)}}
& \AblScore{56.46}{\textcolor{TableGain}{(+1.48)}}
& \AblScore{80.00}{\textcolor{TableGain}{(+3.65)}}
& \AblSecondCell{\AblScore{42.76}{\textcolor{TableGain}{(+9.47)}}} \\

\cellcolor{TableOurs}HyperGPT-2B
& \cellcolor{TableOurs}\AblBase{\AblYes}
& \cellcolor{TableOurs}\AblBase{\AblYes}
& \cellcolor{TableOurs}\AblBase{\AblYes}
& \AblFirstCell{\AblScore{\textbf{62.99}}{\textcolor{TableGain}{(+5.06)}}}
& \AblFirstCell{\AblScore{\textbf{75.79}}{\textcolor{TableGain}{(+2.93)}}}
& \AblScore{55.64}{\textcolor{TableGain}{(+6.61)}}
& \AblFirstCell{\AblScore{\textbf{64.45}}{\textcolor{TableGain}{(+8.22)}}}
& \AblSecondCell{\AblScore{48.32}{\textcolor{TableGain}{(+5.29)}}}
& \AblSecondCell{\AblScore{57.01}{\textcolor{TableGain}{(+2.03)}}}
& \AblFirstCell{\AblScore{\textbf{81.93}}{\textcolor{TableGain}{(+5.58)}}}
& \AblScore{39.89}{\textcolor{TableGain}{(+6.60)}} \\
\arrayrulecolor{TableRule}\specialrule{1pt}{0pt}{0pt}
\end{tabular}%
}
\par\vspace{5pt}
{\fontsize{8.5}{10.5}\selectfont\raggedright\color{TableAccent}
\textcolor{TableBest}{\rule{7pt}{6pt}}\ \textbf{Highest score}\qquad
\textcolor{TableSecond}{\rule{7pt}{6pt}}\ Second-highest score\par}
\endgroup
\end{table*}

\subsection{Visualization of fine-grained tasks}\label{app:sec:qualitative_analysis}

We visualize 34 fine-grained task categories of HyperBench in pages~\pageref{task:Extremum localization}--%
\pageref{task:Dominance-based retrieval}. In the spectral analysis examples, models are required to locate the maximum response within a specified wavelength interval and recognize the overall shape of a spectral curve despite local fluctuations. In the spatial understanding examples, models are expected to distinguish anomalous pixels from the background, localize and count target objects, and determine the relative direction between land-cover classes. They are also required to identify regions where changes occur between paired observations. Given a reference spectrum, models need to identify the matching region within the query area. For material composition analysis, models are tasked with ranking candidate locations by the abundance of the target material in mixed spectra. These examples illustrate how HyperBench requires models to exploit spectral and spatial evidence to answer questions under diverse semantic conditions.

\input{appendix/qualitative-examples}

%% file: appendix/qualitative-examples.tex
\begingroup
\setlength{\textfloatsep}{6pt plus 1pt minus 1pt}
\setlength{\floatsep}{6pt plus 1pt minus 1pt}

\newcommand{\HyperGPTQualitativePair}[3]{%
  \begin{figure*}[p]
    \centering
    \phantomsection
    \label{task:#1}
    \vspace*{2pt}
    {\large\bfseries #1}\par\vspace{4pt}
    \includegraphics[width=.94\textwidth,height=.41\textheight,keepaspectratio]{#2}\par\vspace{5pt}
    \includegraphics[width=.94\textwidth,height=.41\textheight,keepaspectratio]{#3}
  \end{figure*}
}

\HyperGPTQualitativePair{Extremum localization}{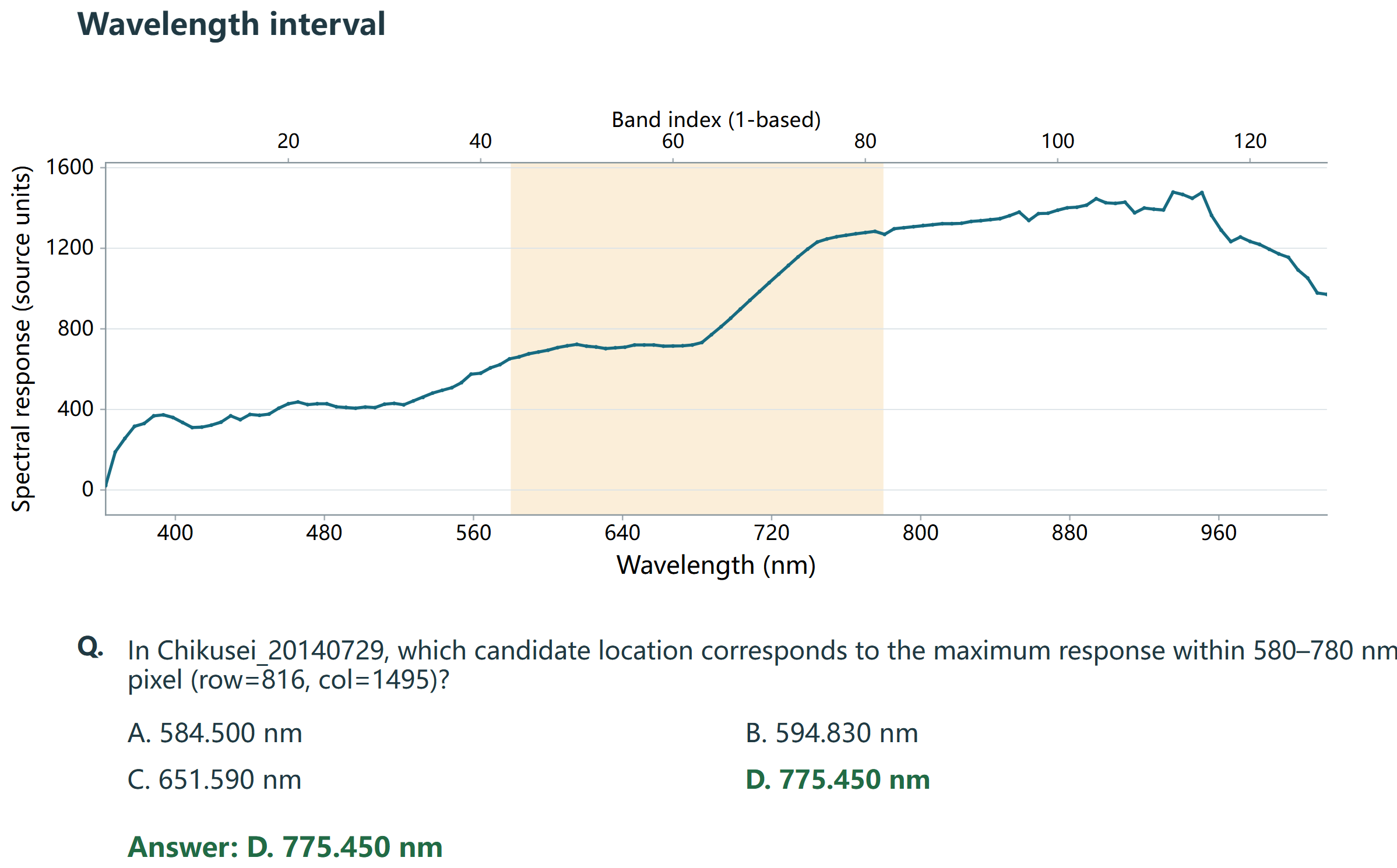}{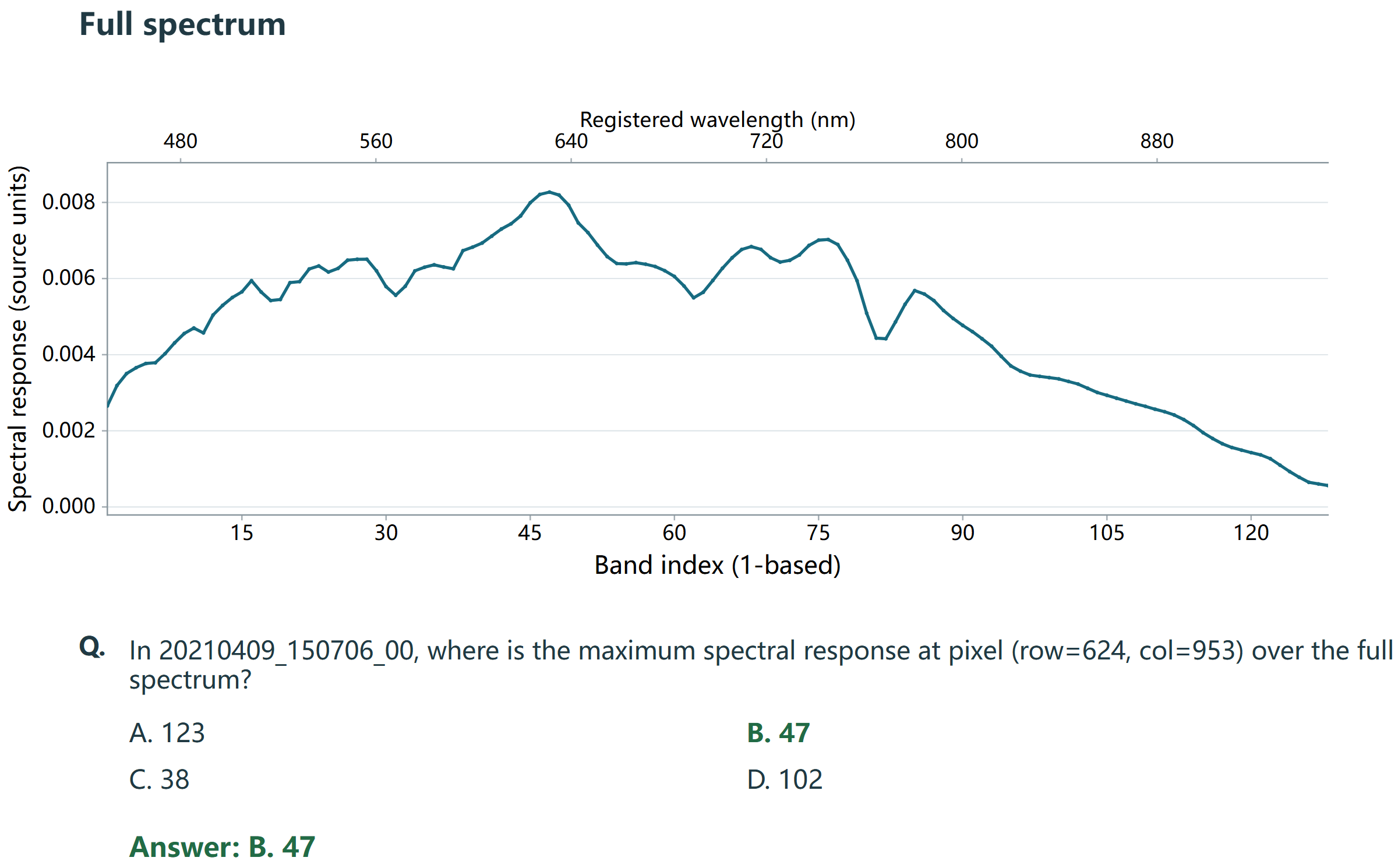}
\clearpage
\HyperGPTQualitativePair{Trend recognition}{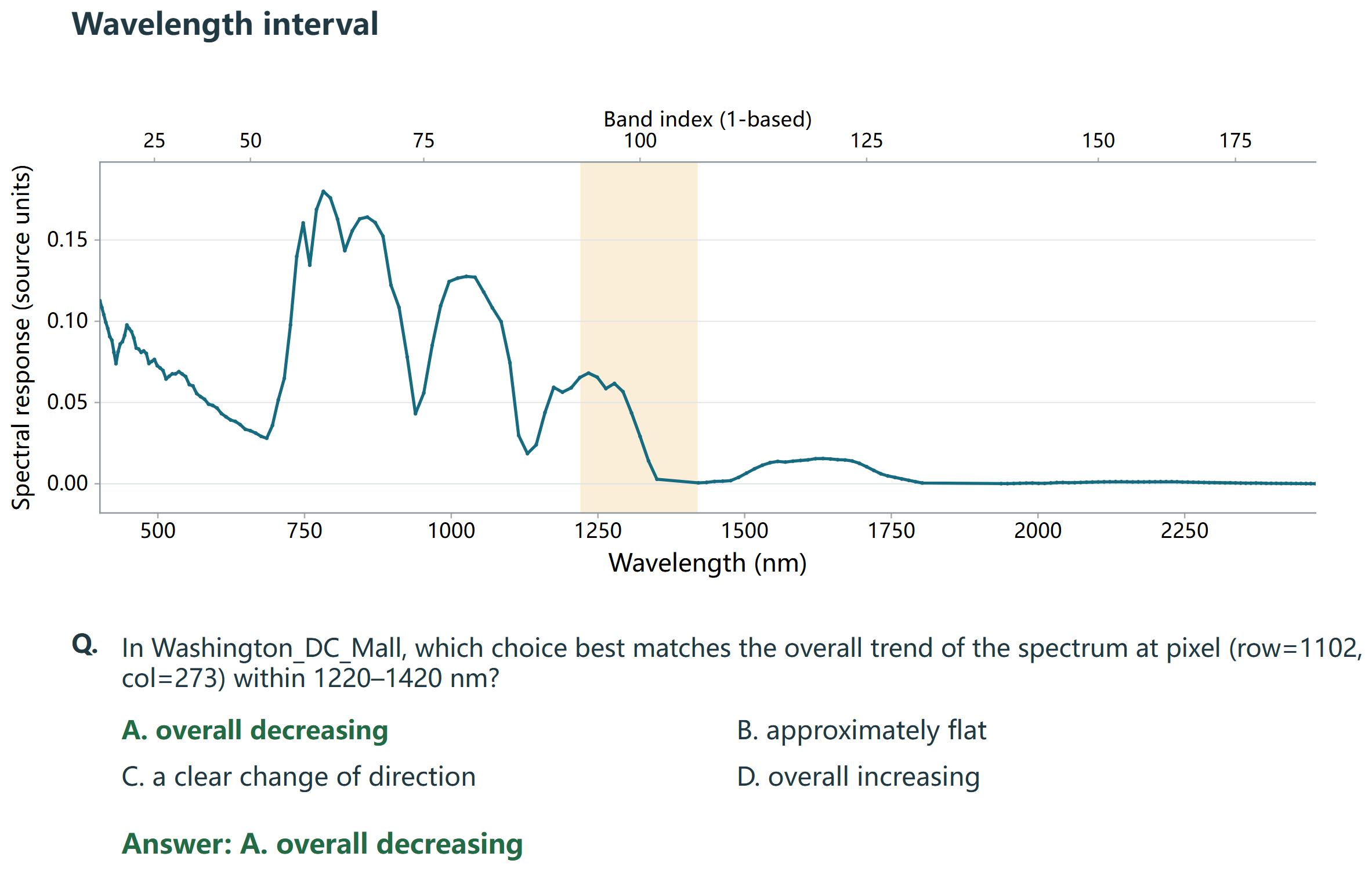}{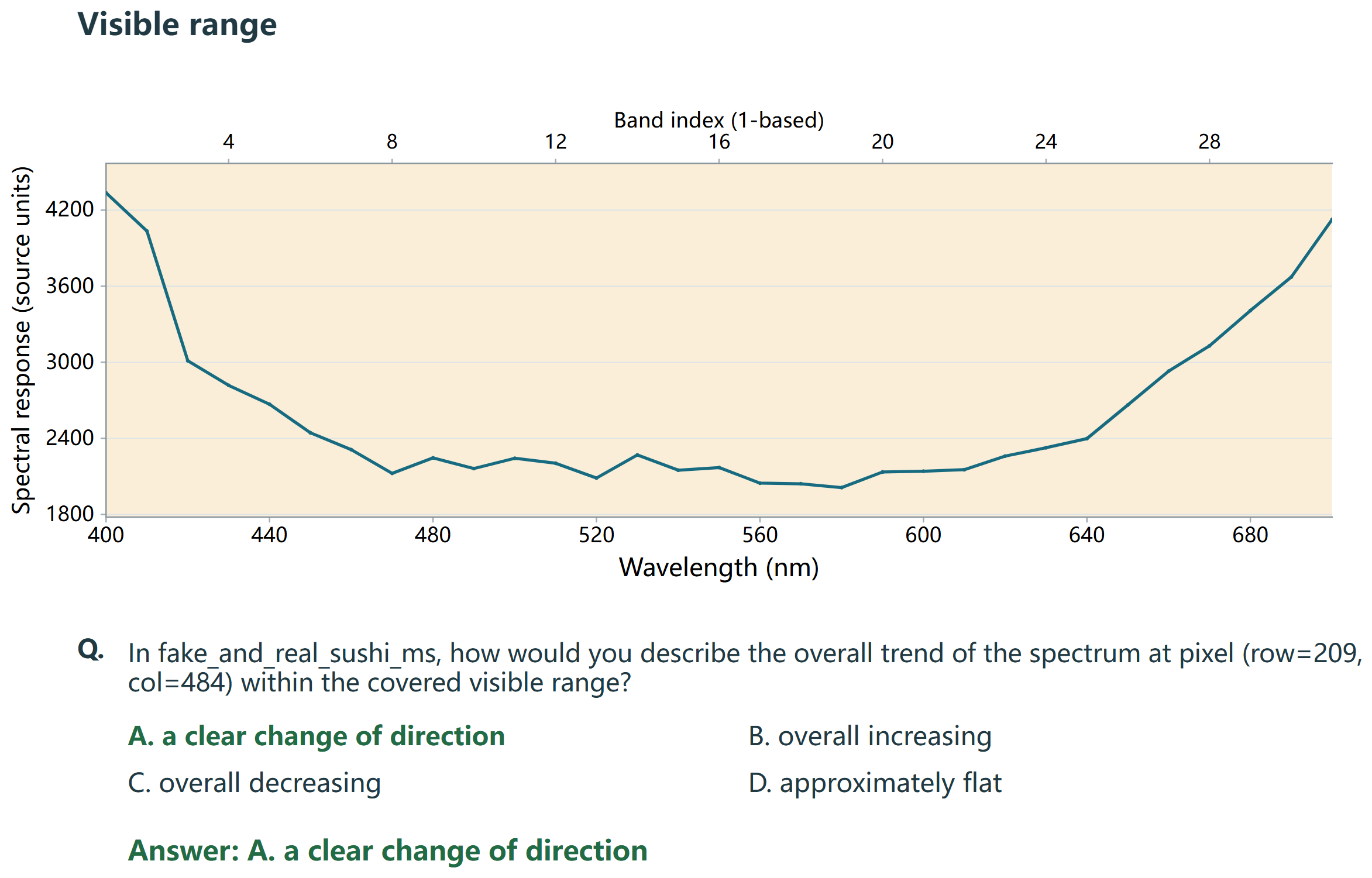}
\clearpage
\HyperGPTQualitativePair{Shape recognition}{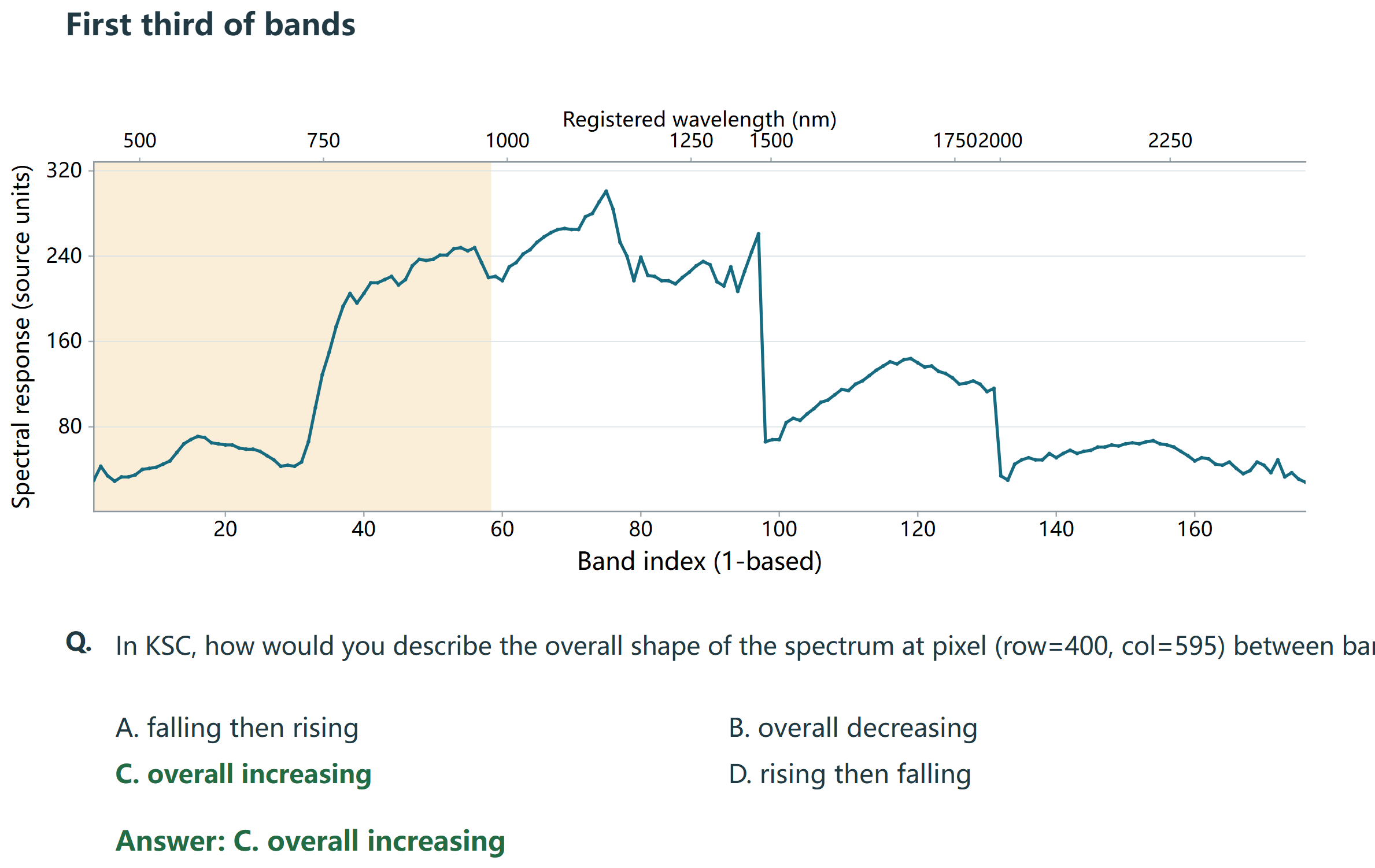}{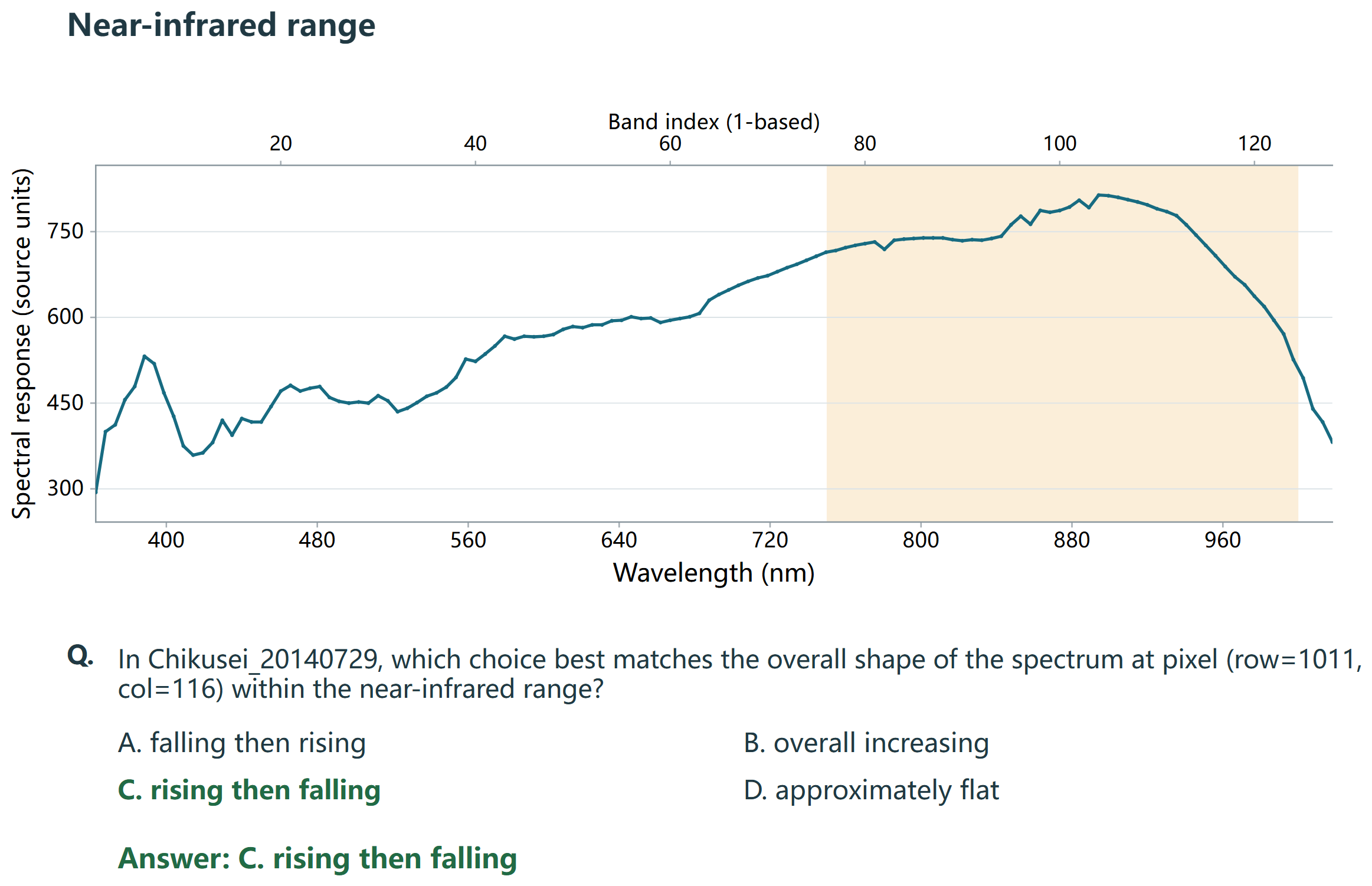}
\clearpage
\HyperGPTQualitativePair{Extrema counting}{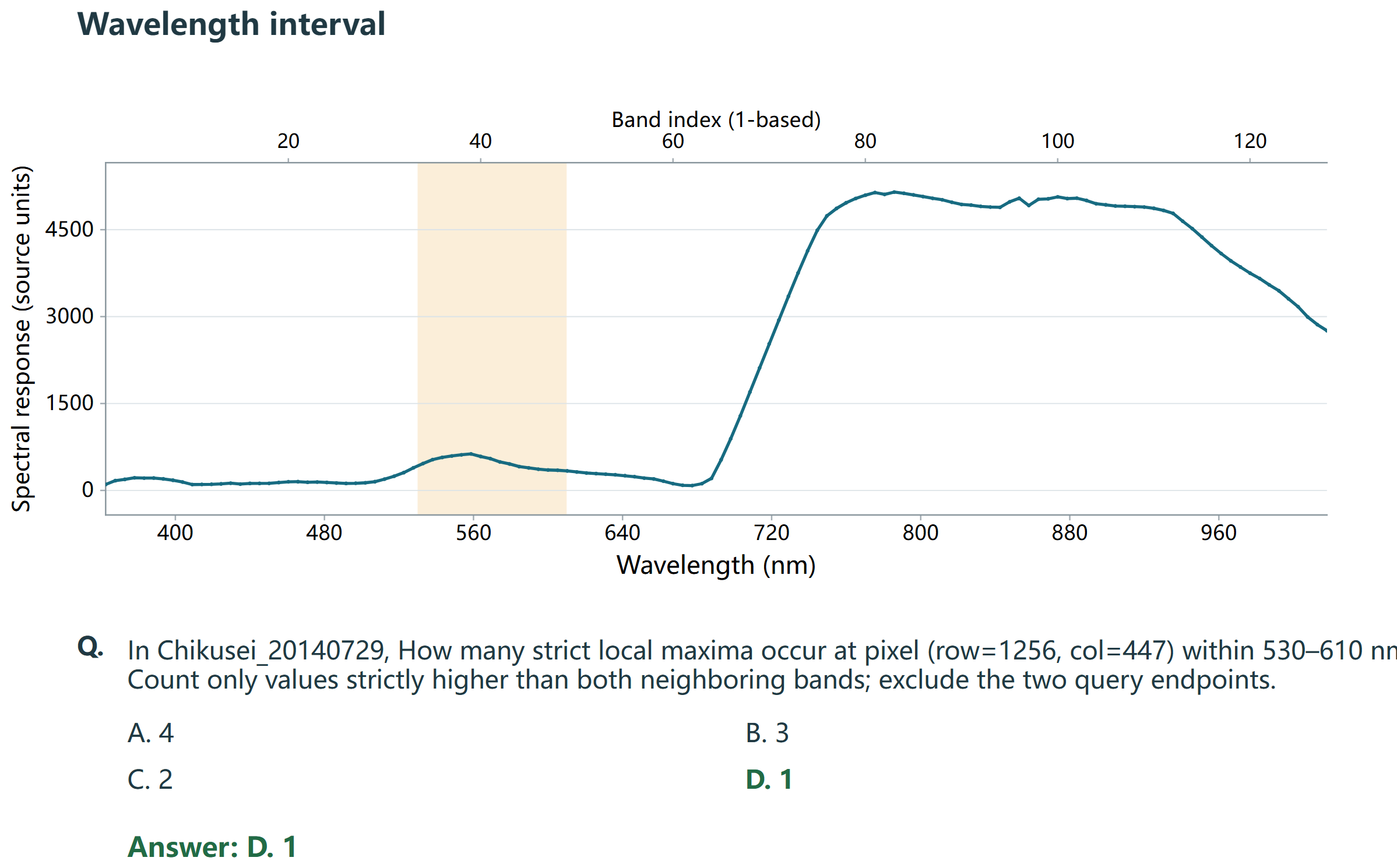}{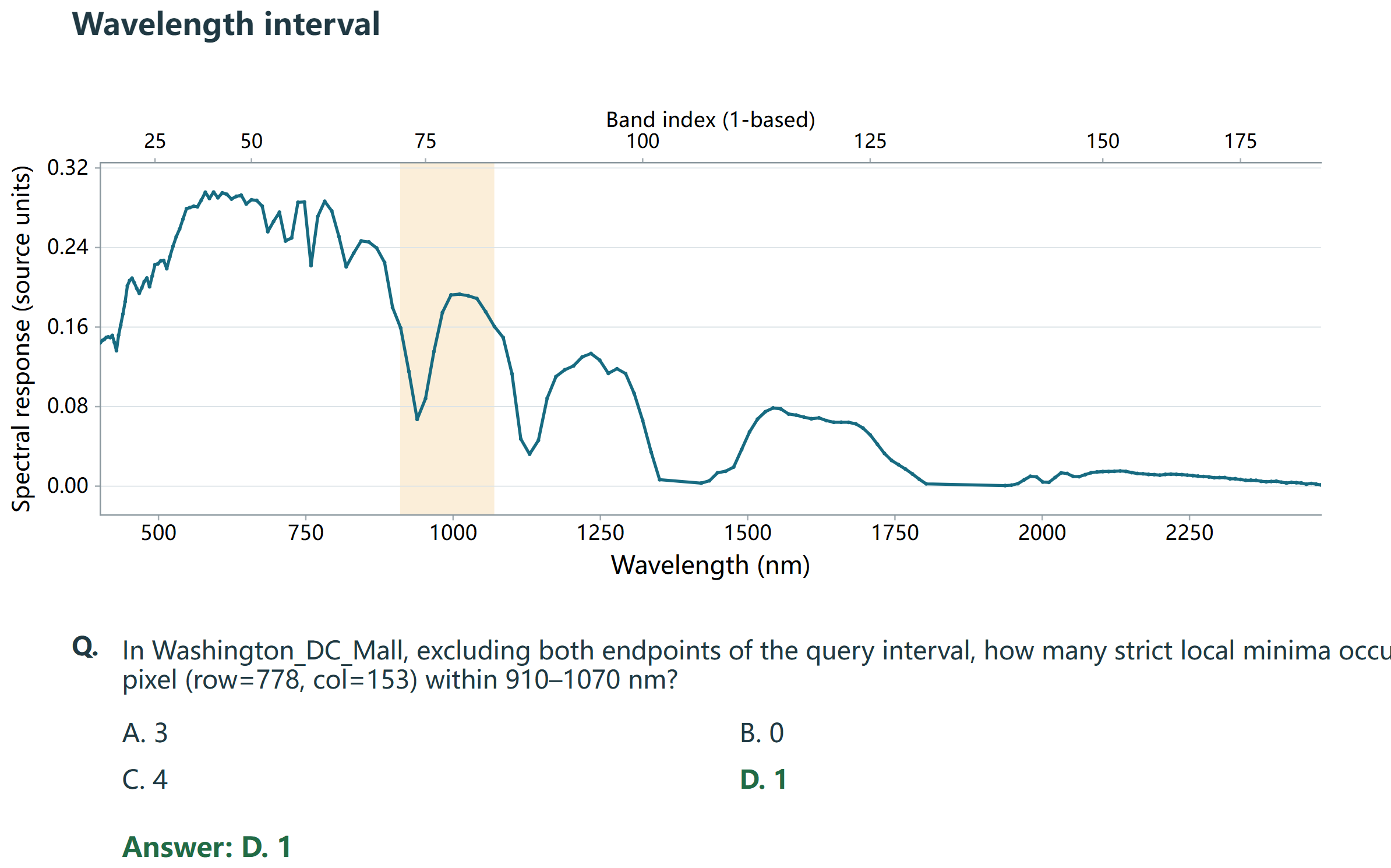}
\clearpage
\HyperGPTQualitativePair{Pixel classification}{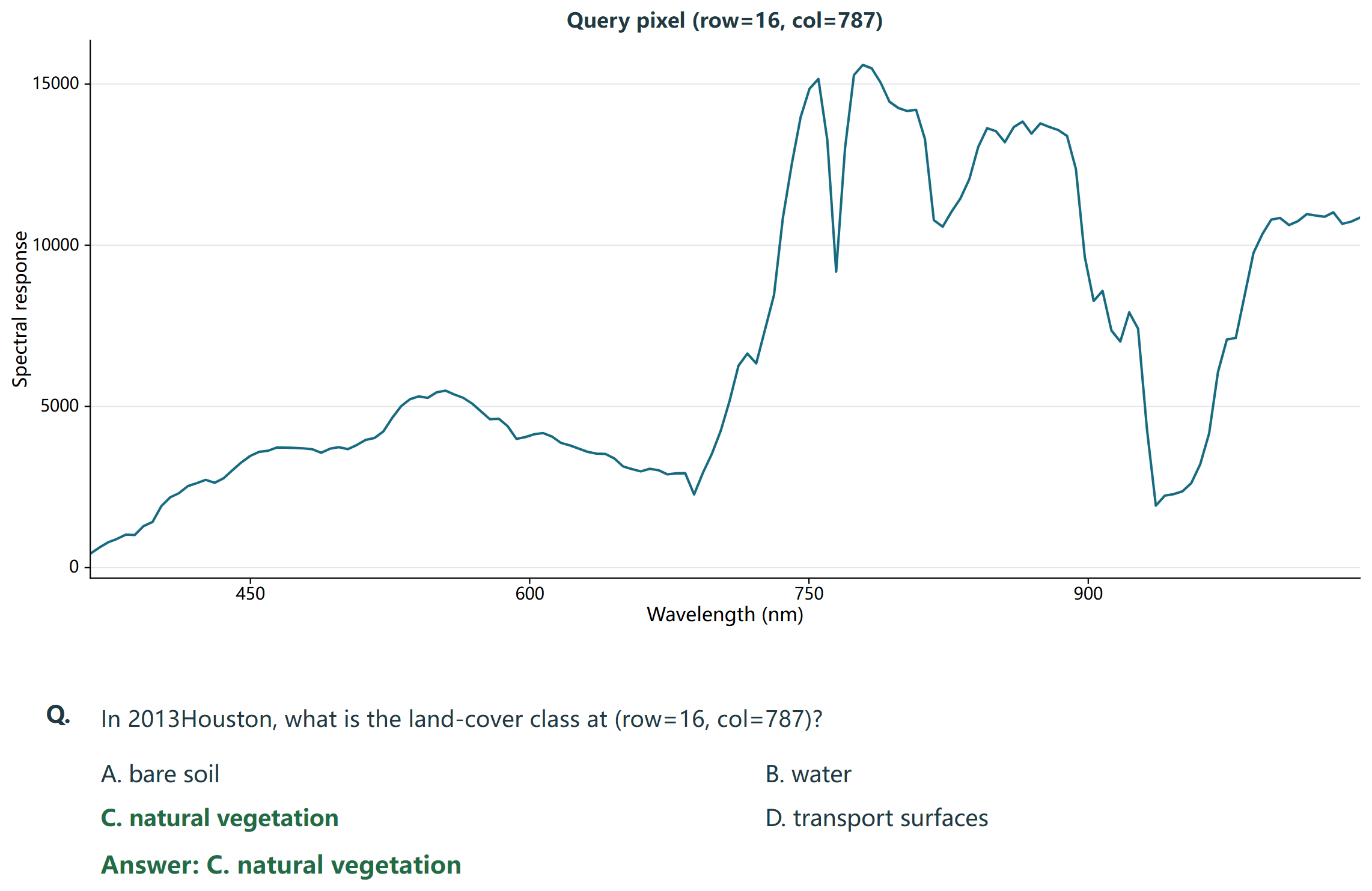}{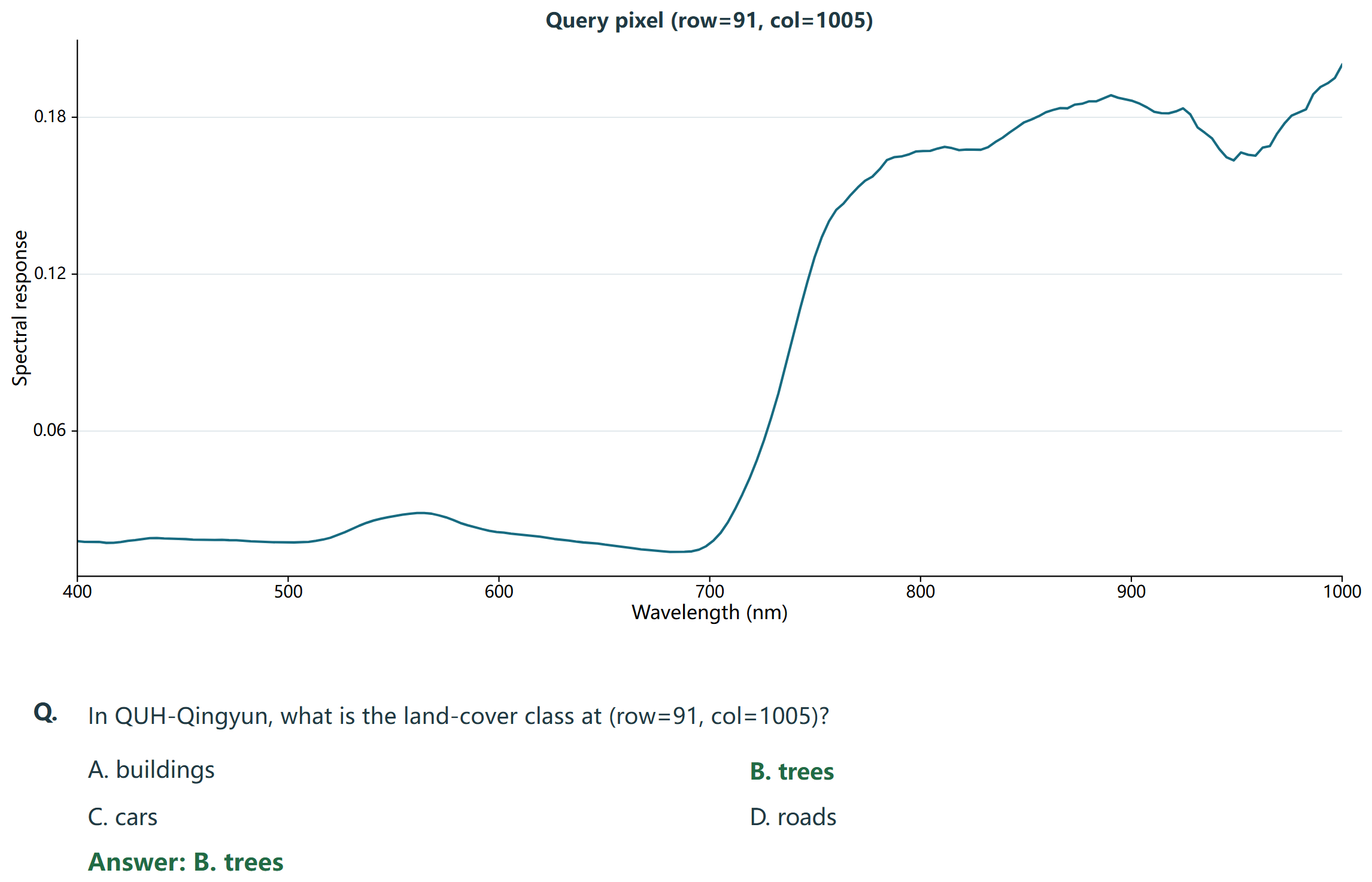}
\clearpage
\HyperGPTQualitativePair{Class-guided pixel retrieval}{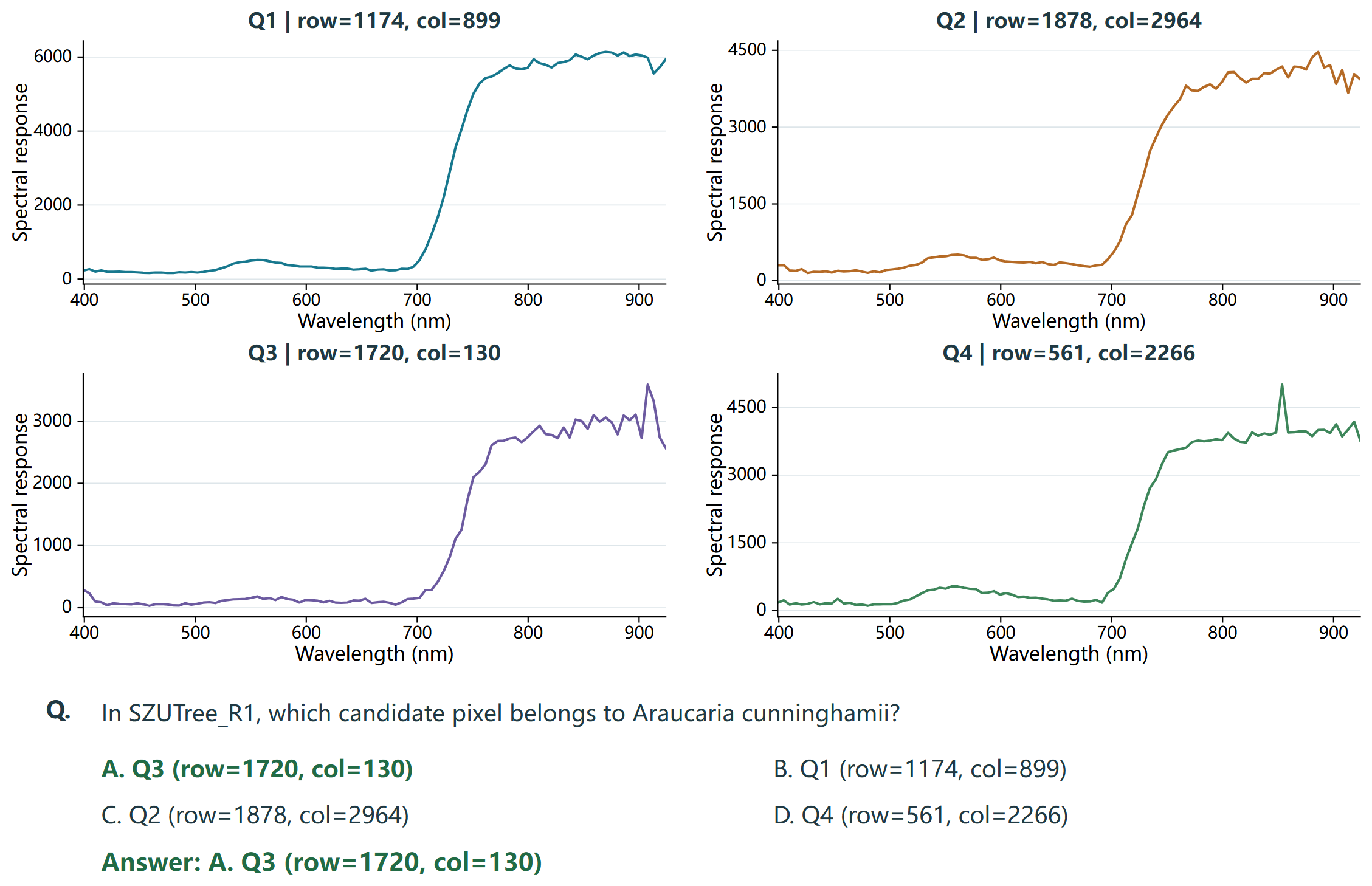}{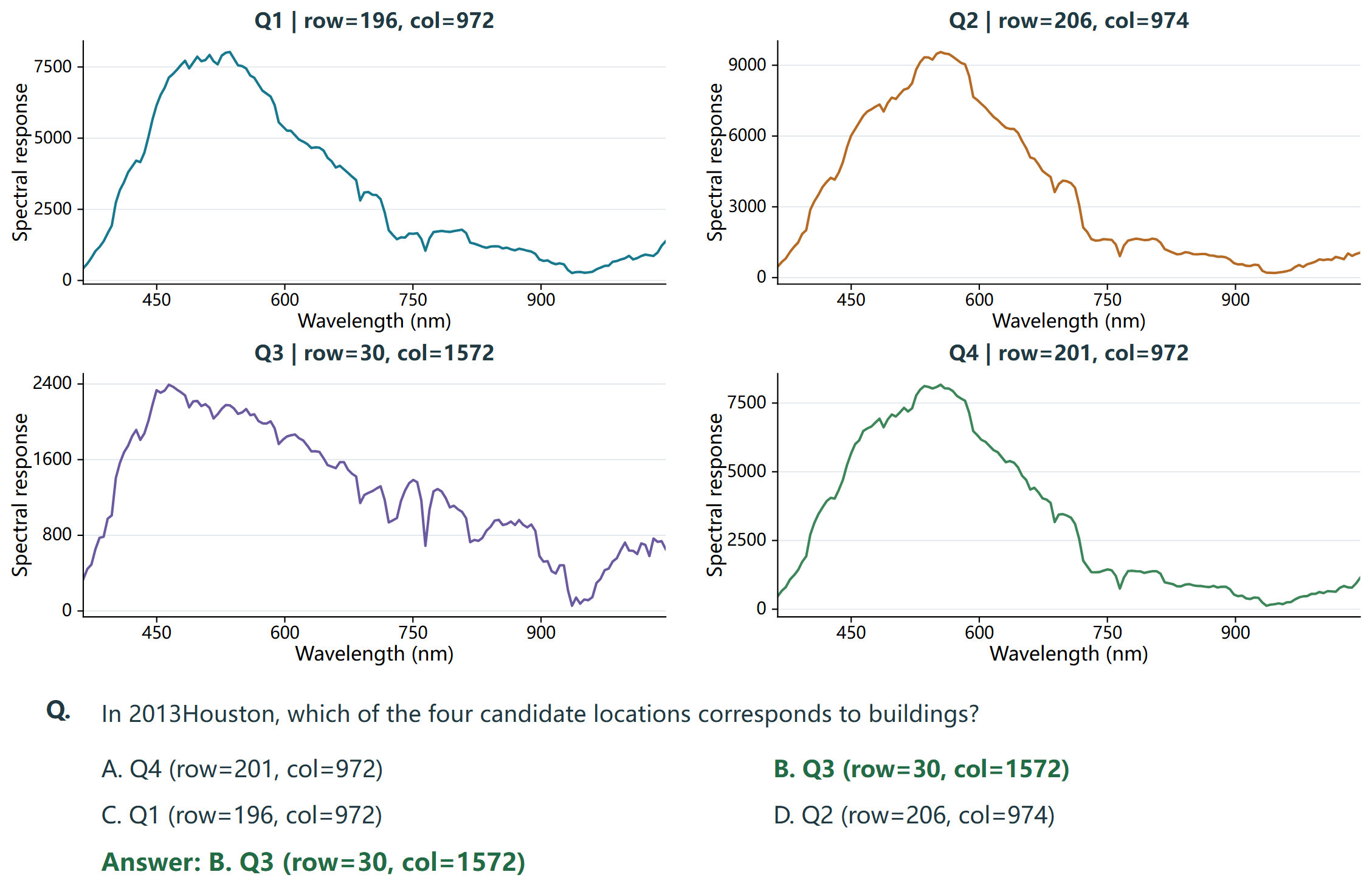}
\clearpage
\HyperGPTQualitativePair{Same-class matching}{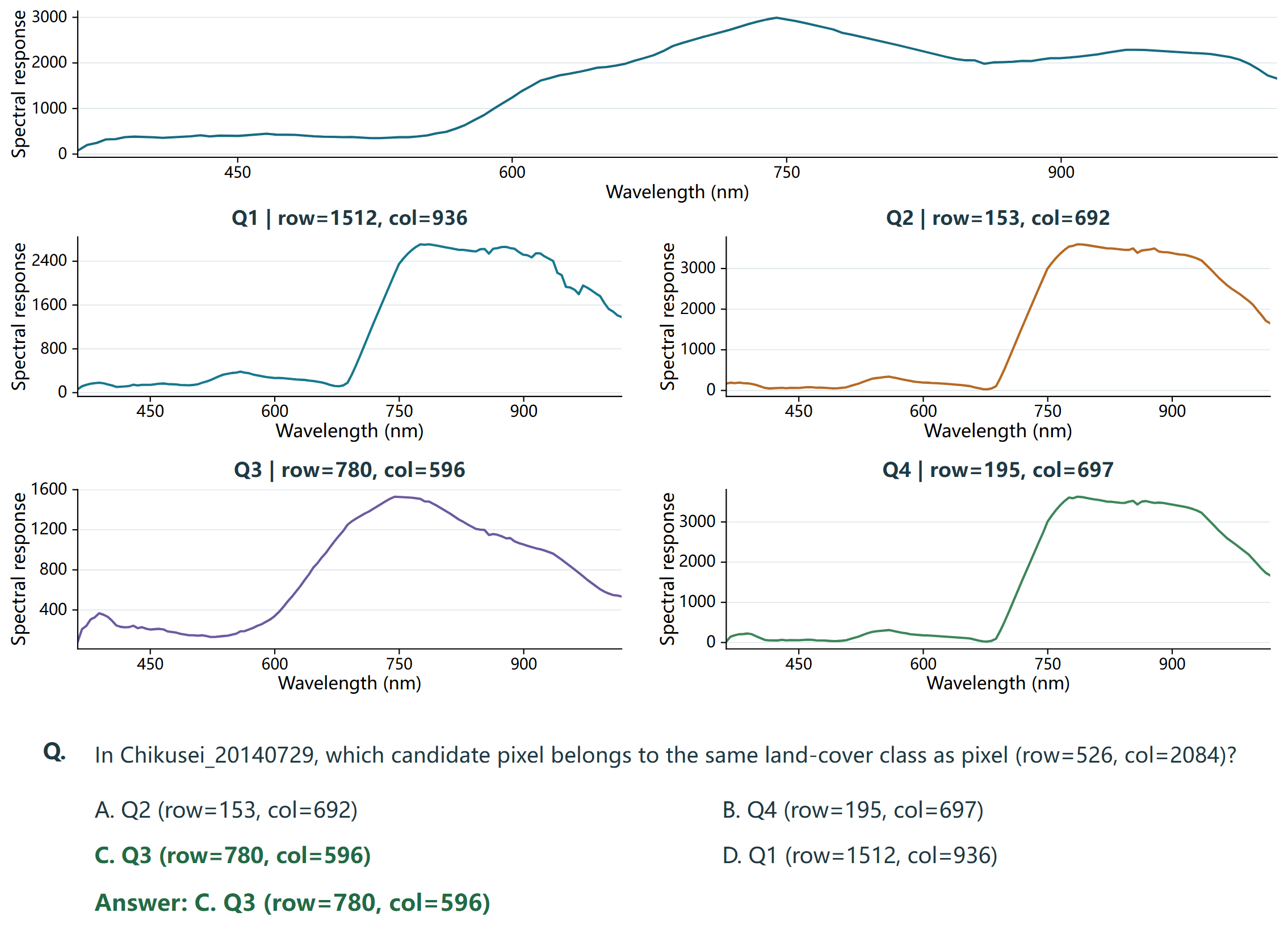}{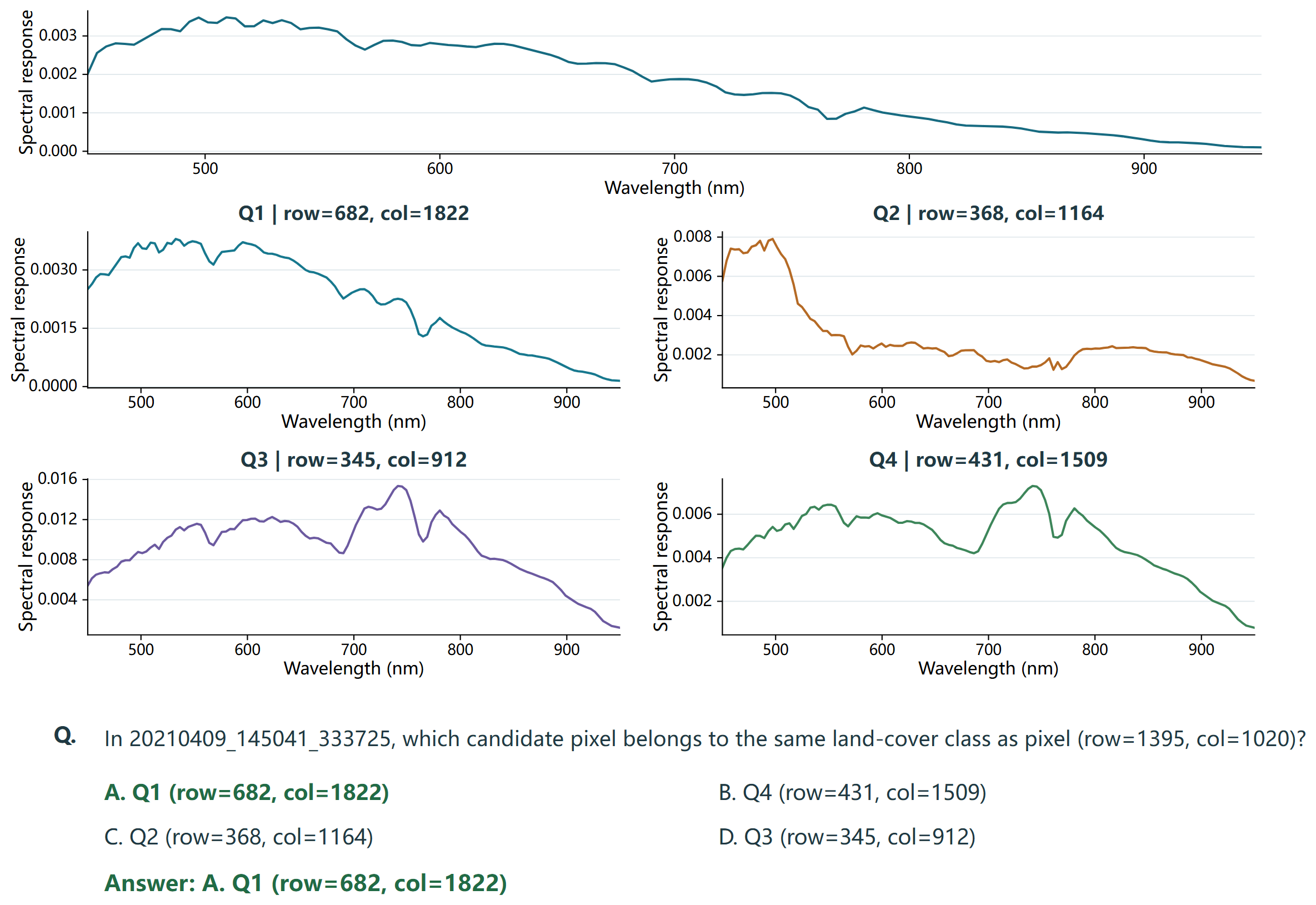}
\clearpage
\HyperGPTQualitativePair{Support-guided classification}{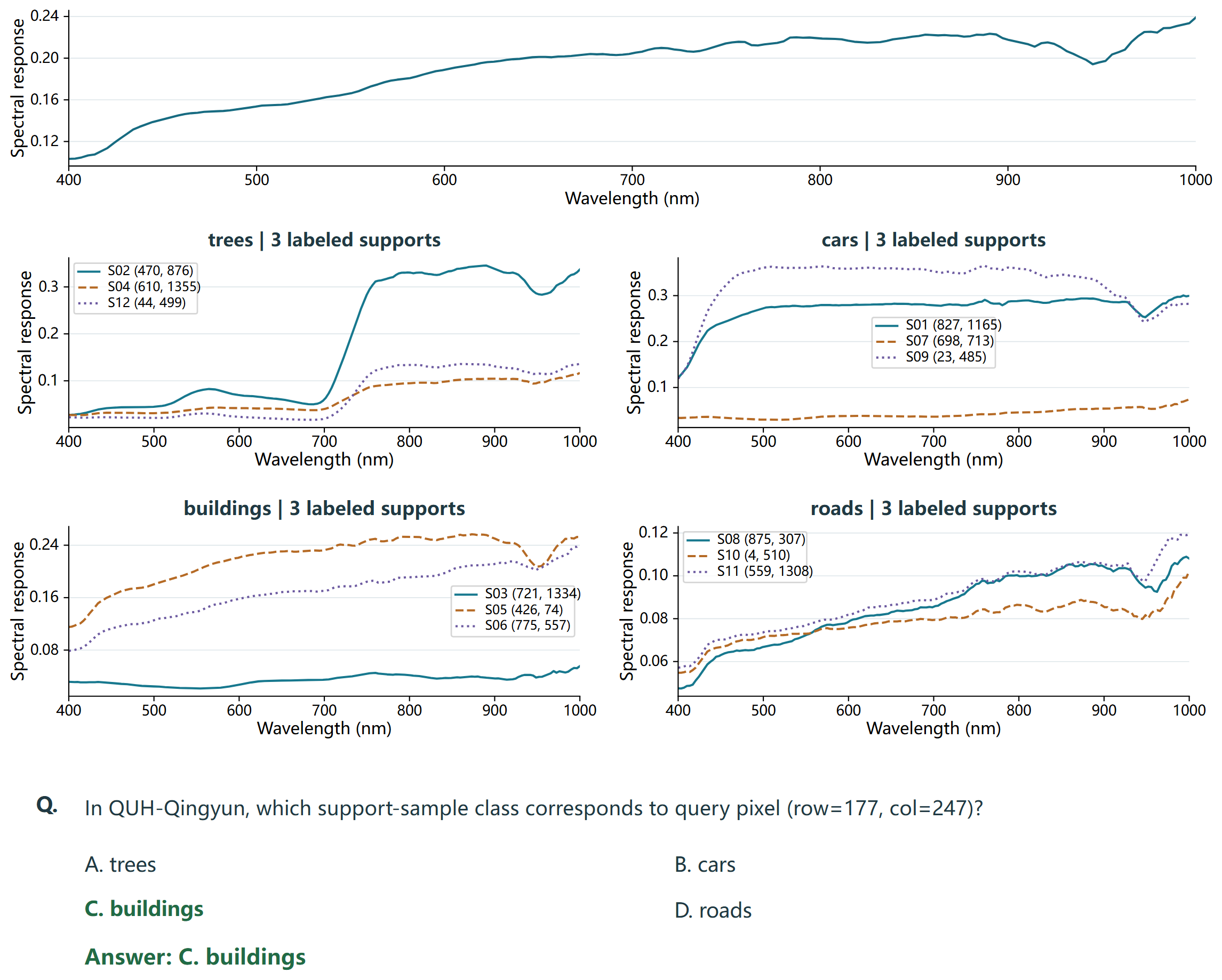}{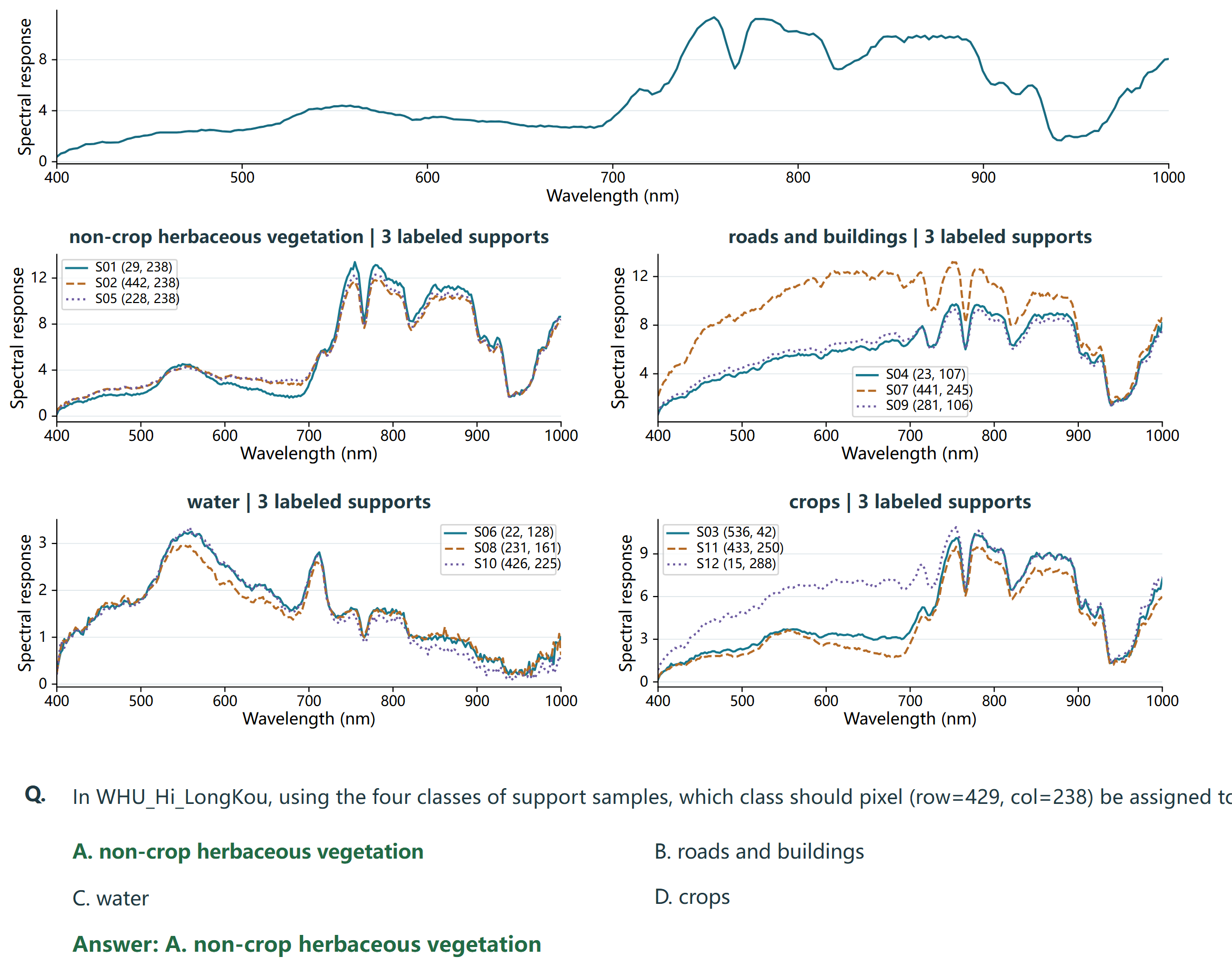}
\clearpage
\HyperGPTQualitativePair{Regional composition}{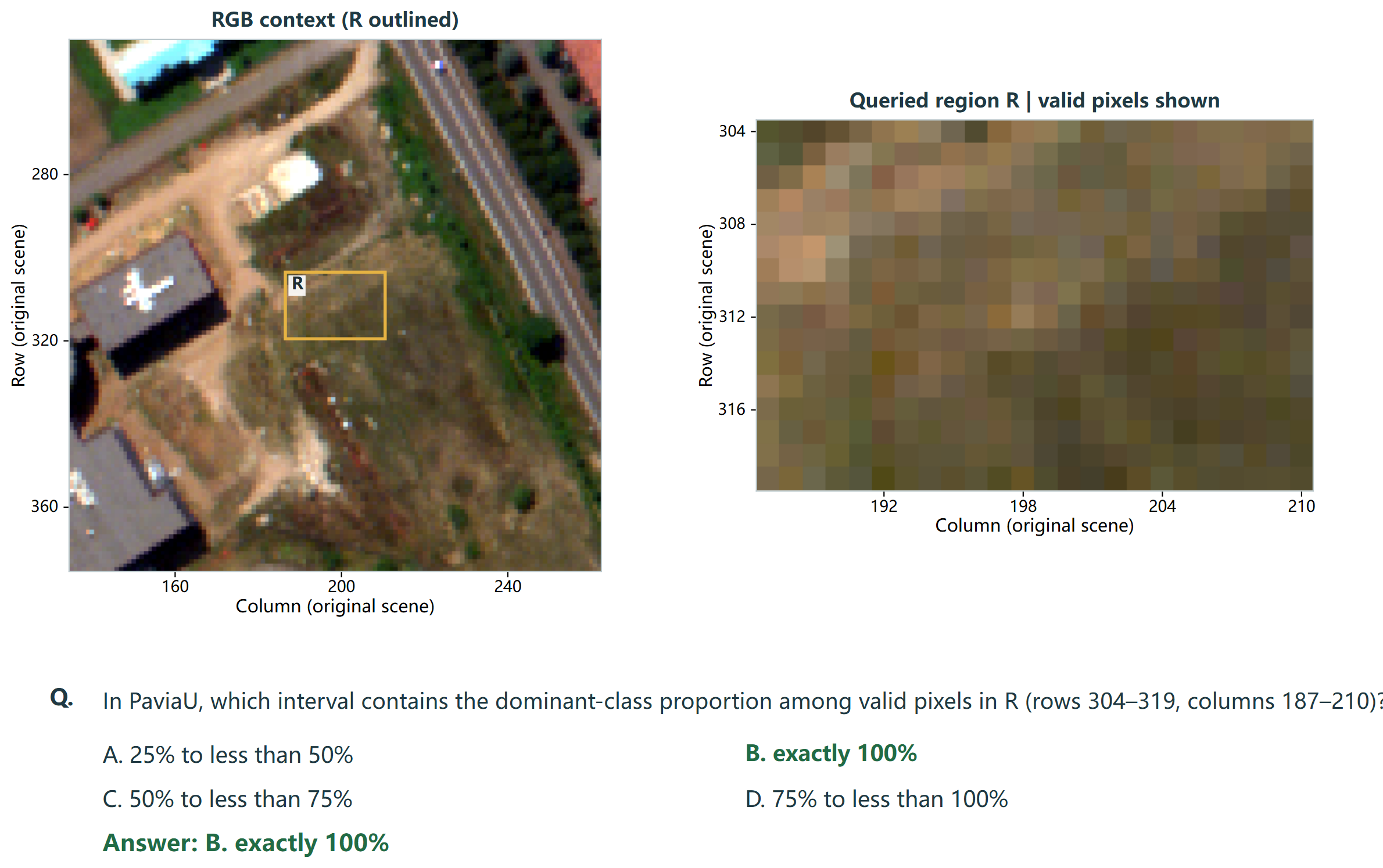}{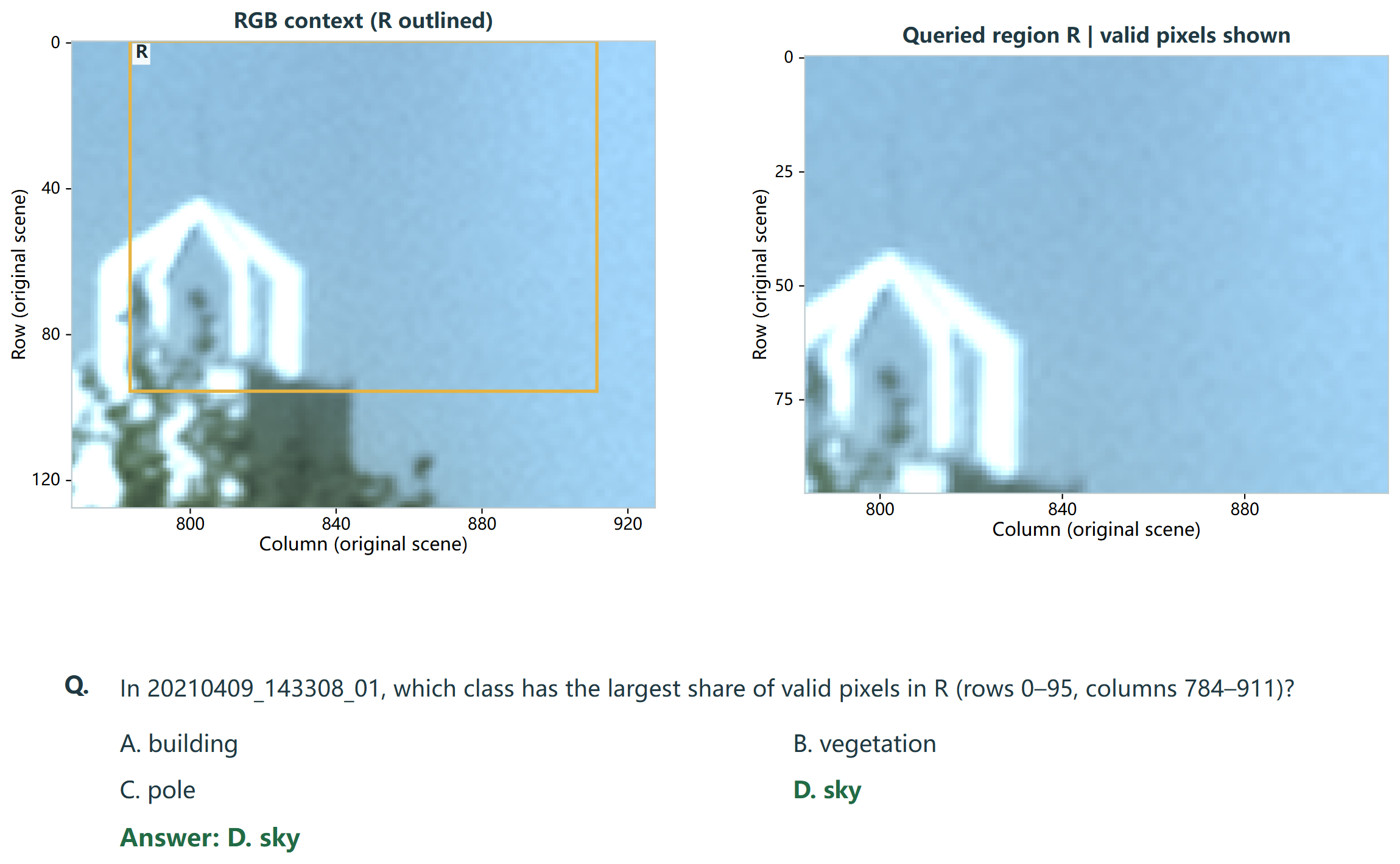}
\clearpage
\HyperGPTQualitativePair{Class spatial relations}{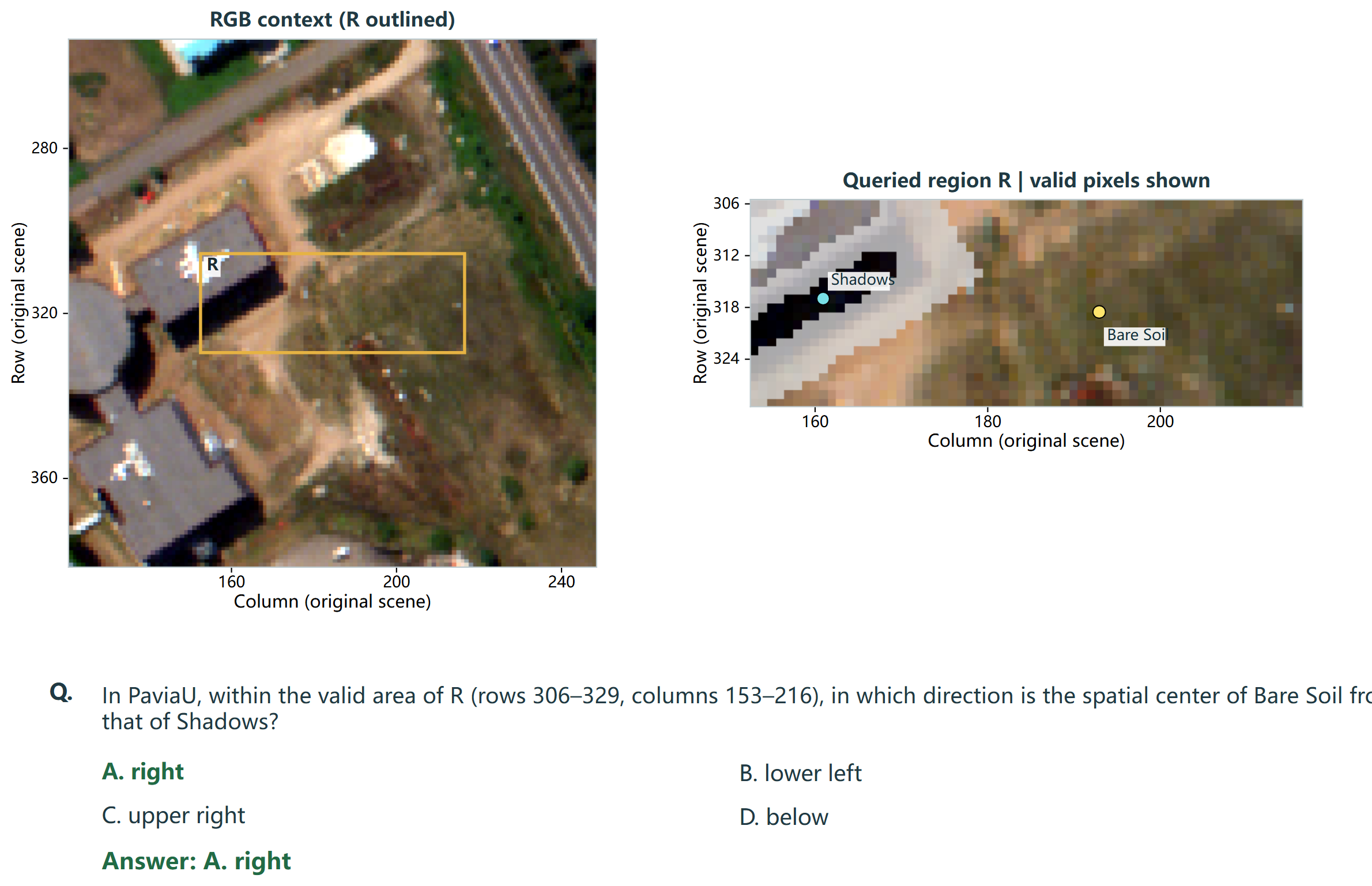}{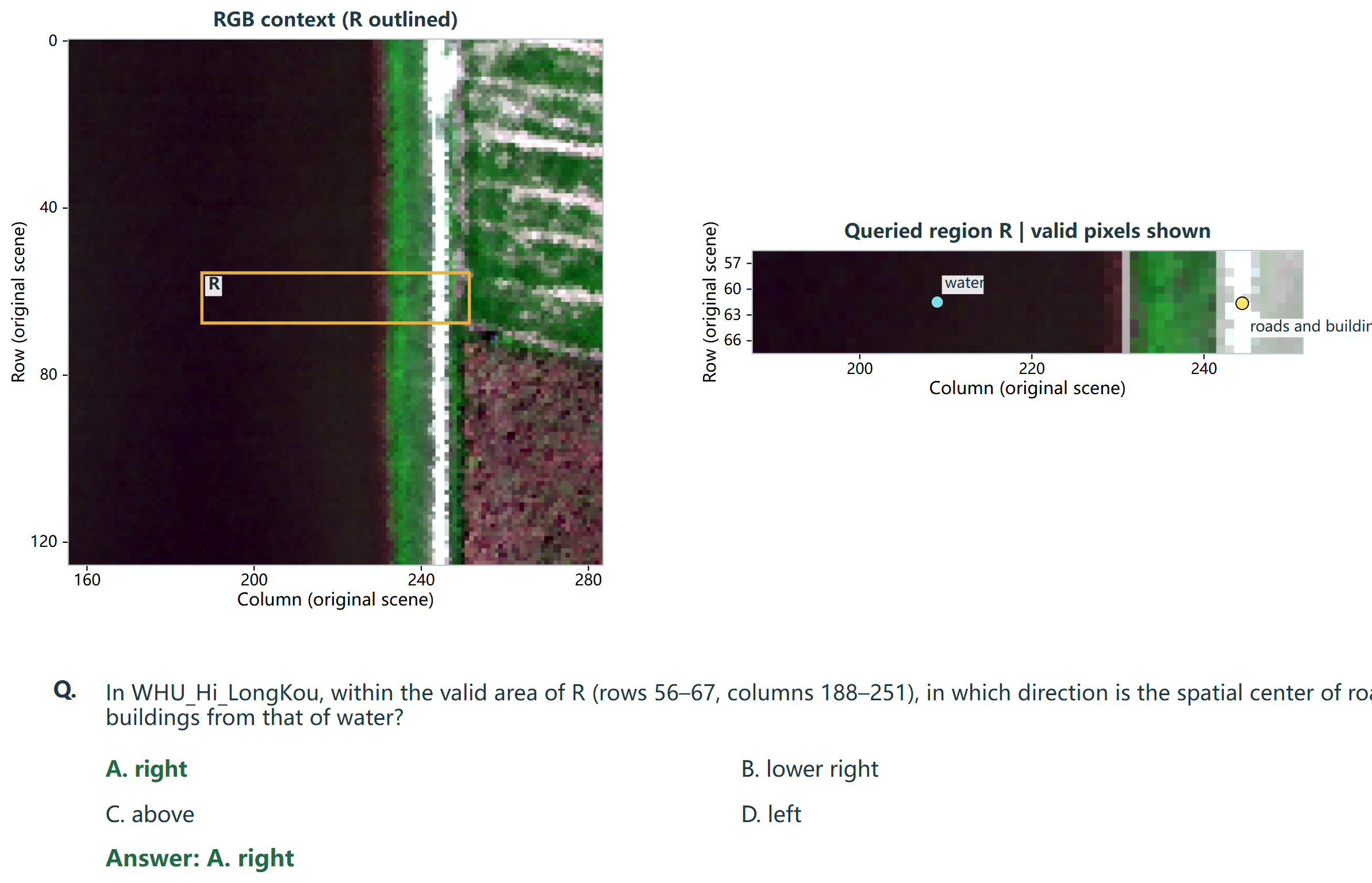}
\clearpage
\HyperGPTQualitativePair{Class-guided region retrieval}{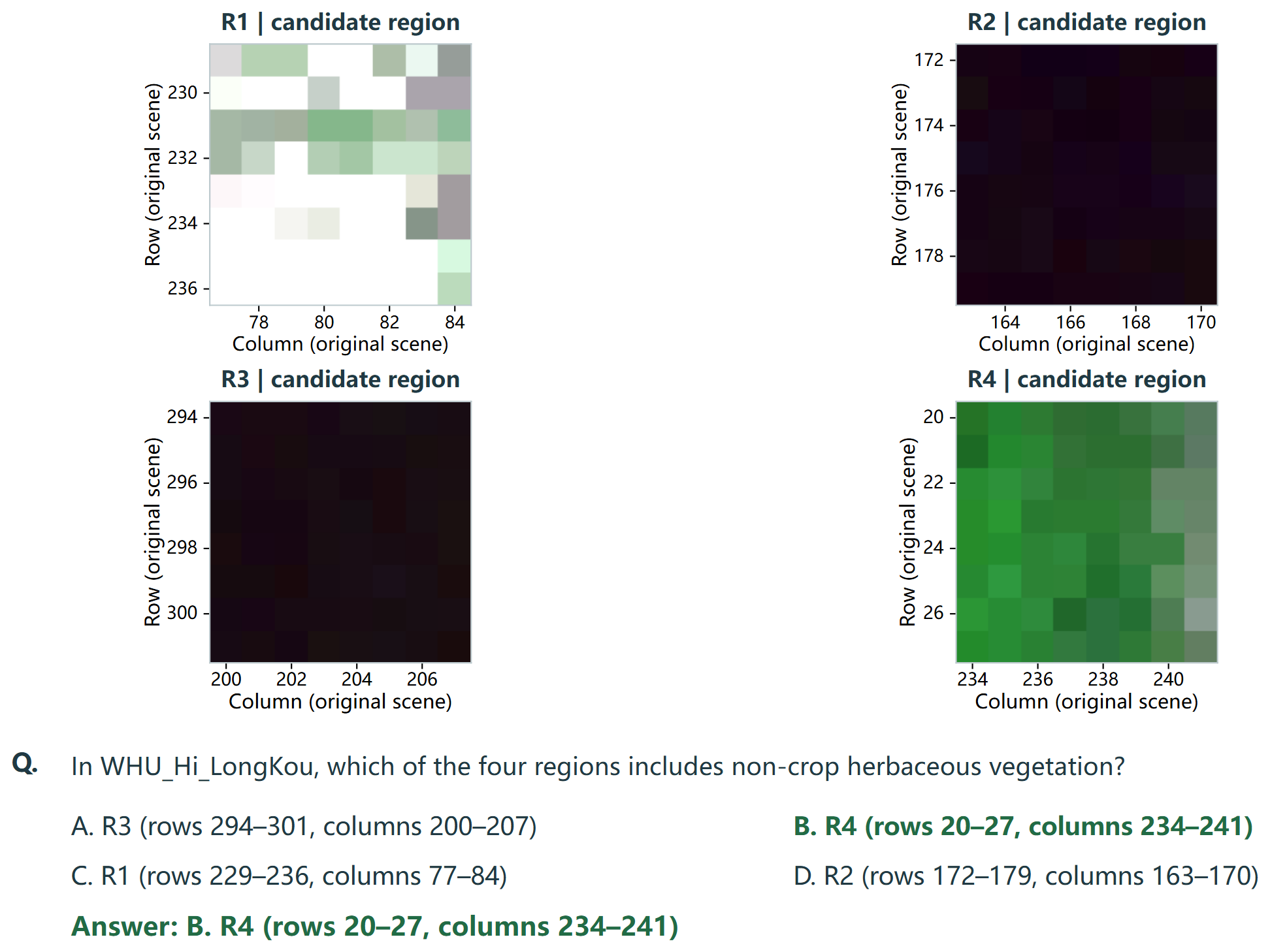}{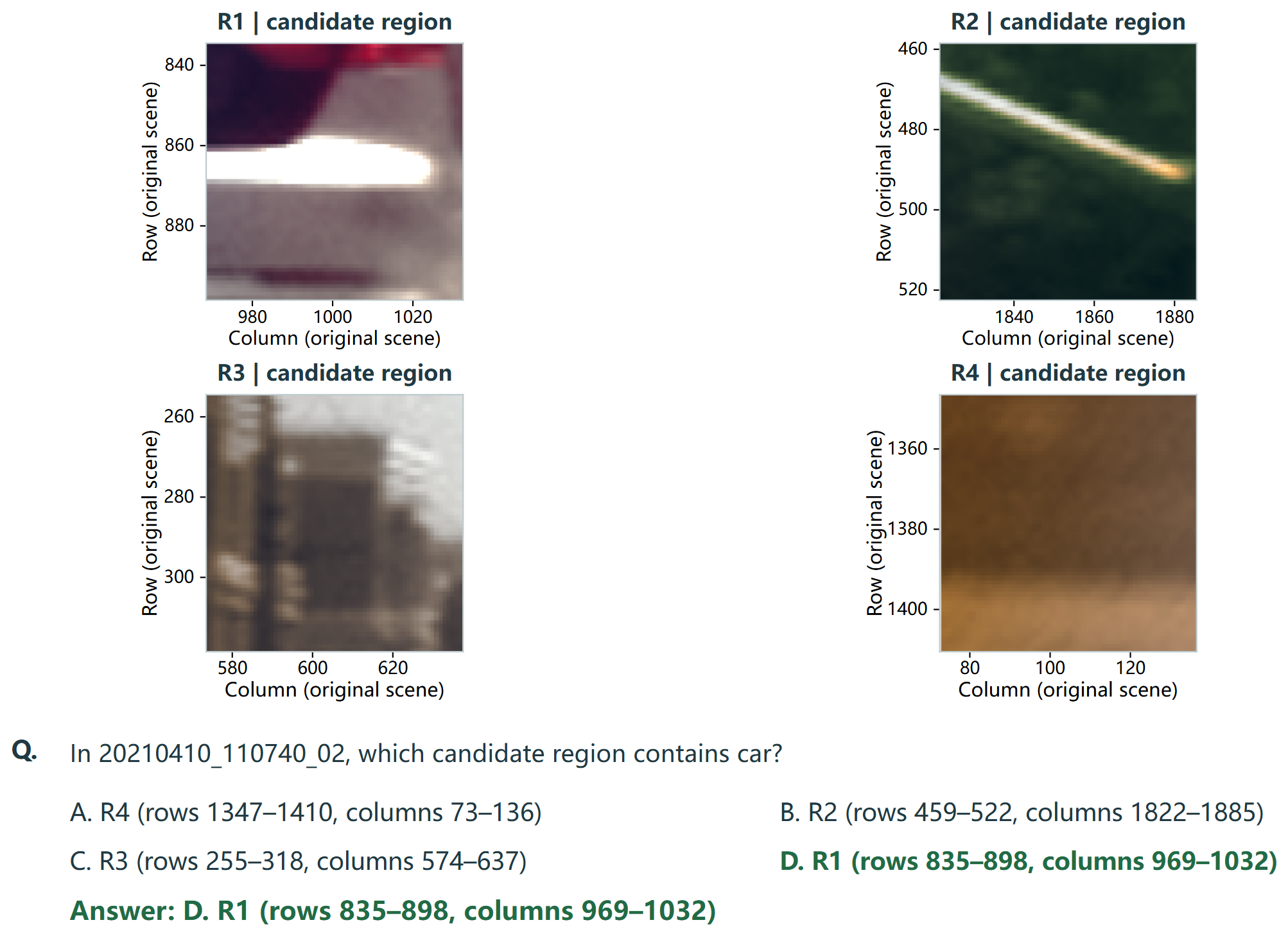}
\clearpage
\HyperGPTQualitativePair{Regional count comparison}{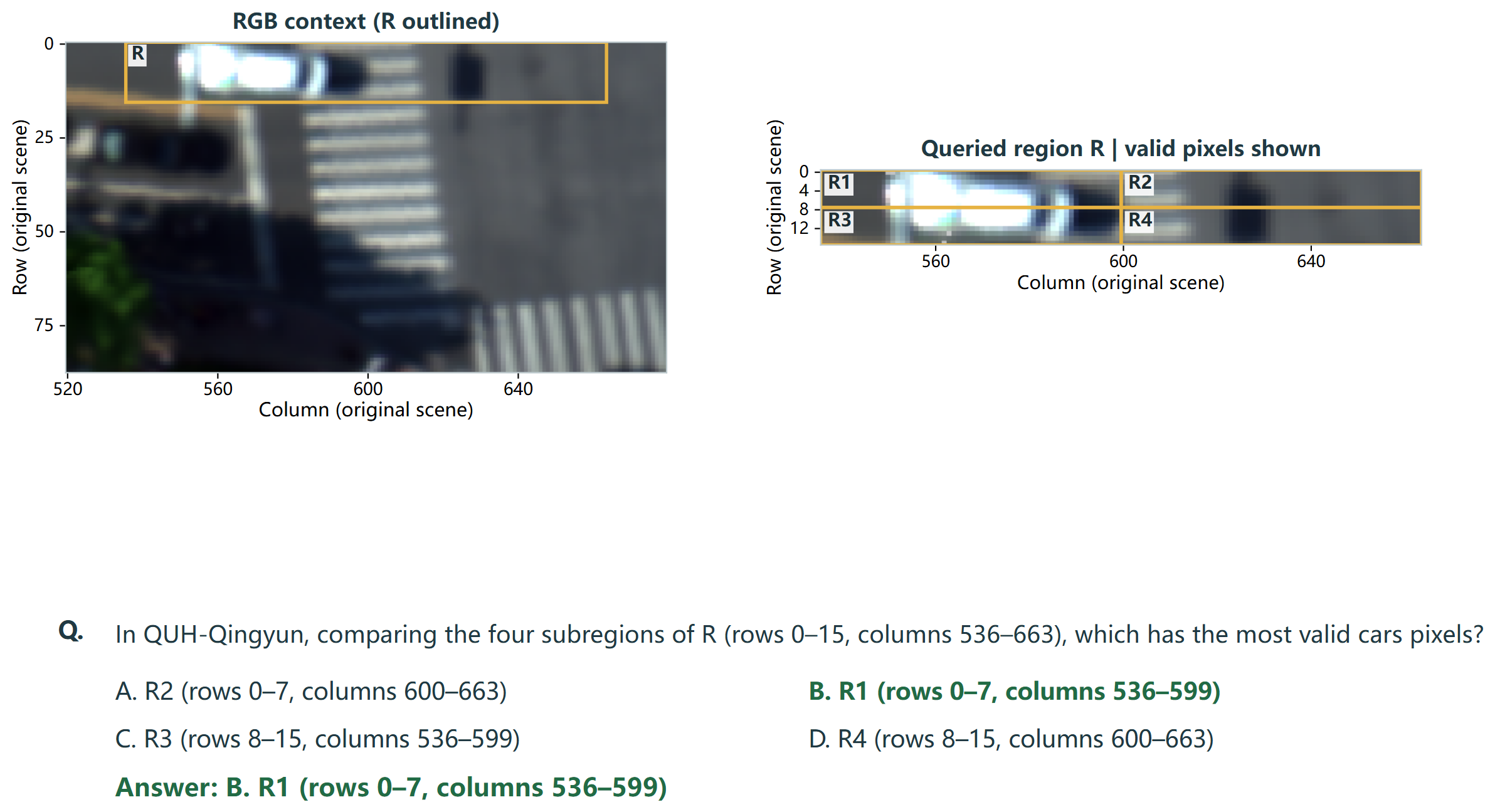}{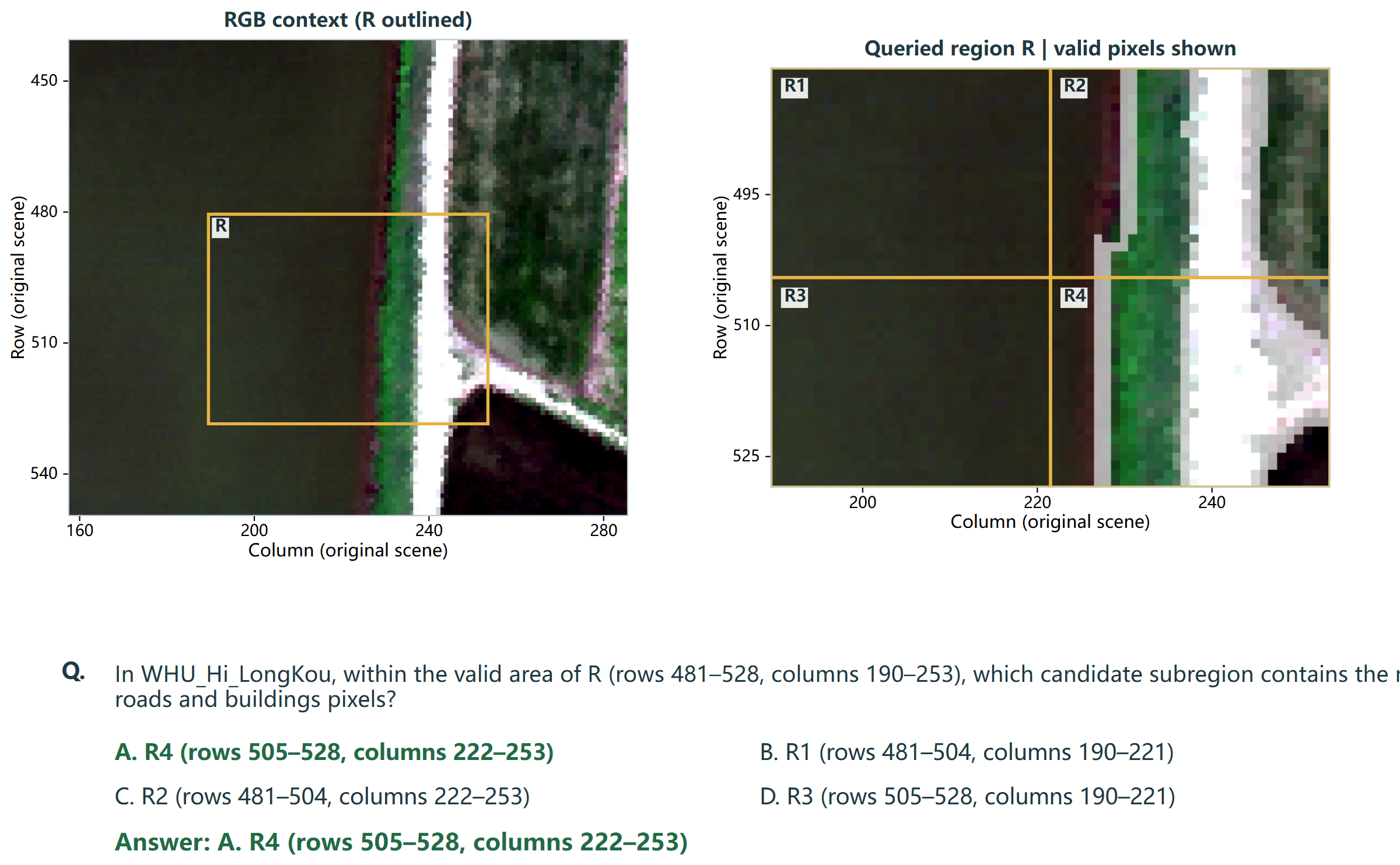}
\clearpage
\HyperGPTQualitativePair{Global counting}{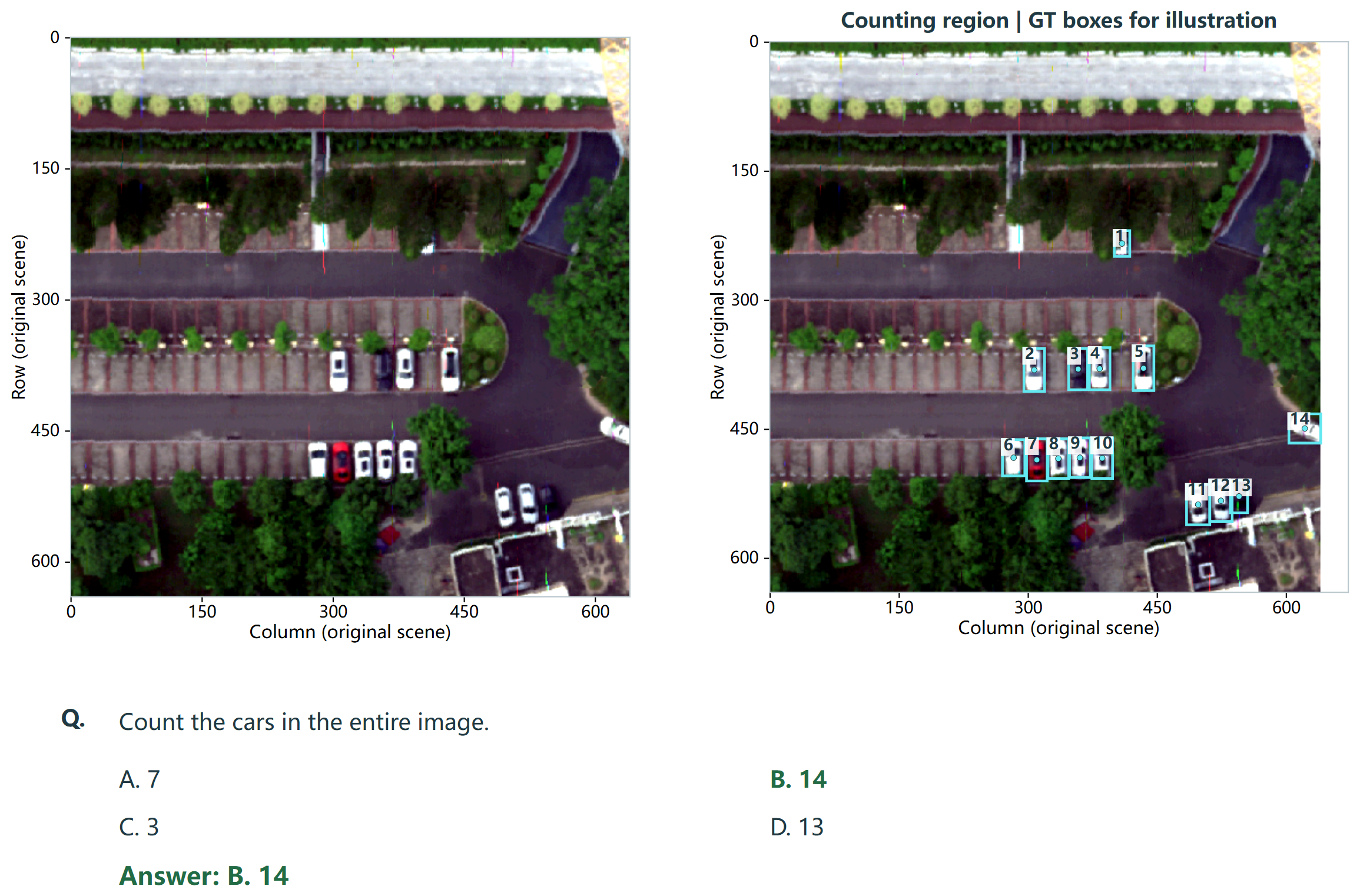}{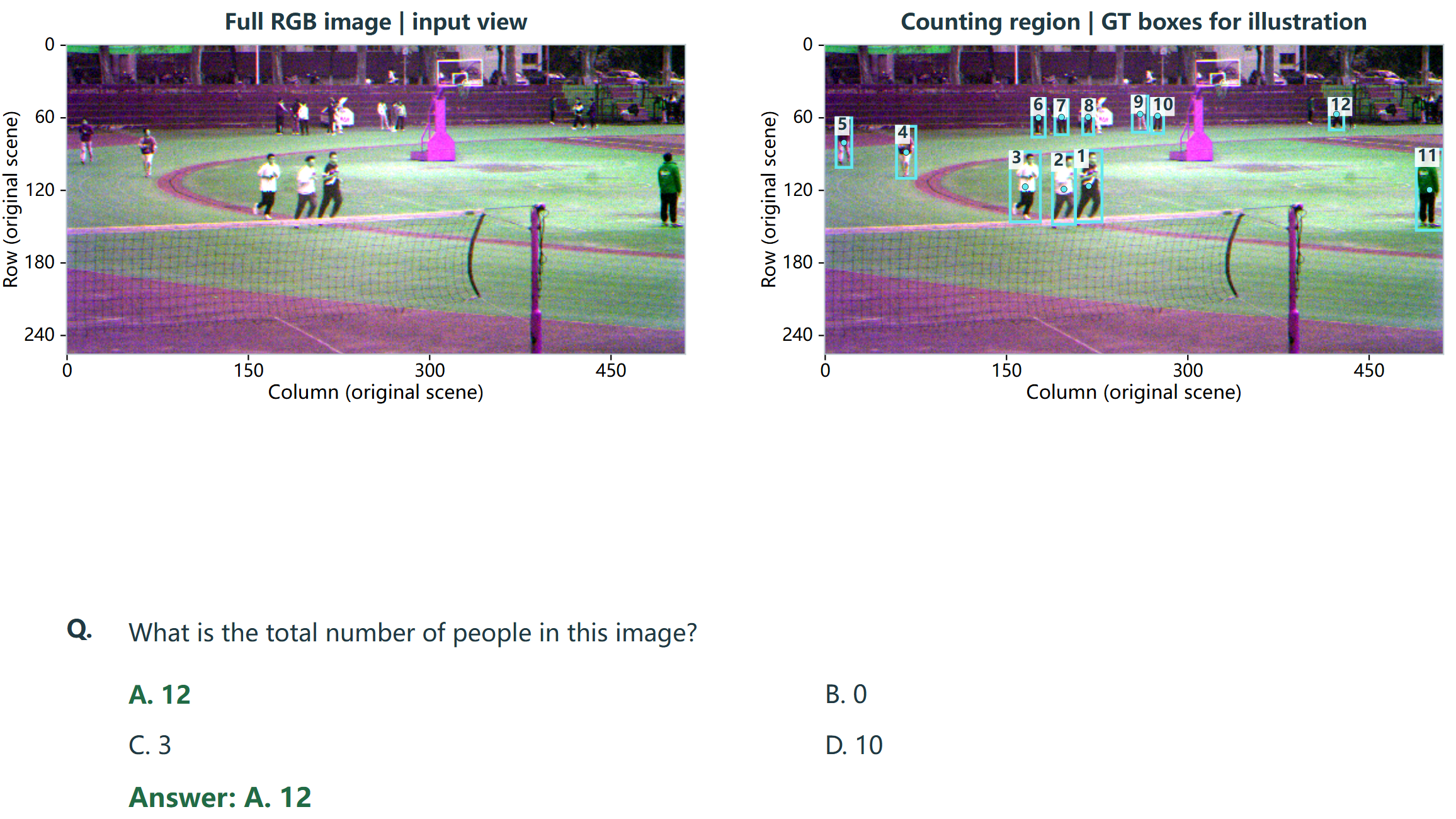}
\clearpage
\HyperGPTQualitativePair{Regional counting}{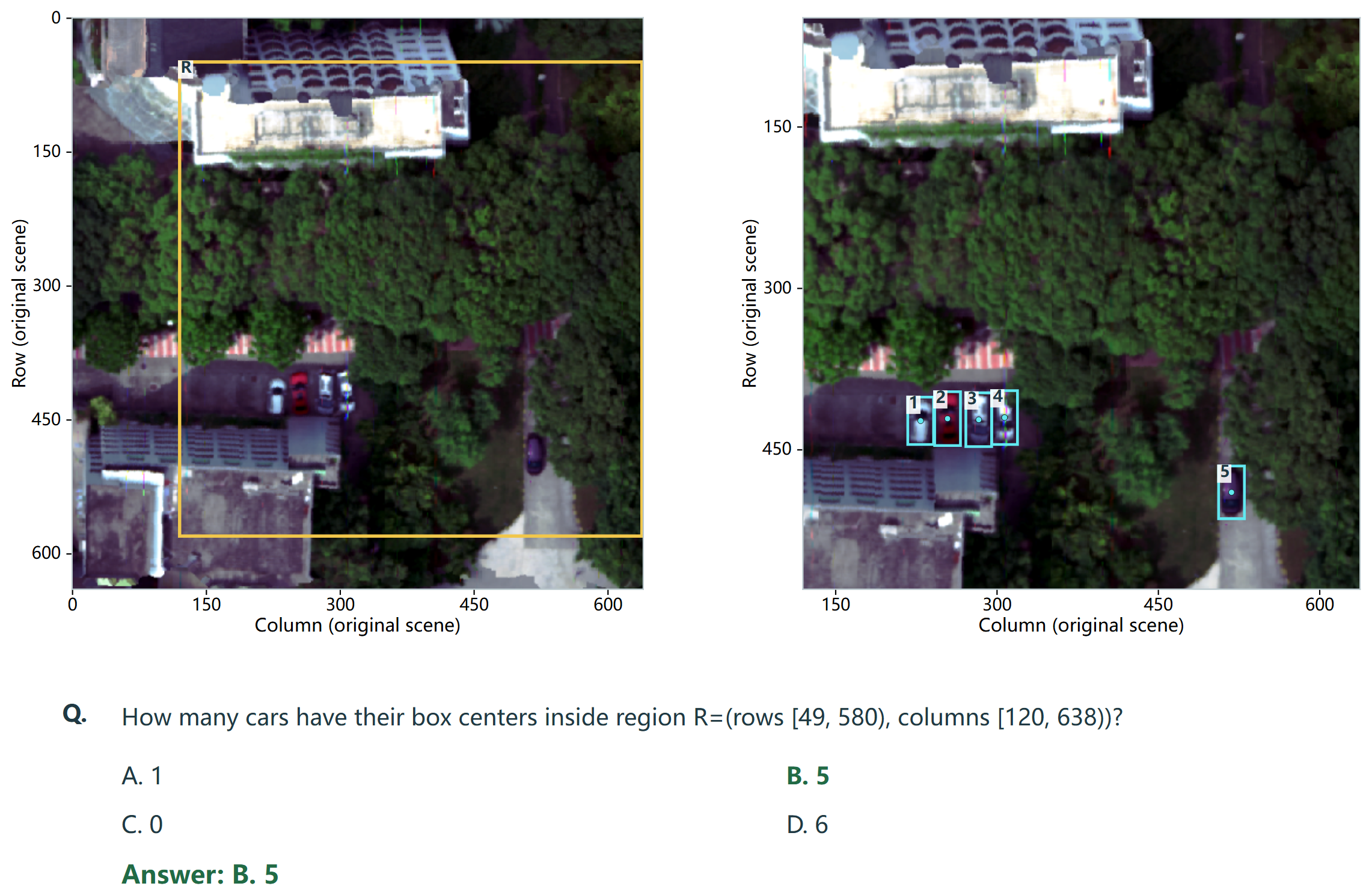}{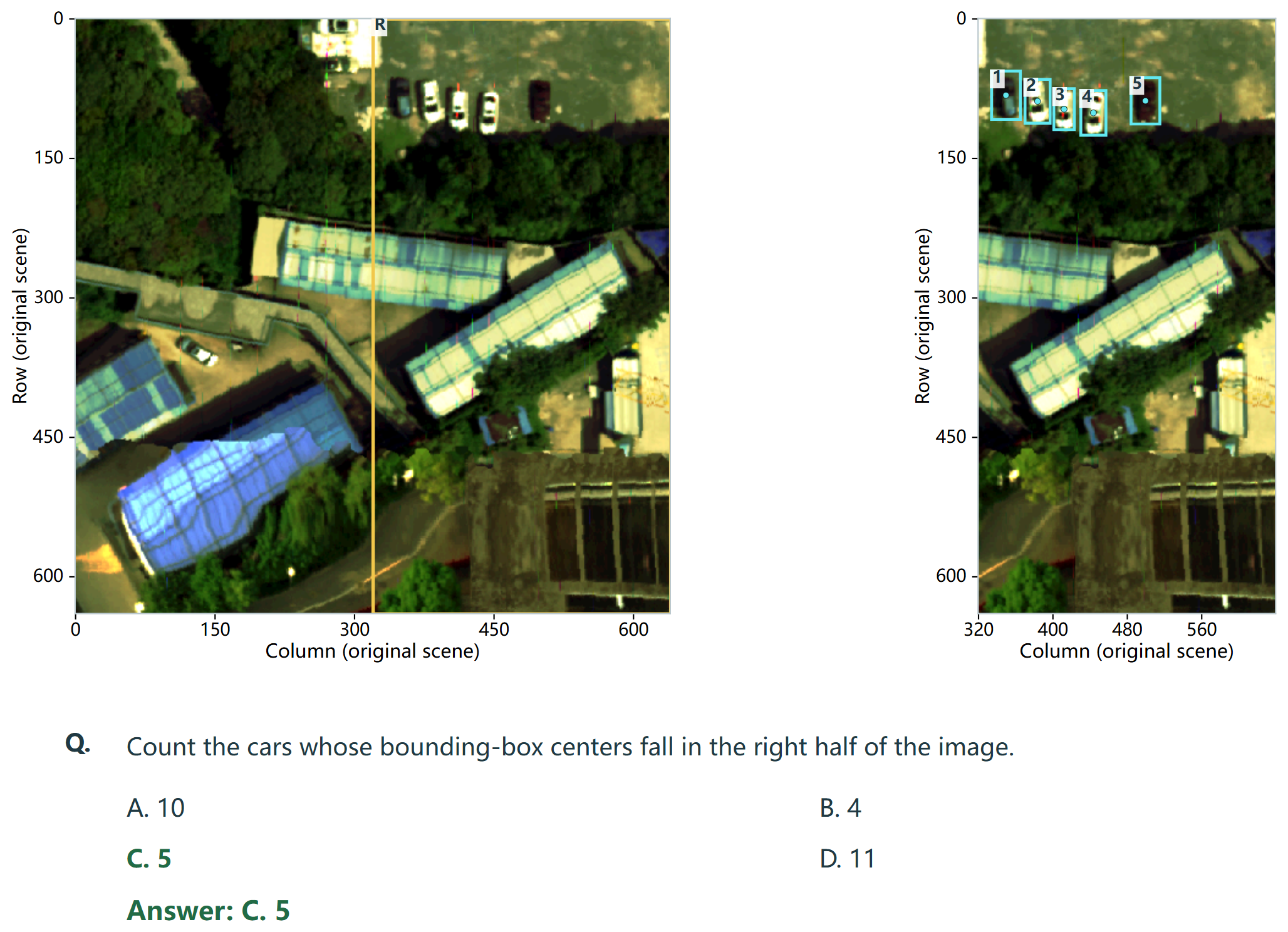}
\clearpage
\HyperGPTQualitativePair{Single-object localization}{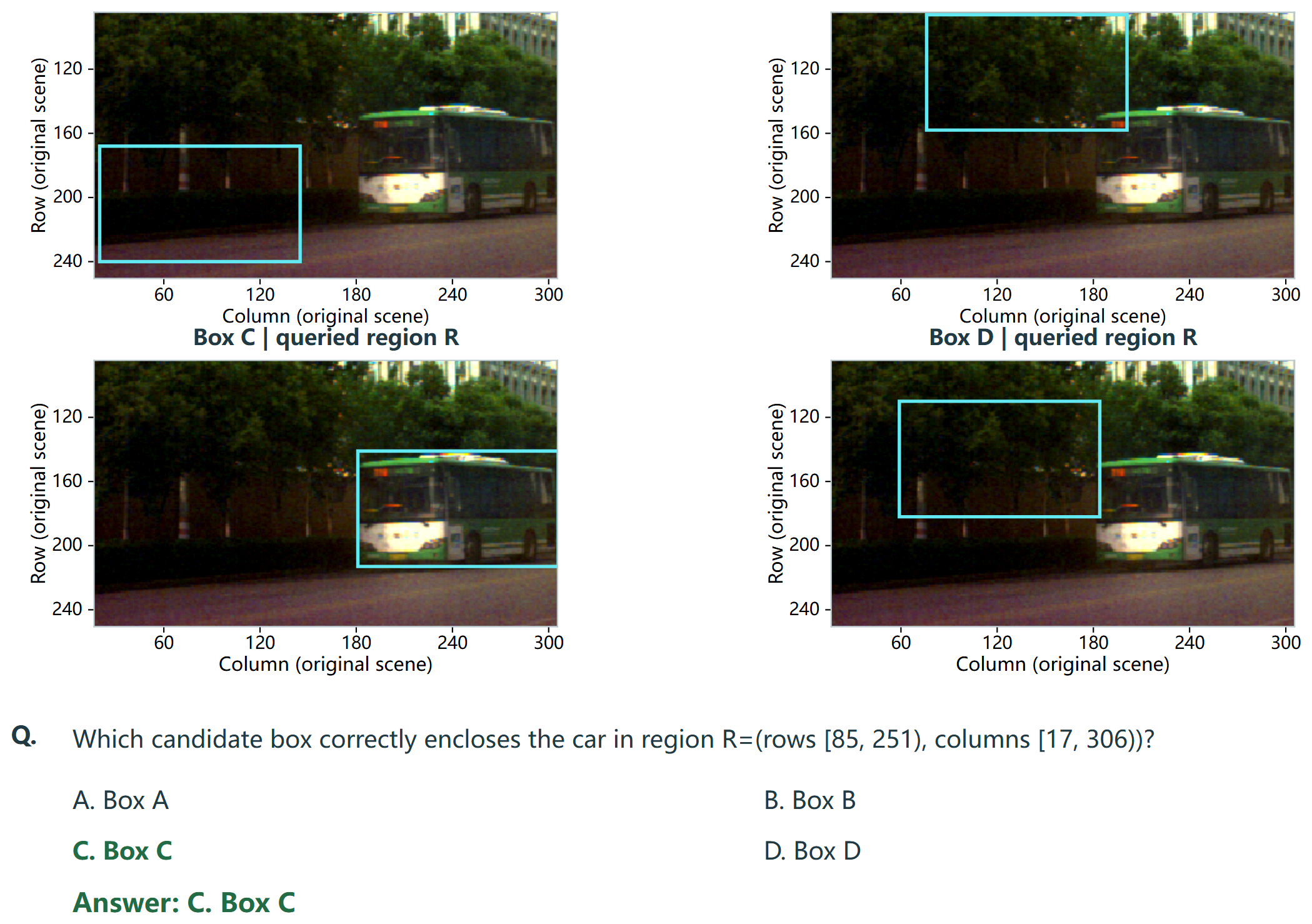}{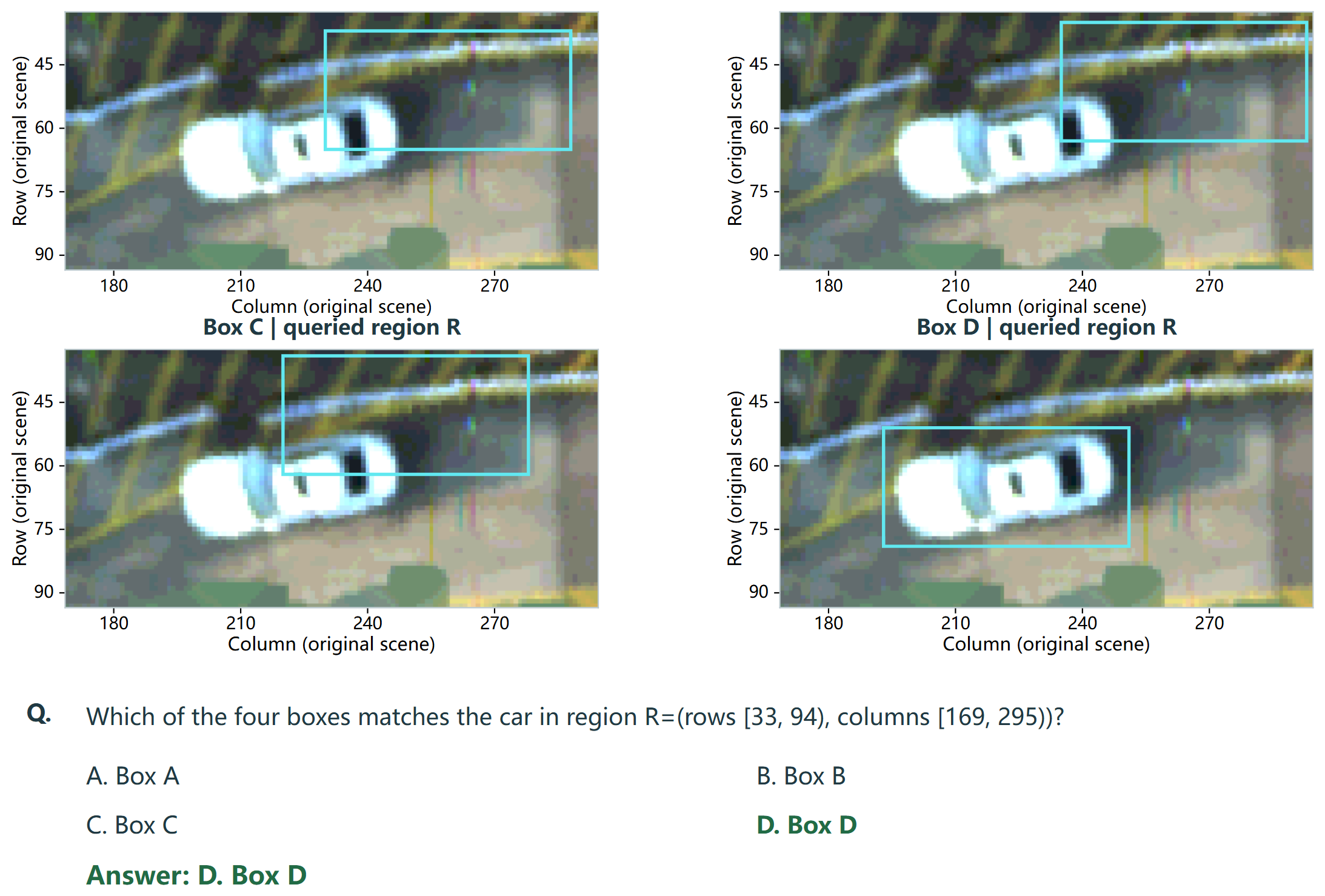}
\clearpage
\HyperGPTQualitativePair{Object-set localization}{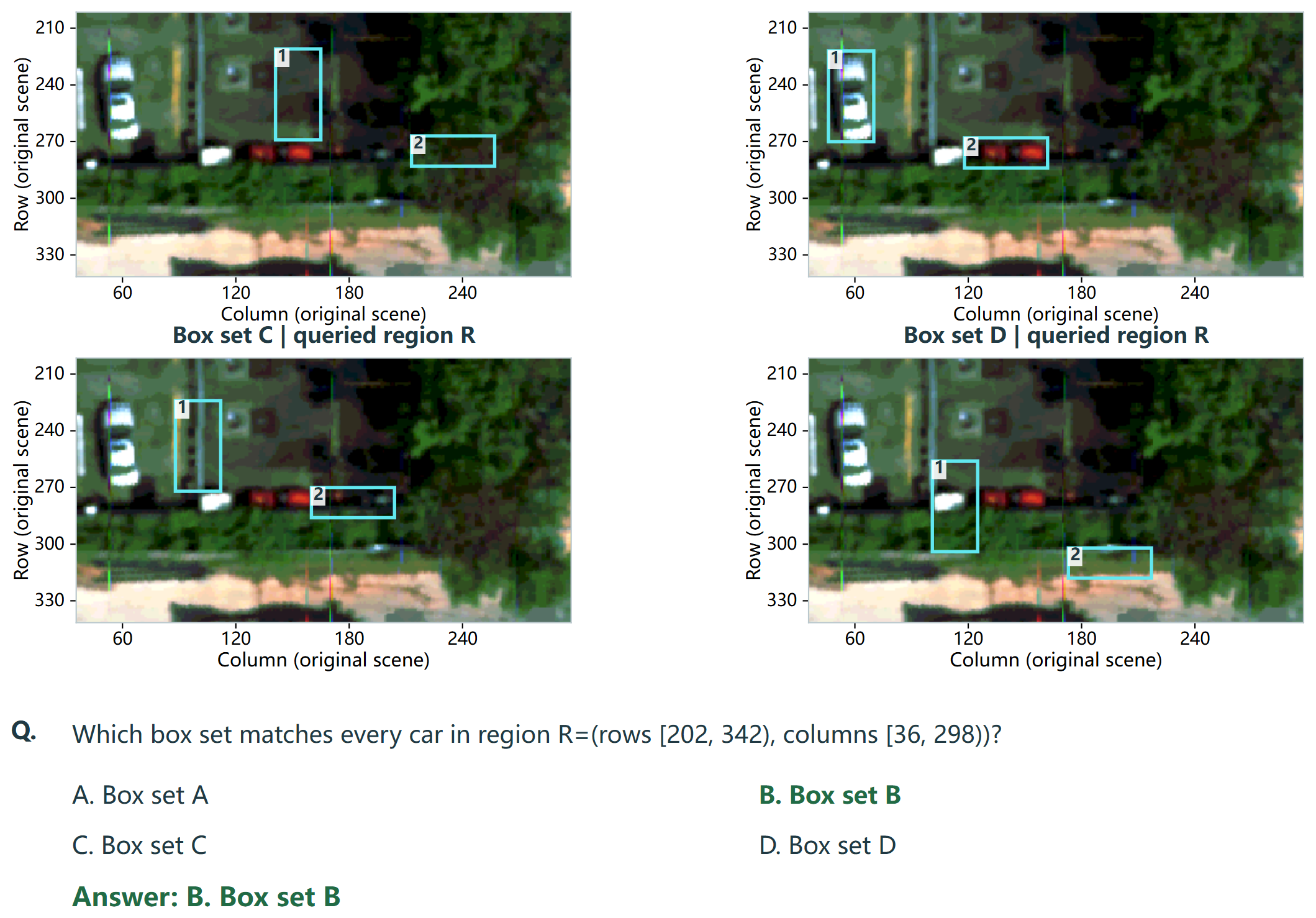}{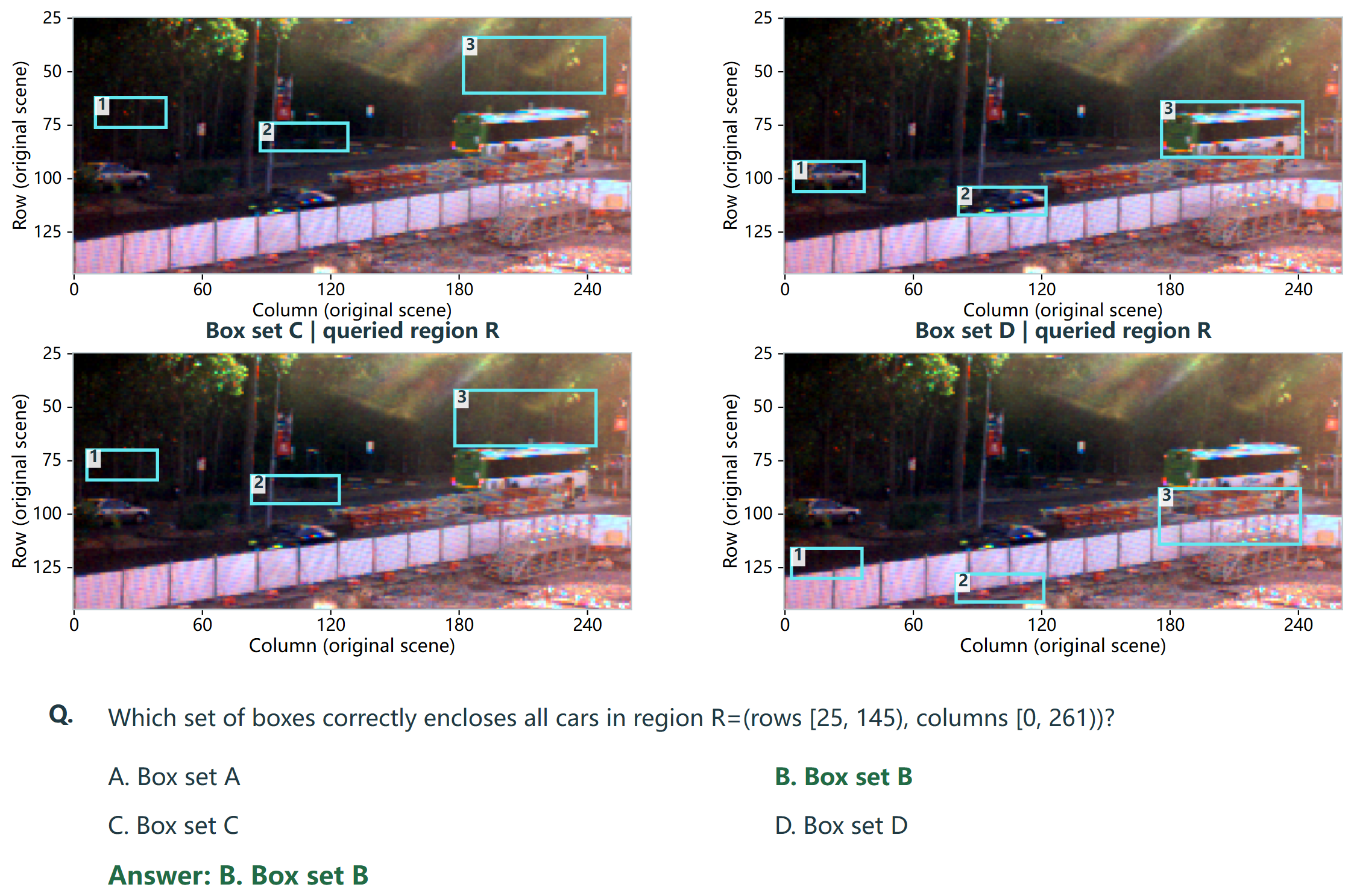}
\clearpage
\HyperGPTQualitativePair{Object spatial relations}{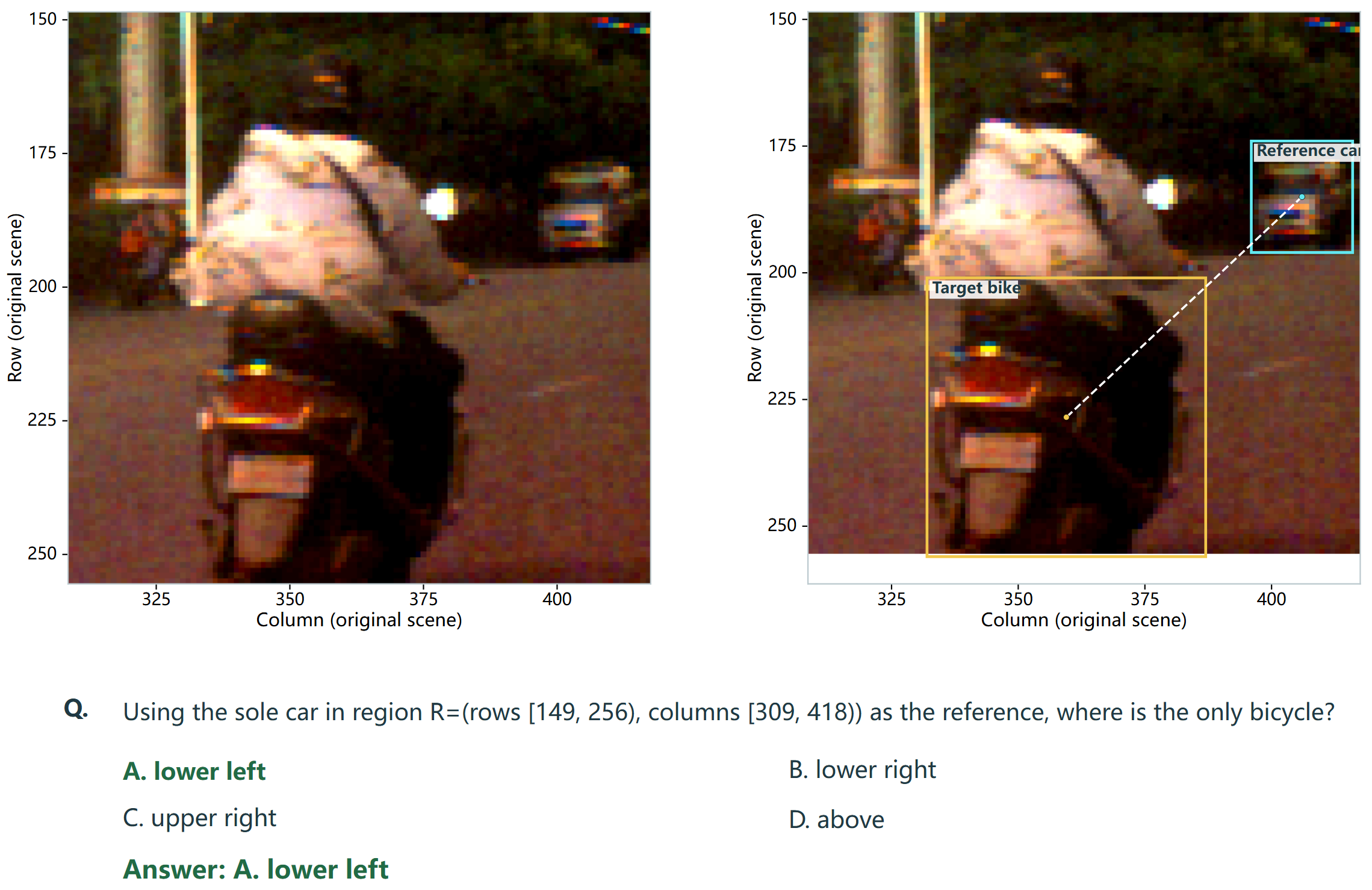}{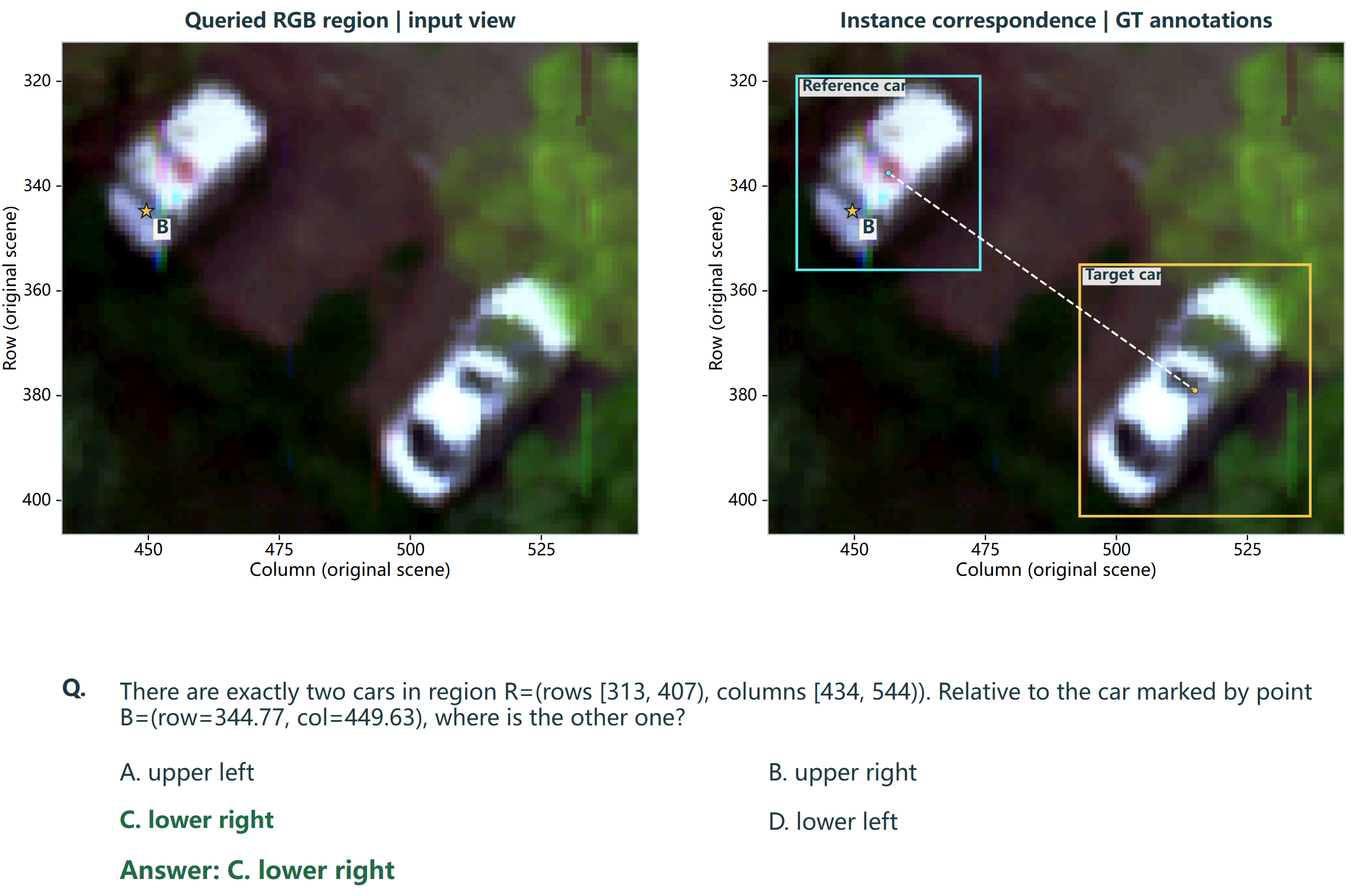}
\clearpage
\HyperGPTQualitativePair{Change presence}{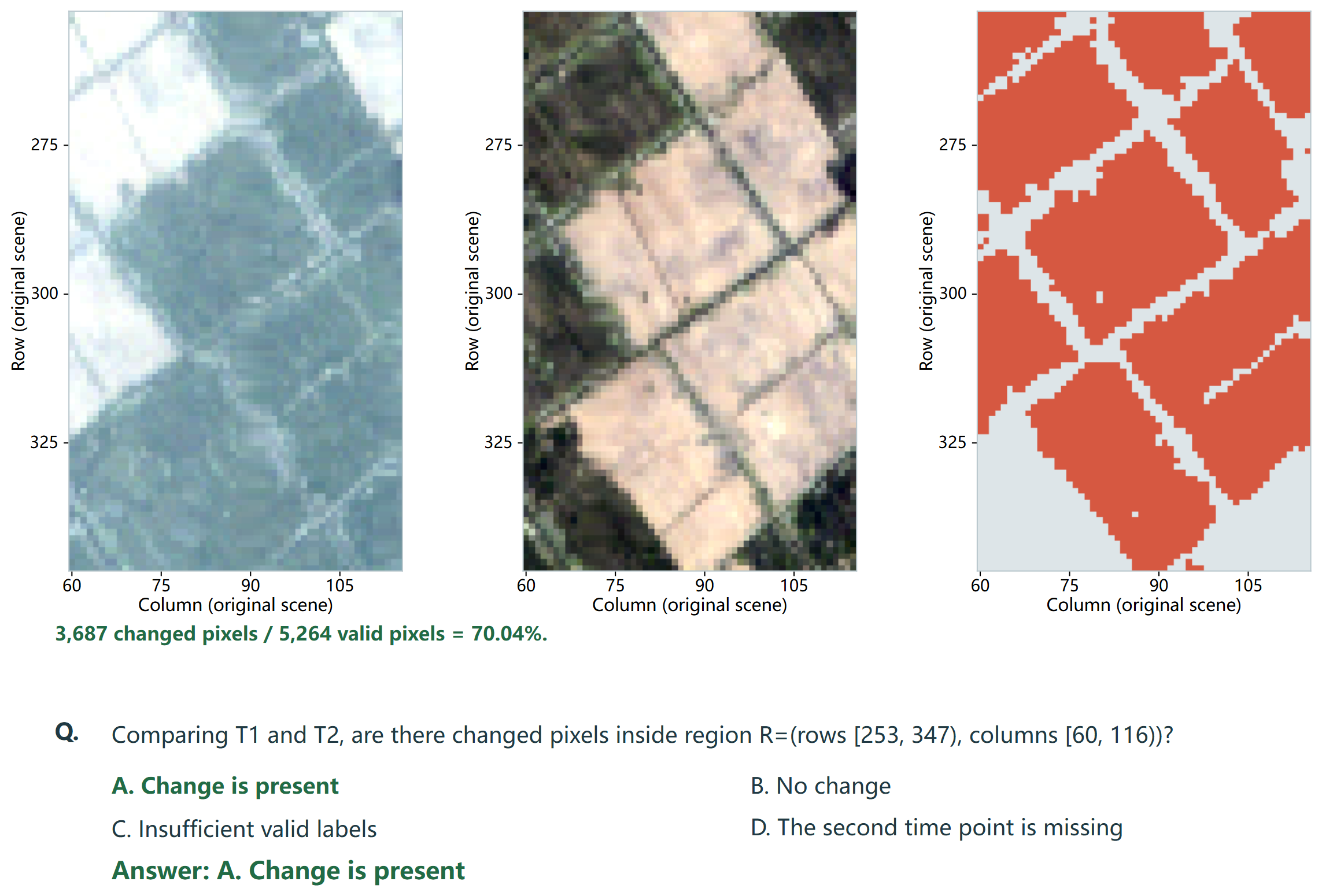}{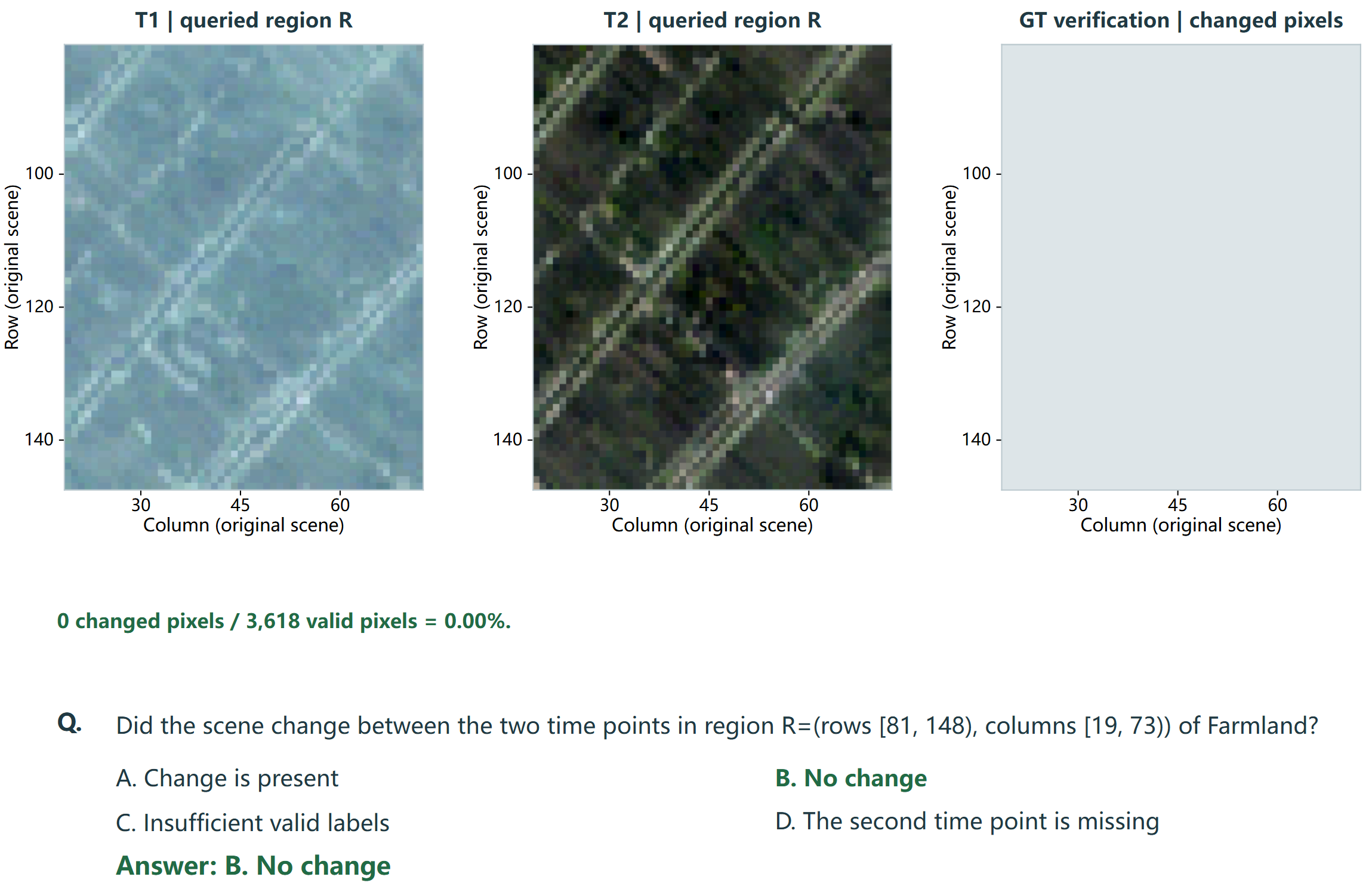}
\clearpage
\HyperGPTQualitativePair{Change localization}{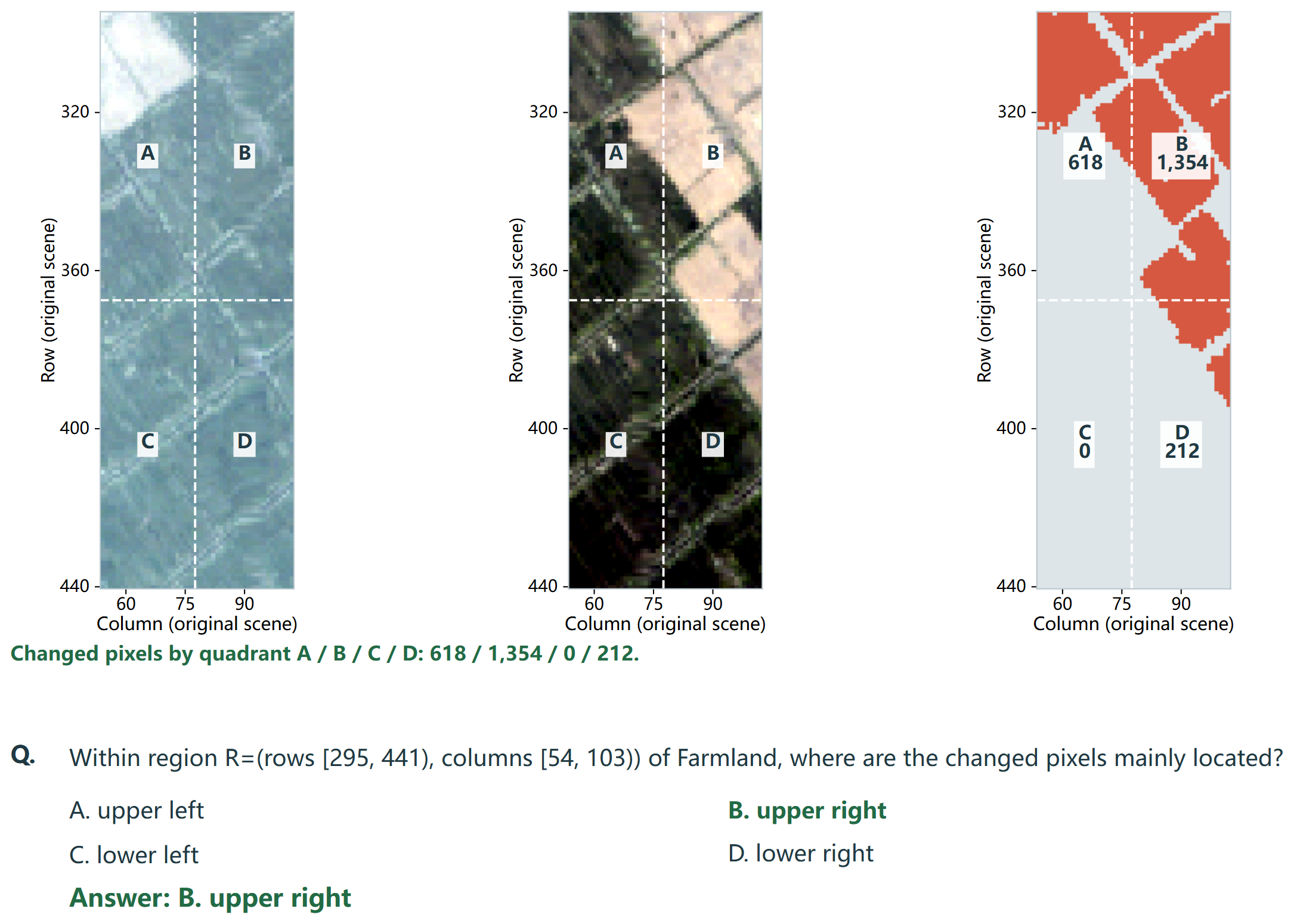}{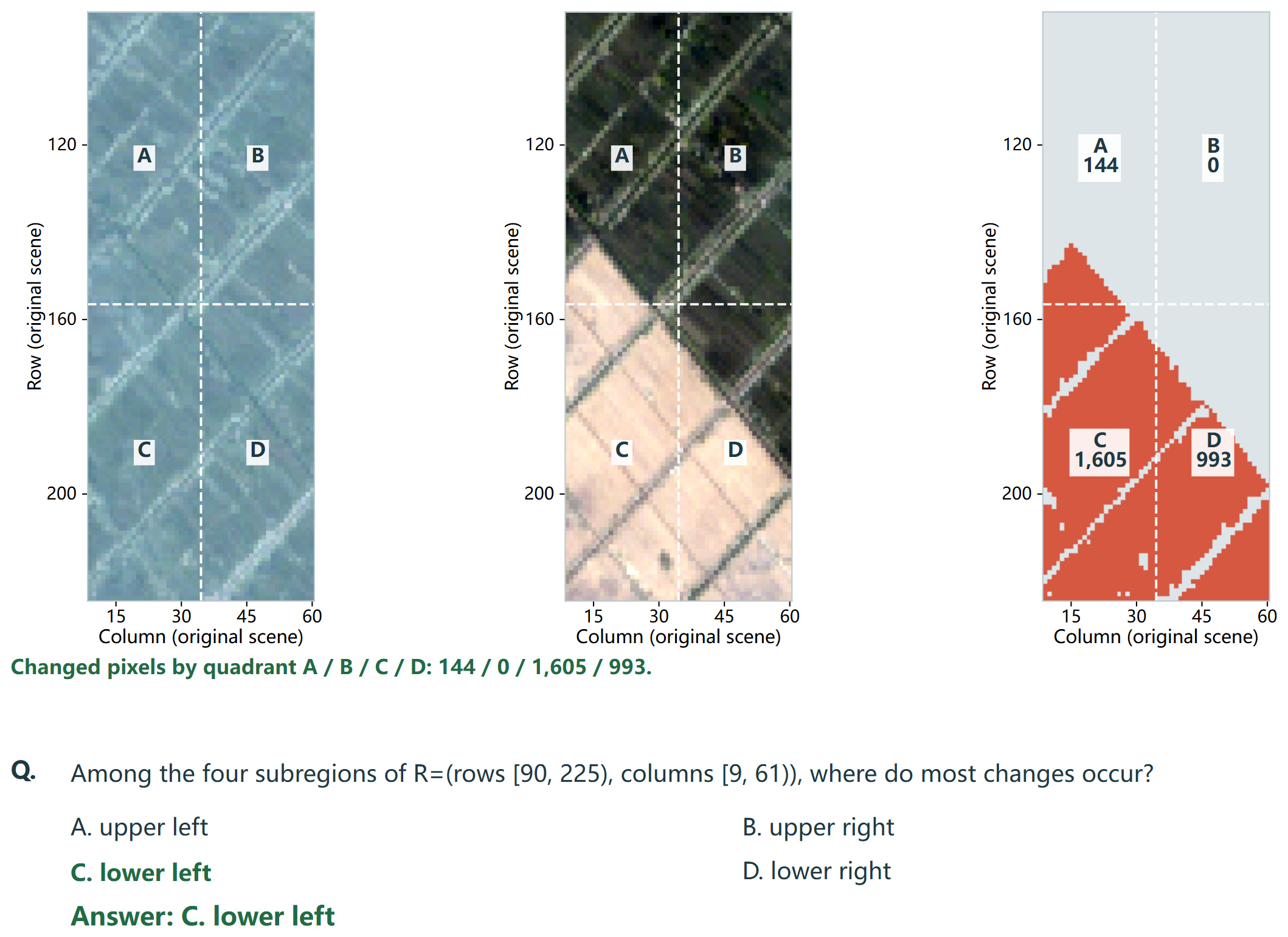}
\clearpage
\HyperGPTQualitativePair{Changed-area fraction}{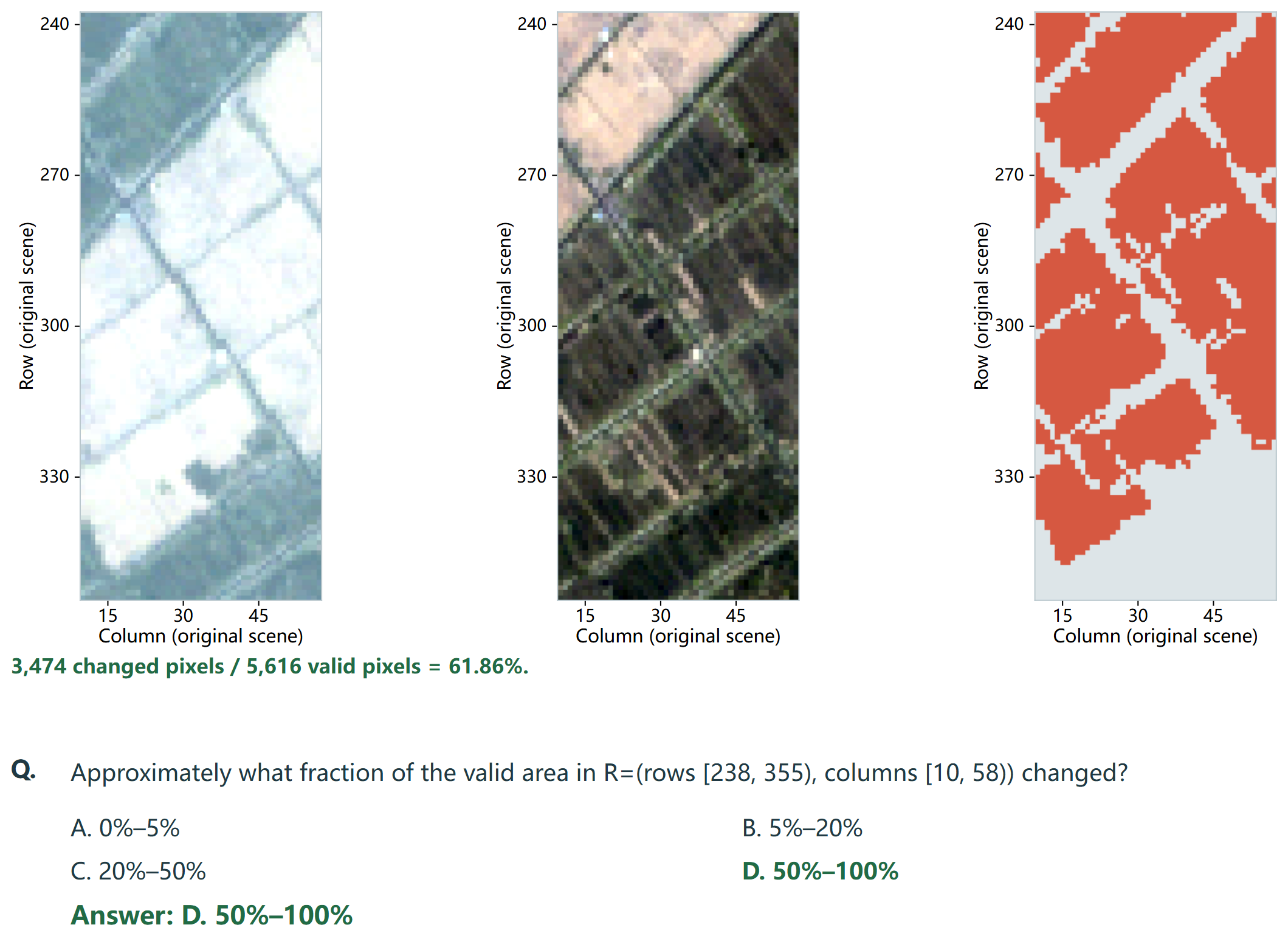}{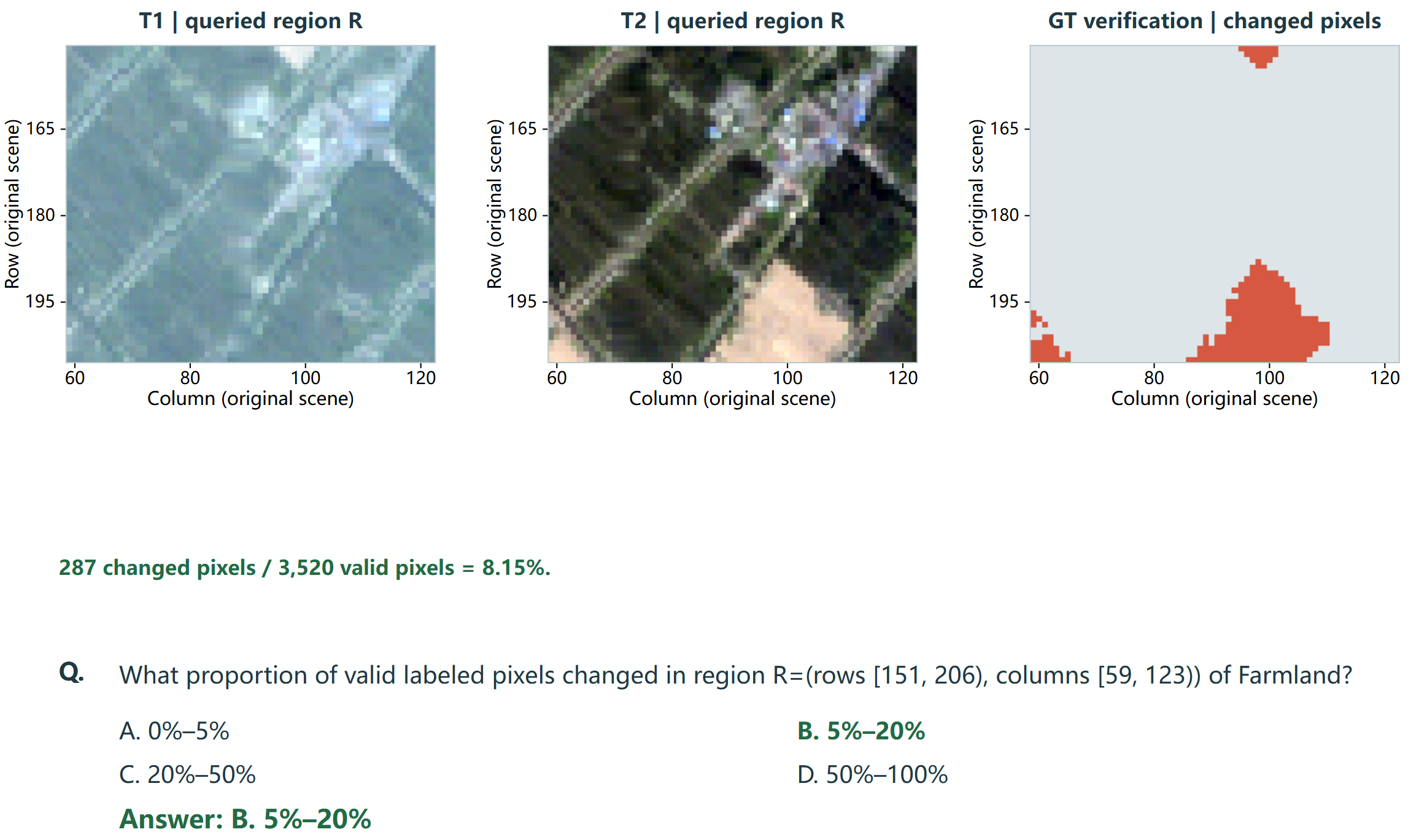}
\clearpage
\HyperGPTQualitativePair{Spectral change localization}{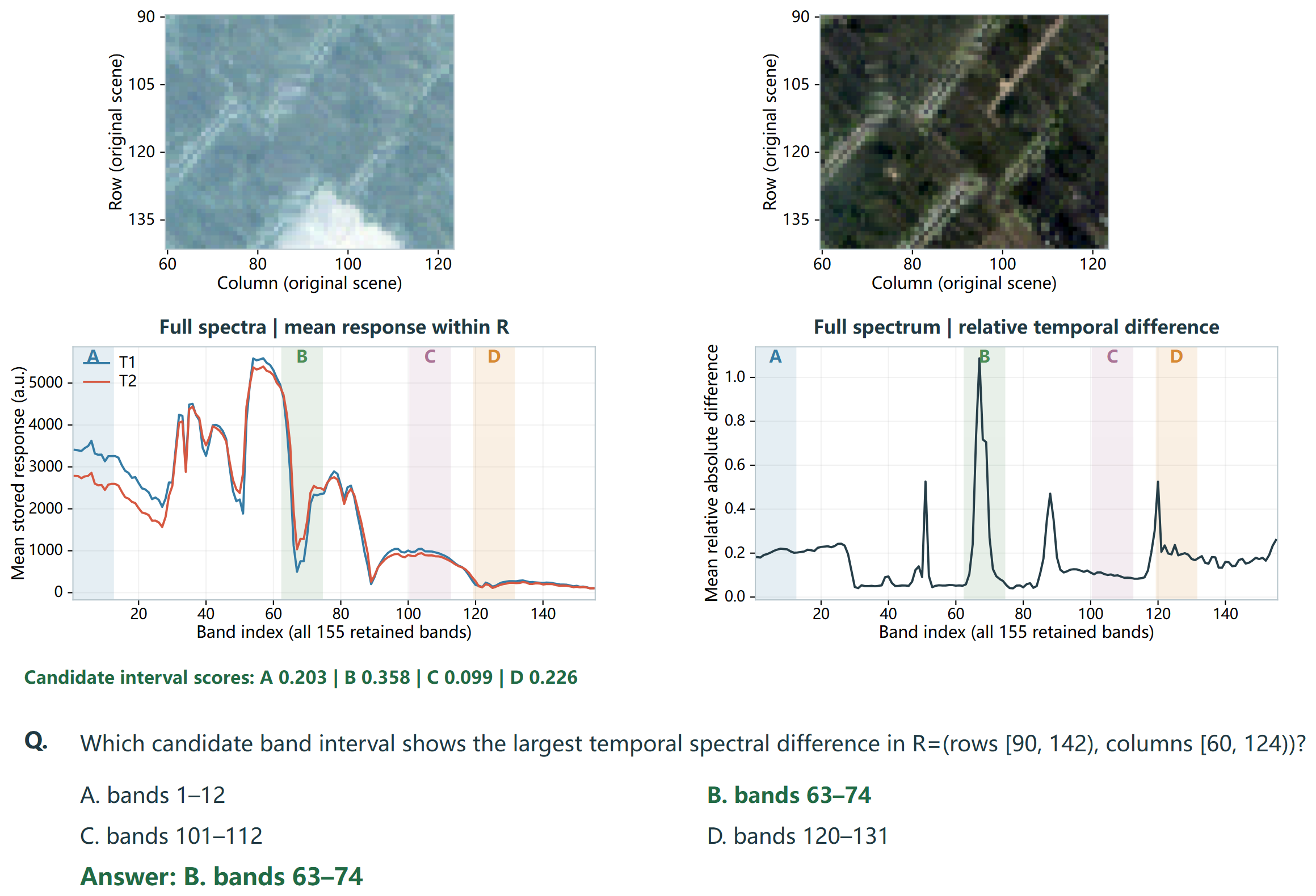}{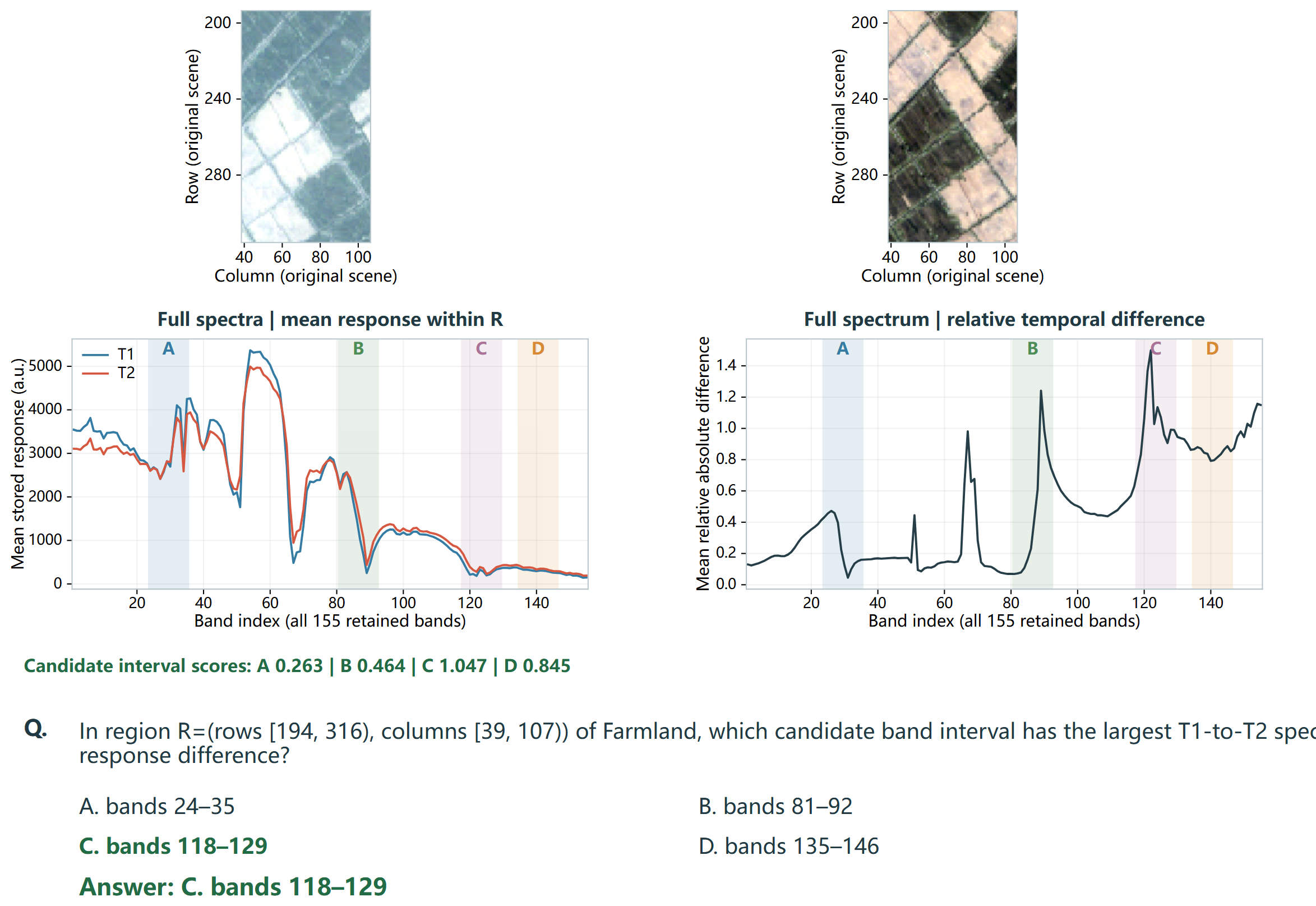}
\clearpage
\HyperGPTQualitativePair{Change-region counting}{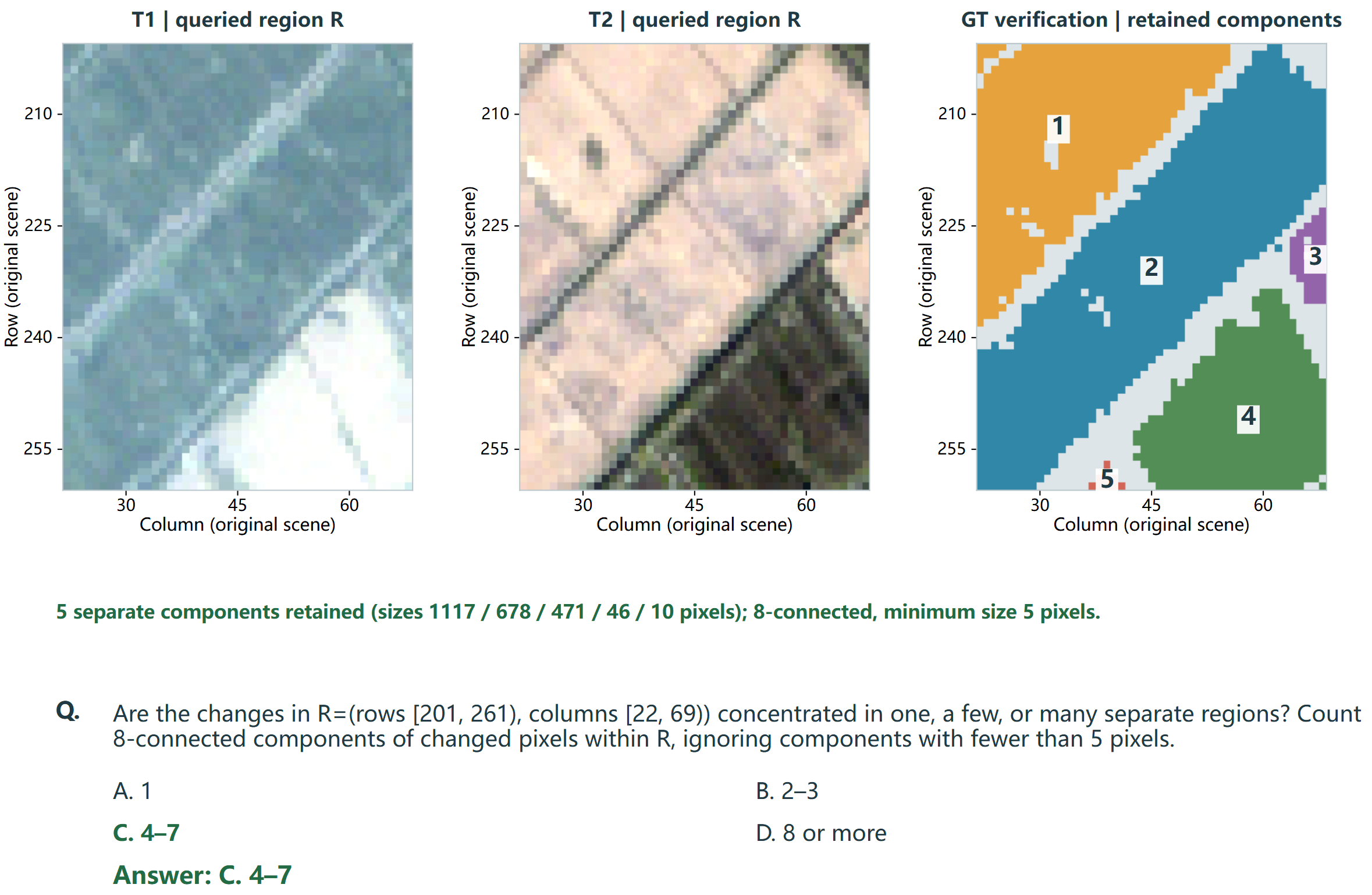}{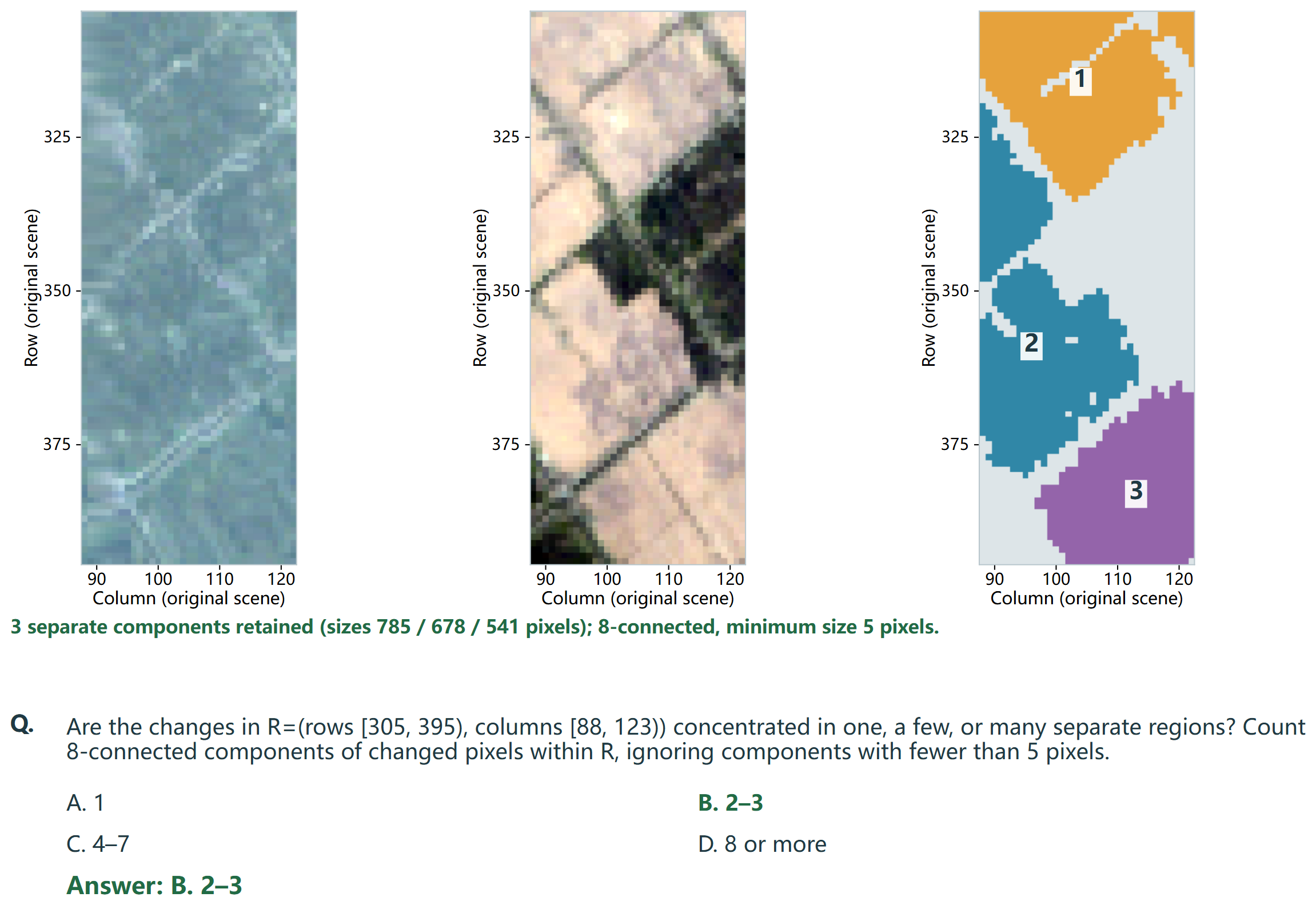}
\clearpage
\HyperGPTQualitativePair{Anomaly presence}{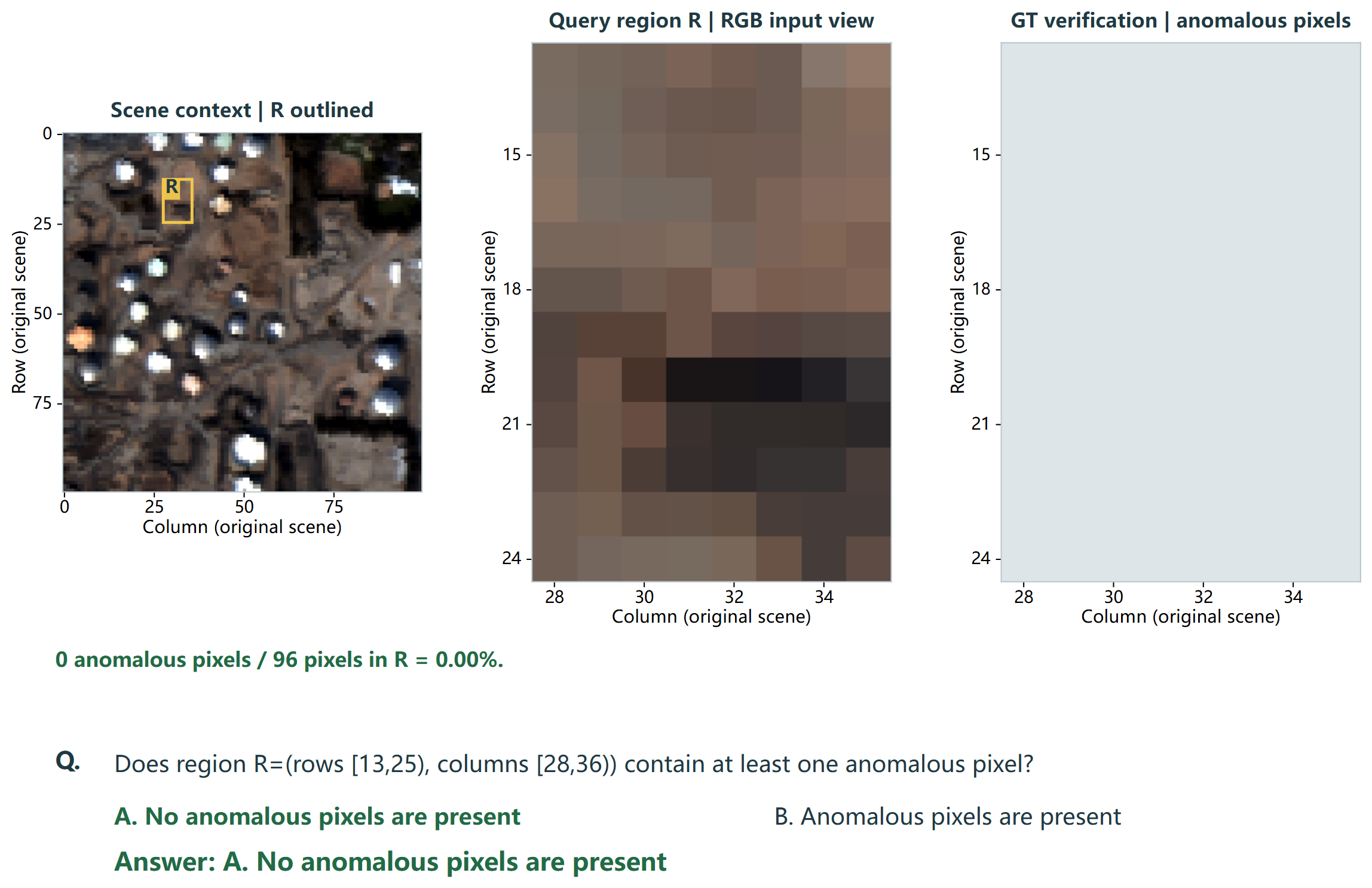}{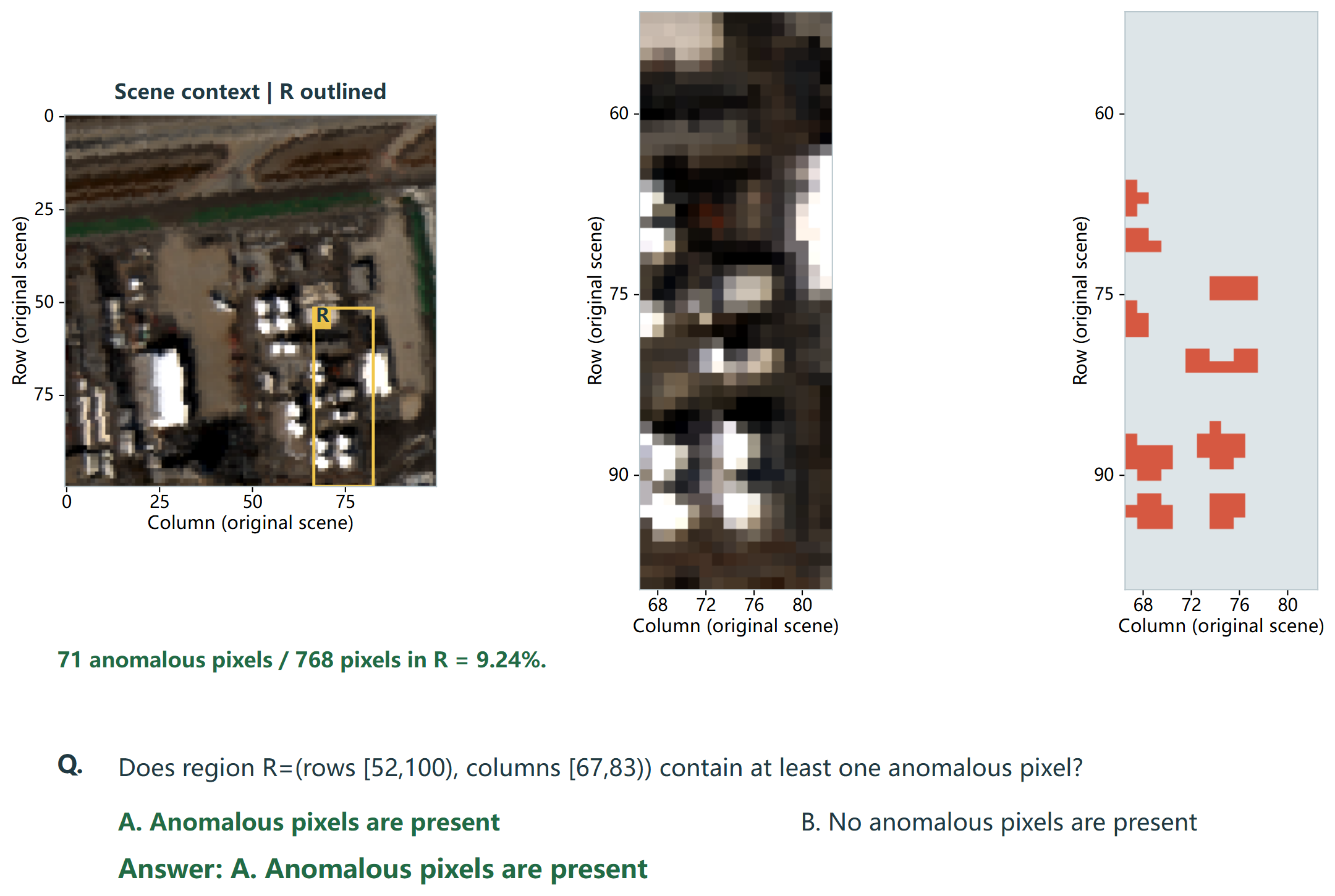}
\clearpage
\HyperGPTQualitativePair{Anomaly localization}{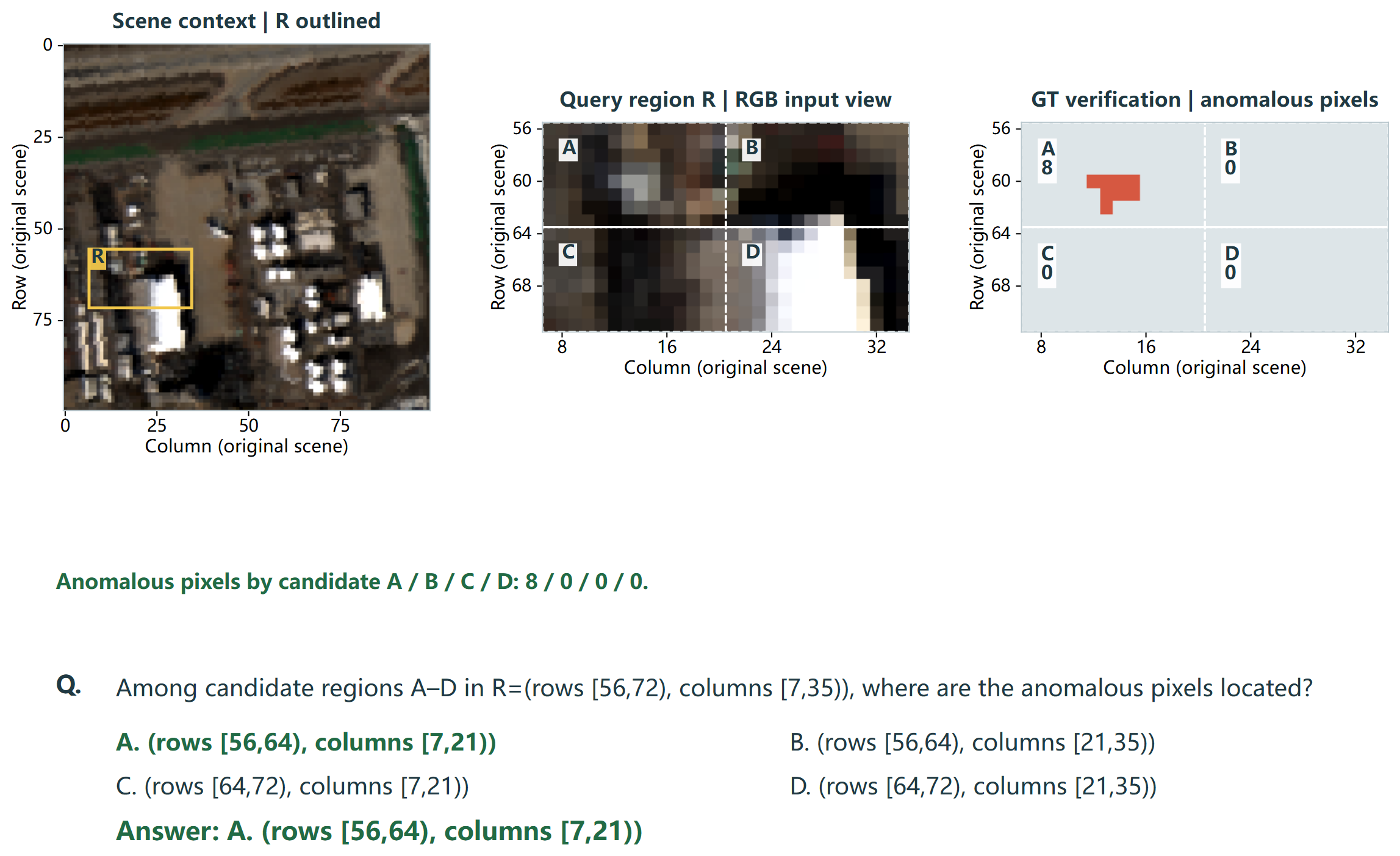}{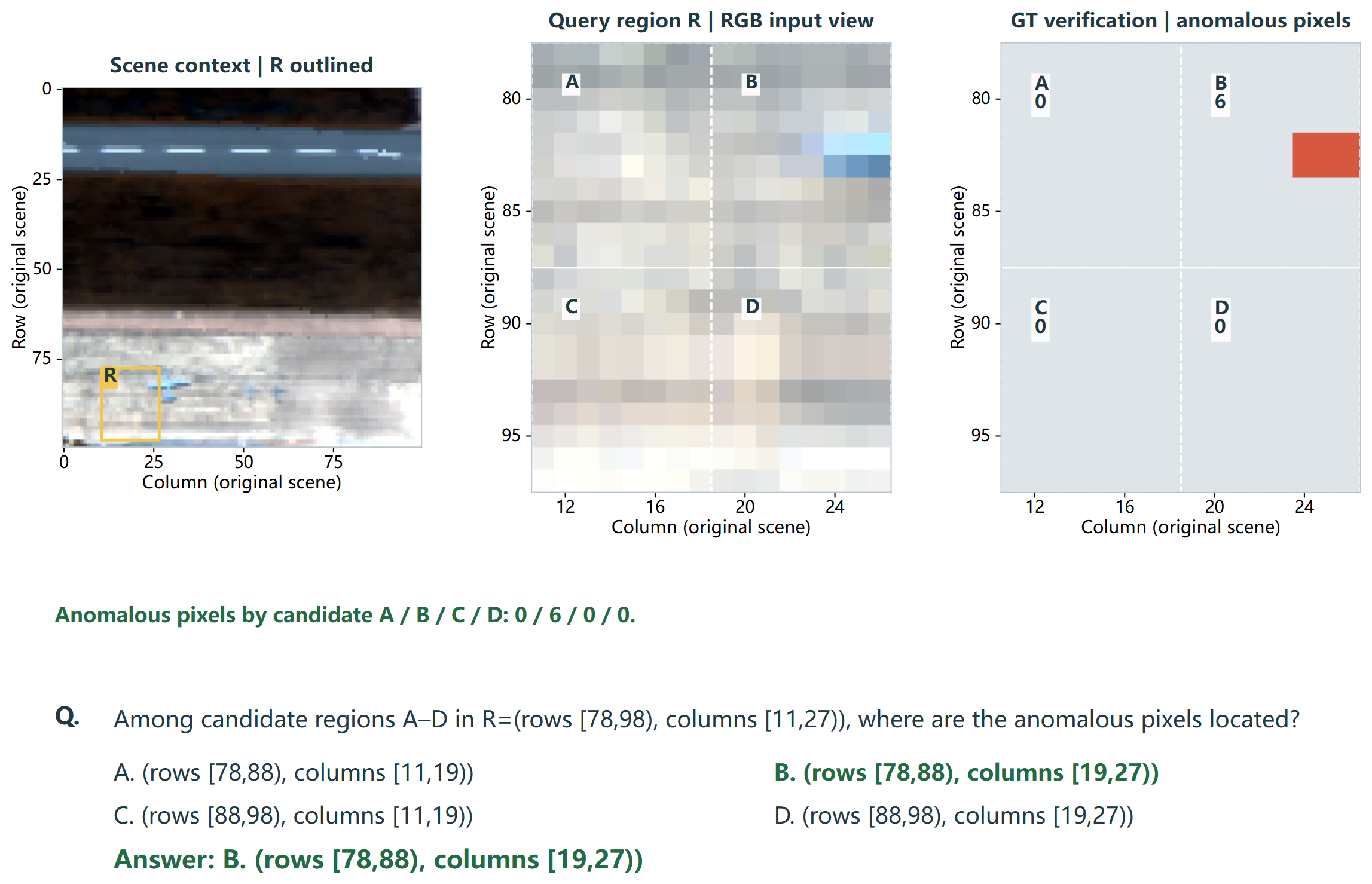}
\clearpage
\HyperGPTQualitativePair{Anomalous-area fraction}{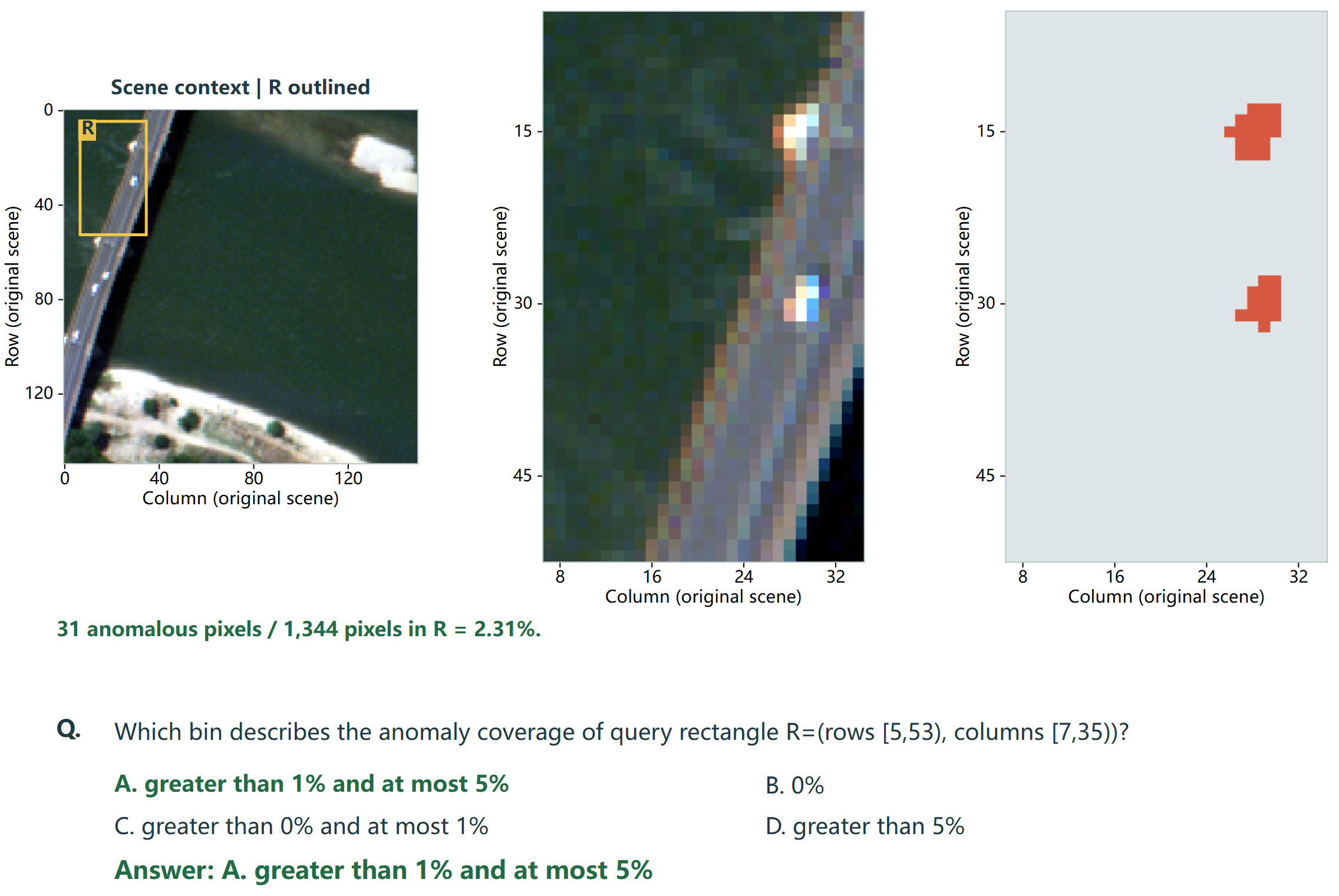}{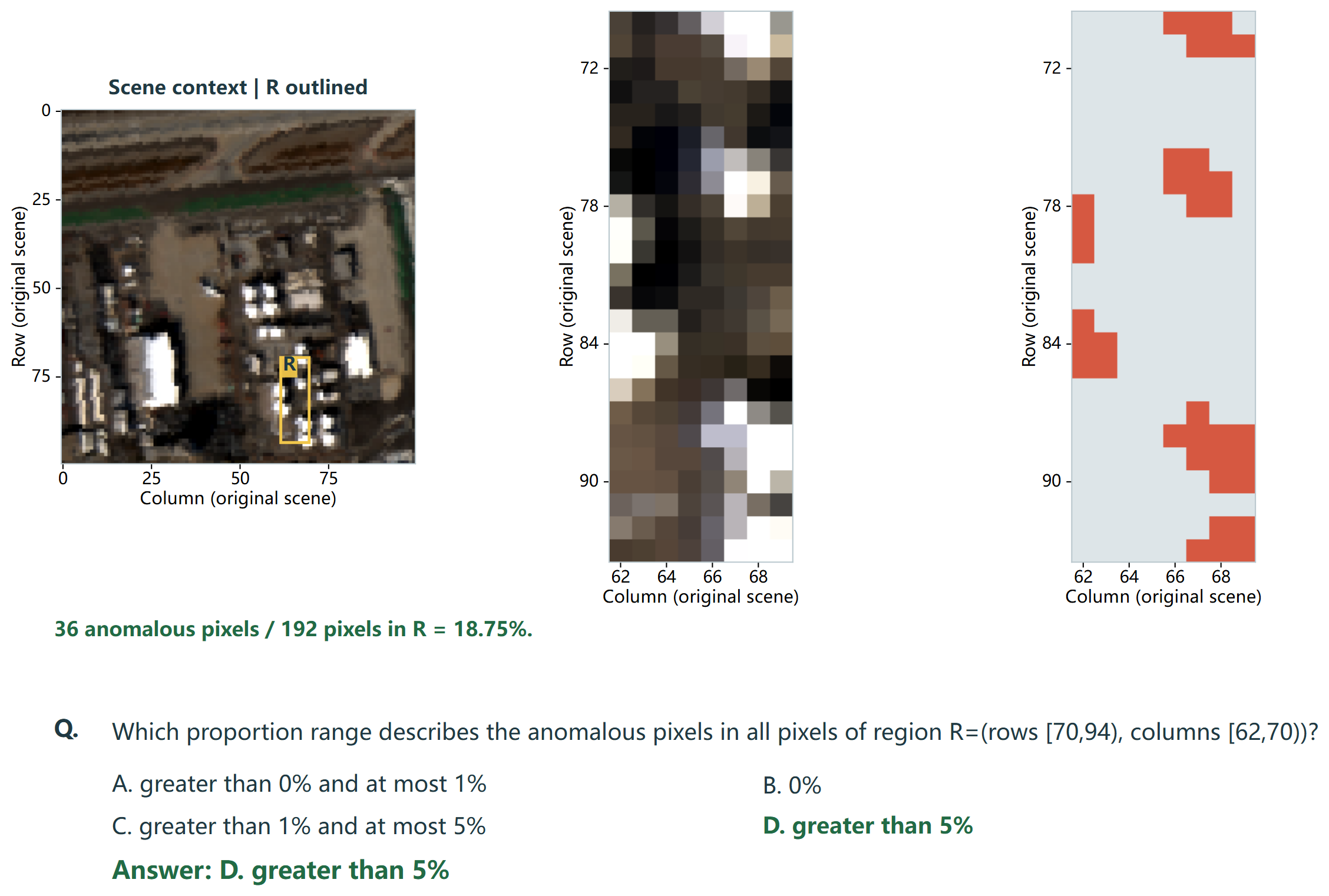}
\clearpage
\HyperGPTQualitativePair{Spectral anomaly contrast}{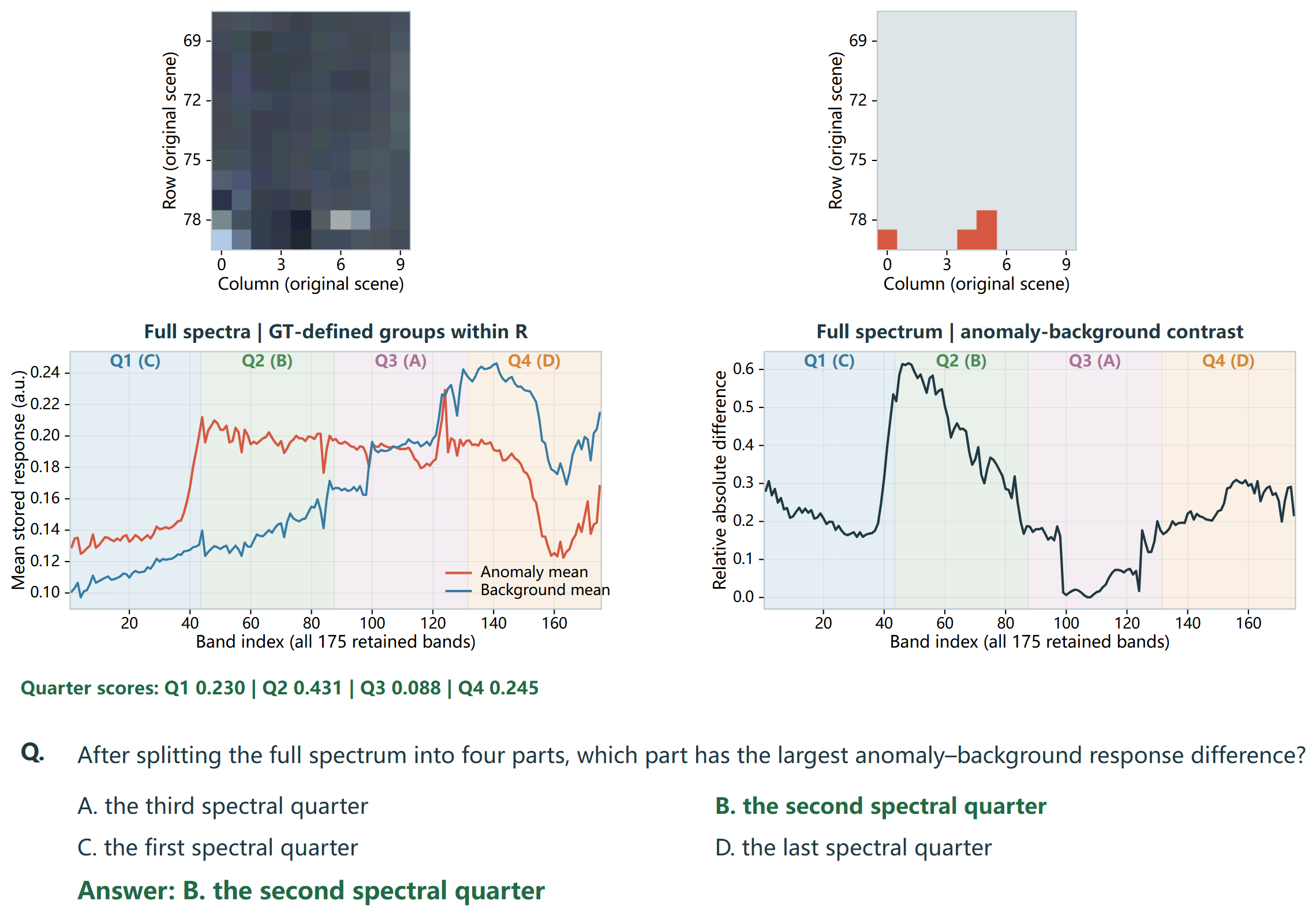}{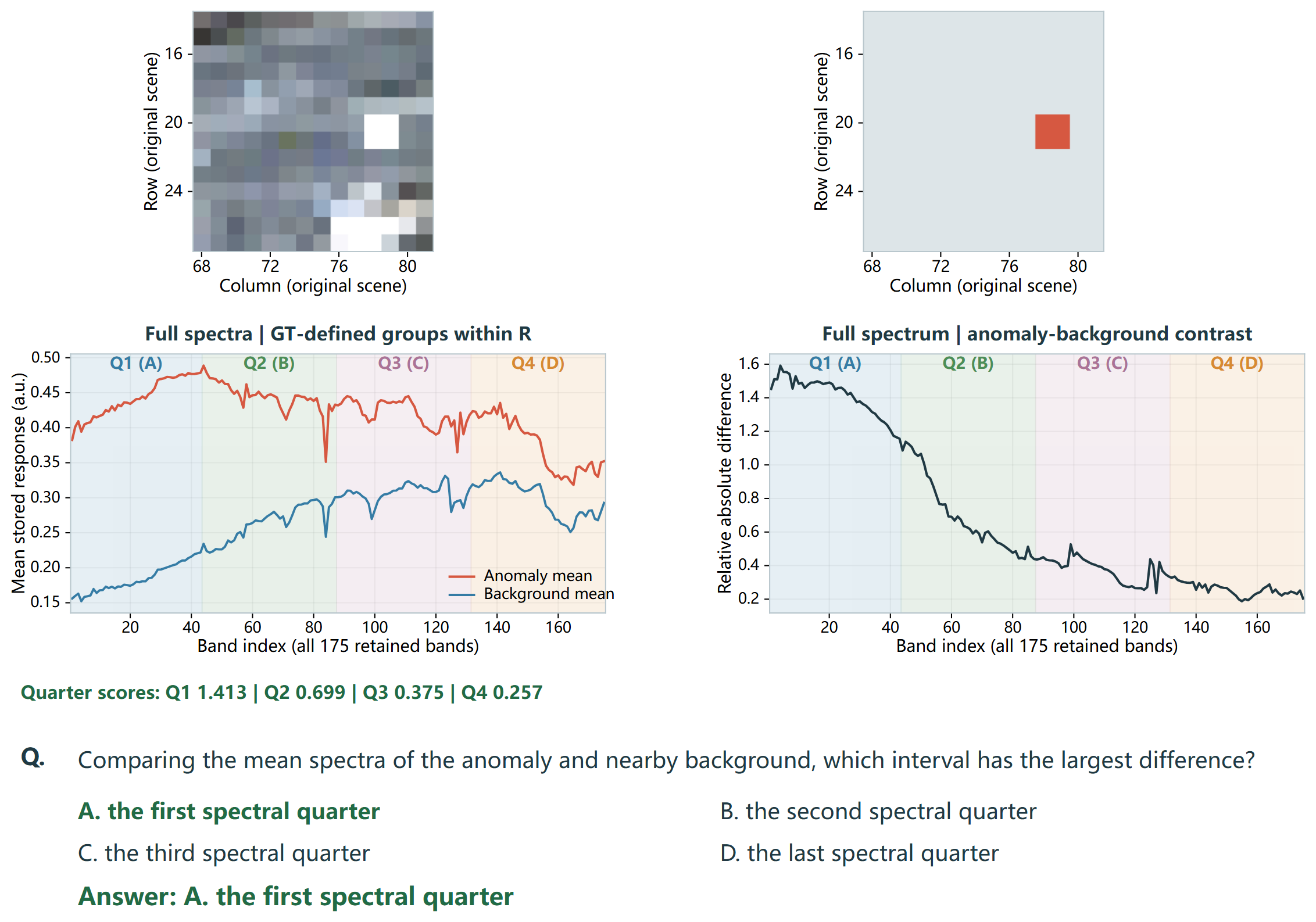}
\clearpage
\HyperGPTQualitativePair{Target presence}{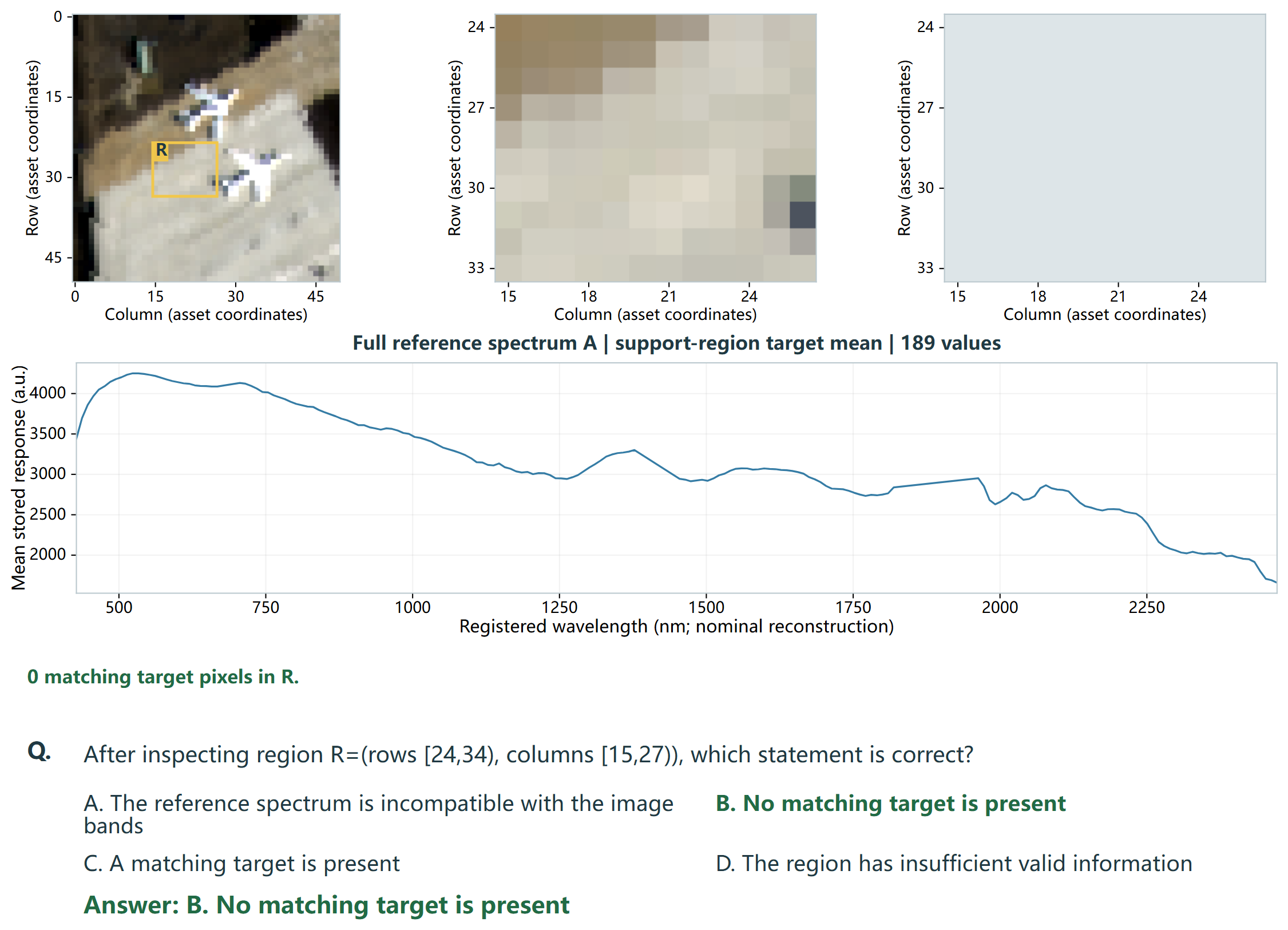}{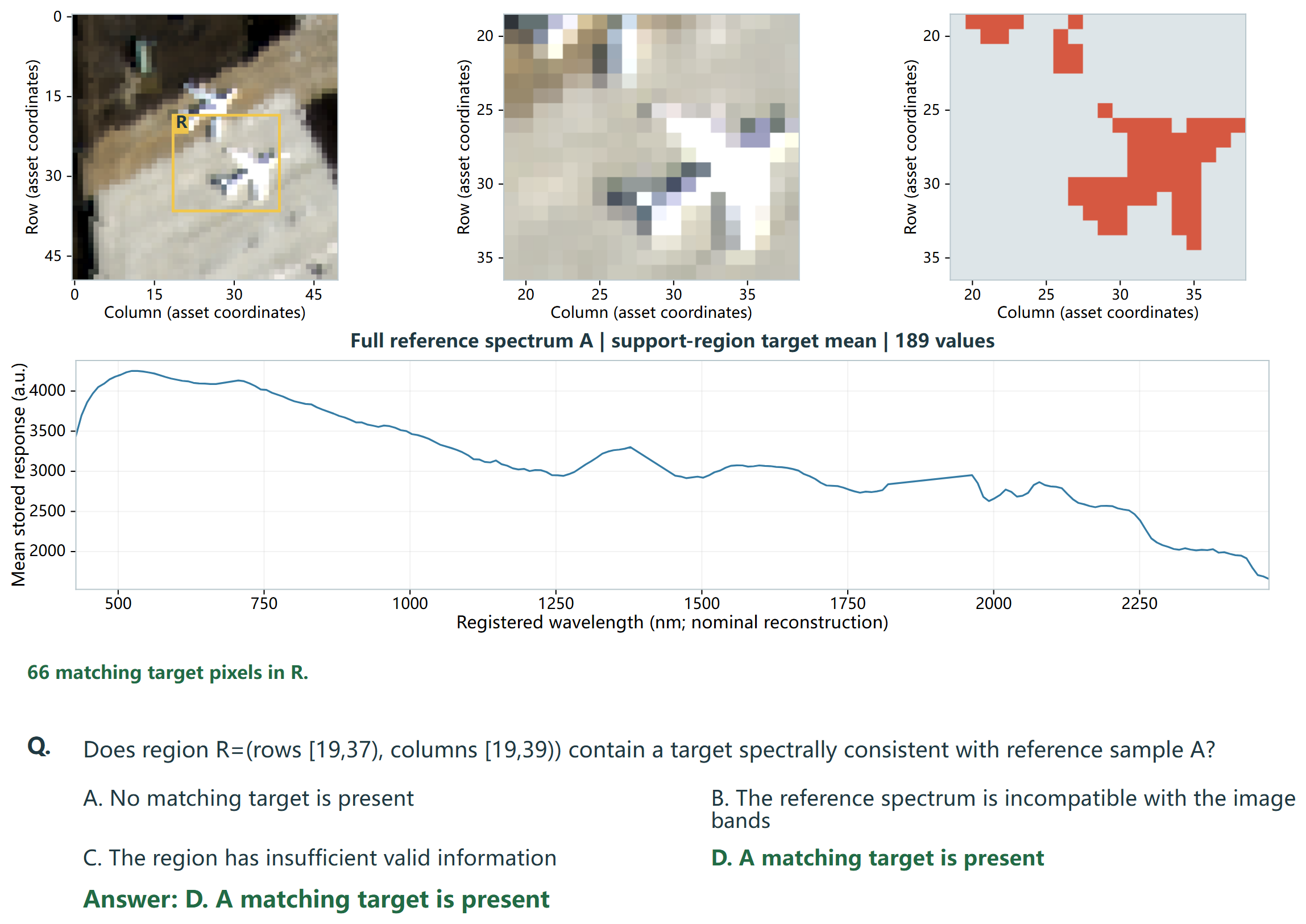}
\clearpage
\HyperGPTQualitativePair{Target localization}{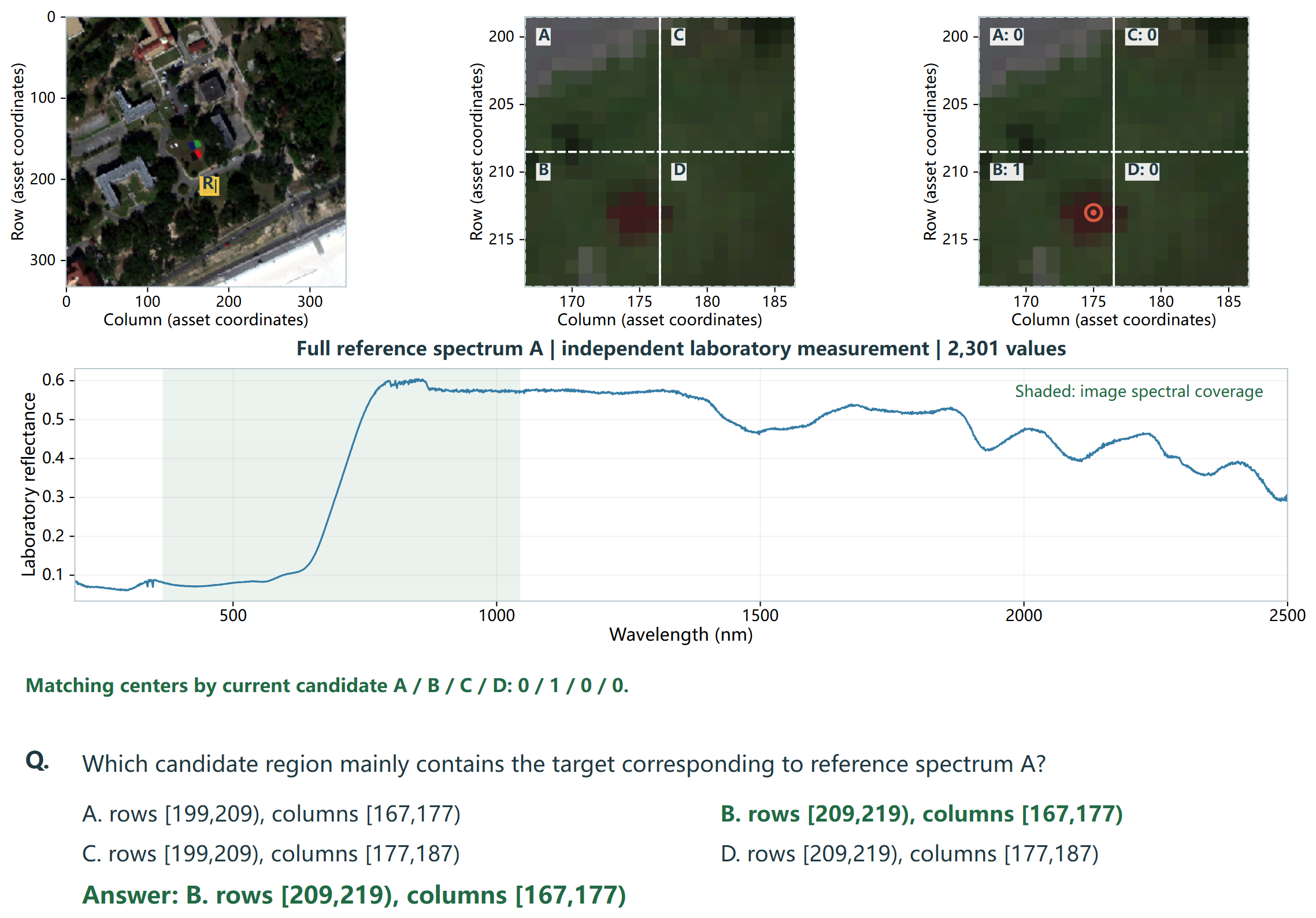}{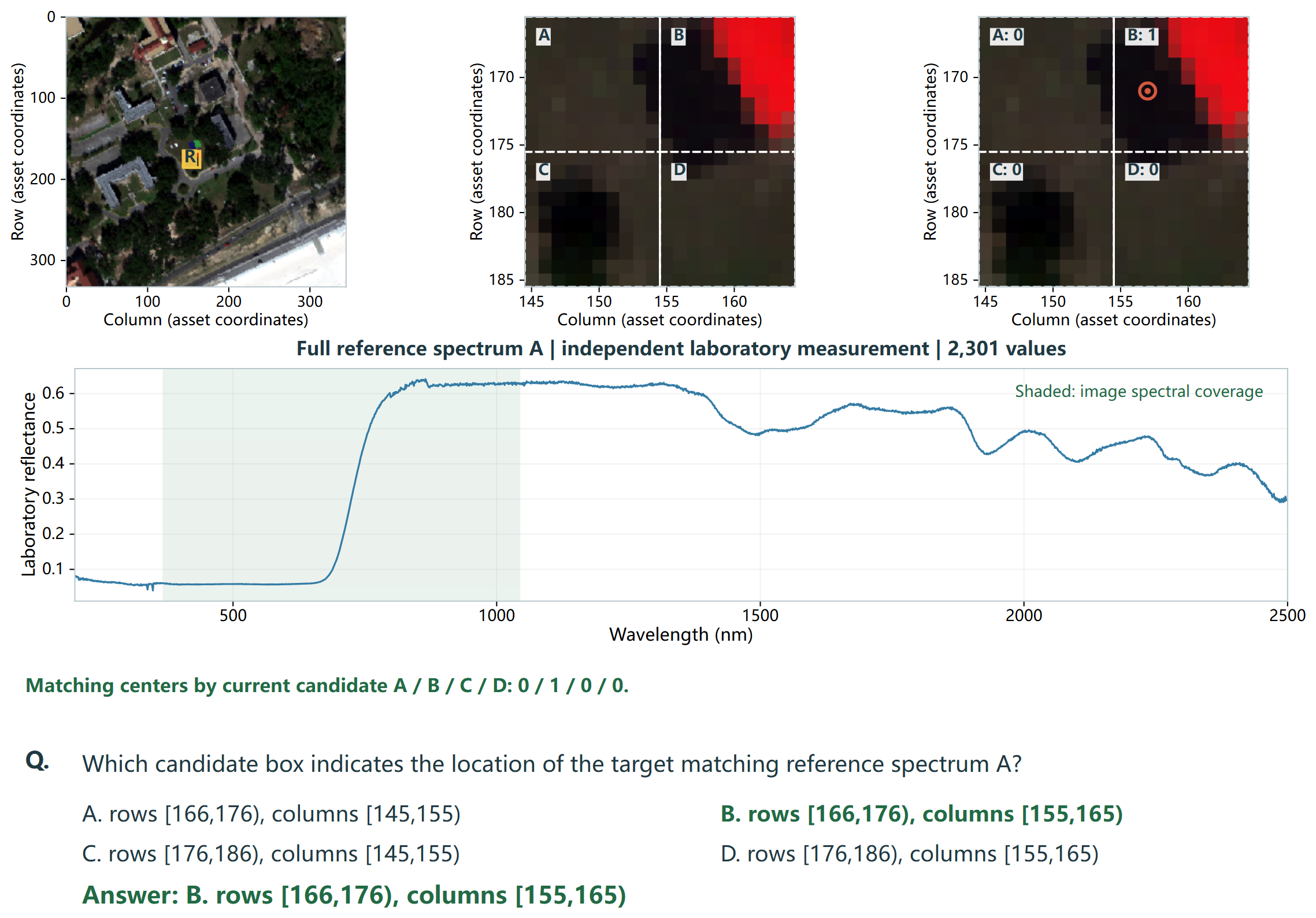}
\clearpage
\HyperGPTQualitativePair{Abundance estimation}{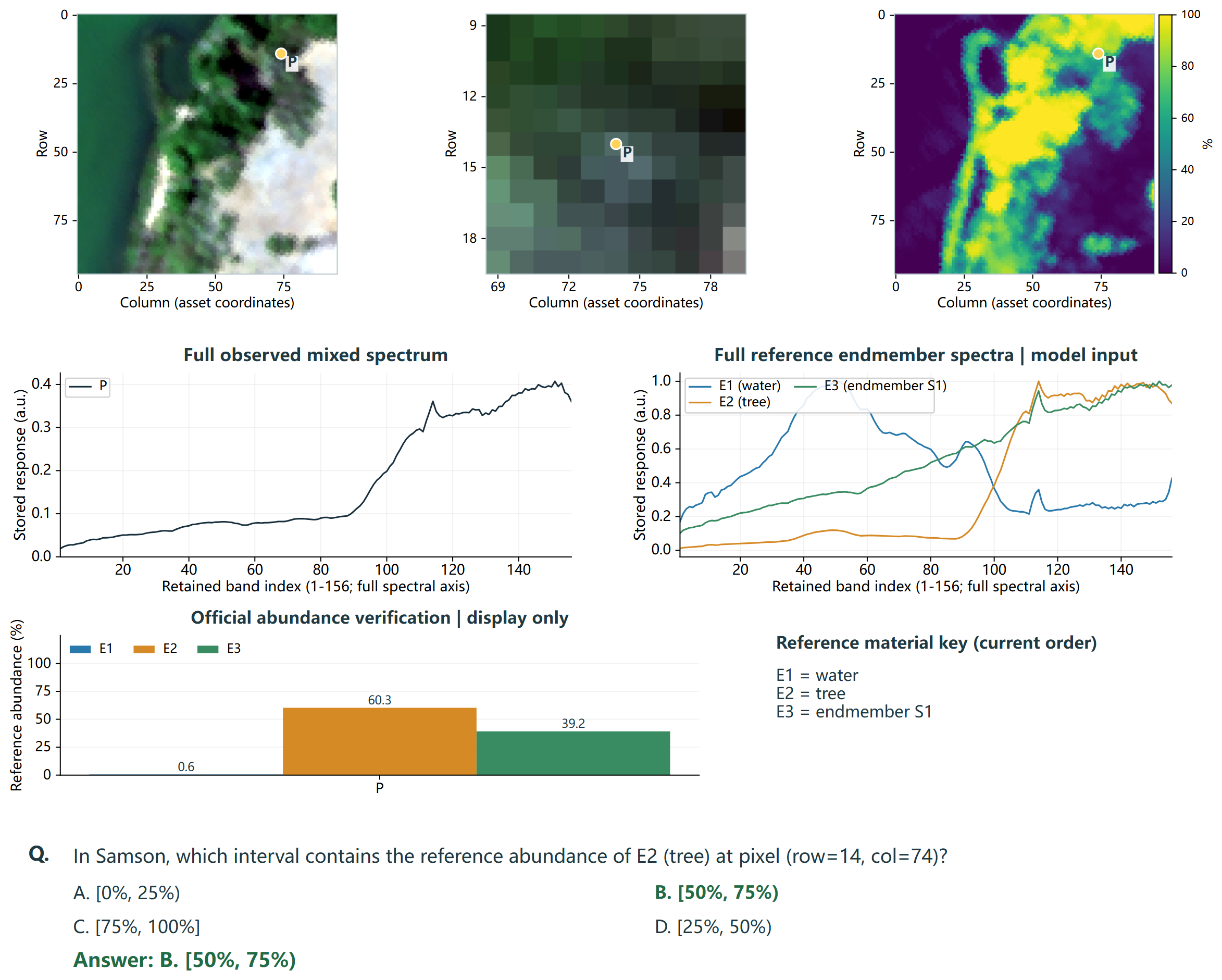}{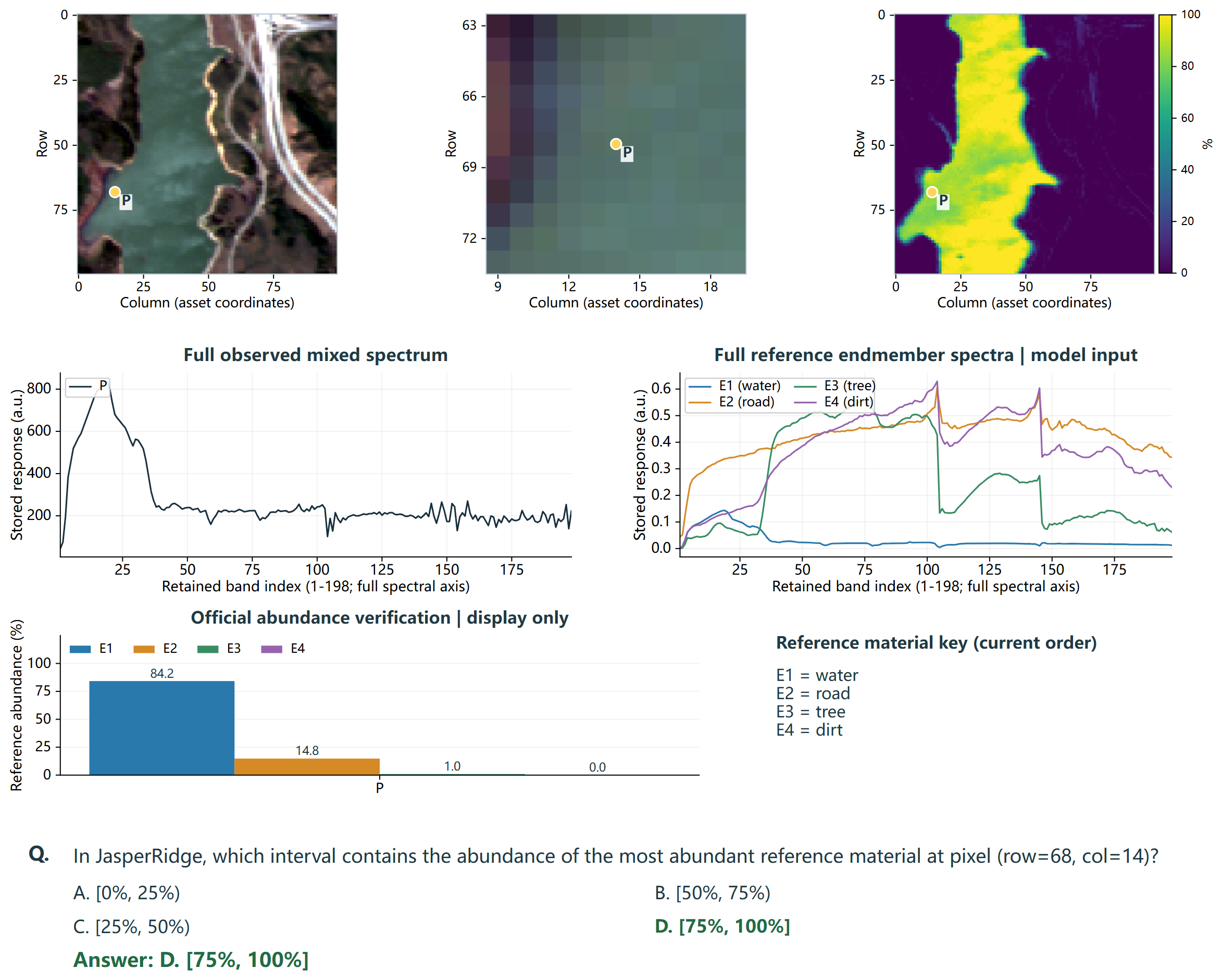}
\clearpage
\HyperGPTQualitativePair{Abundance difference}{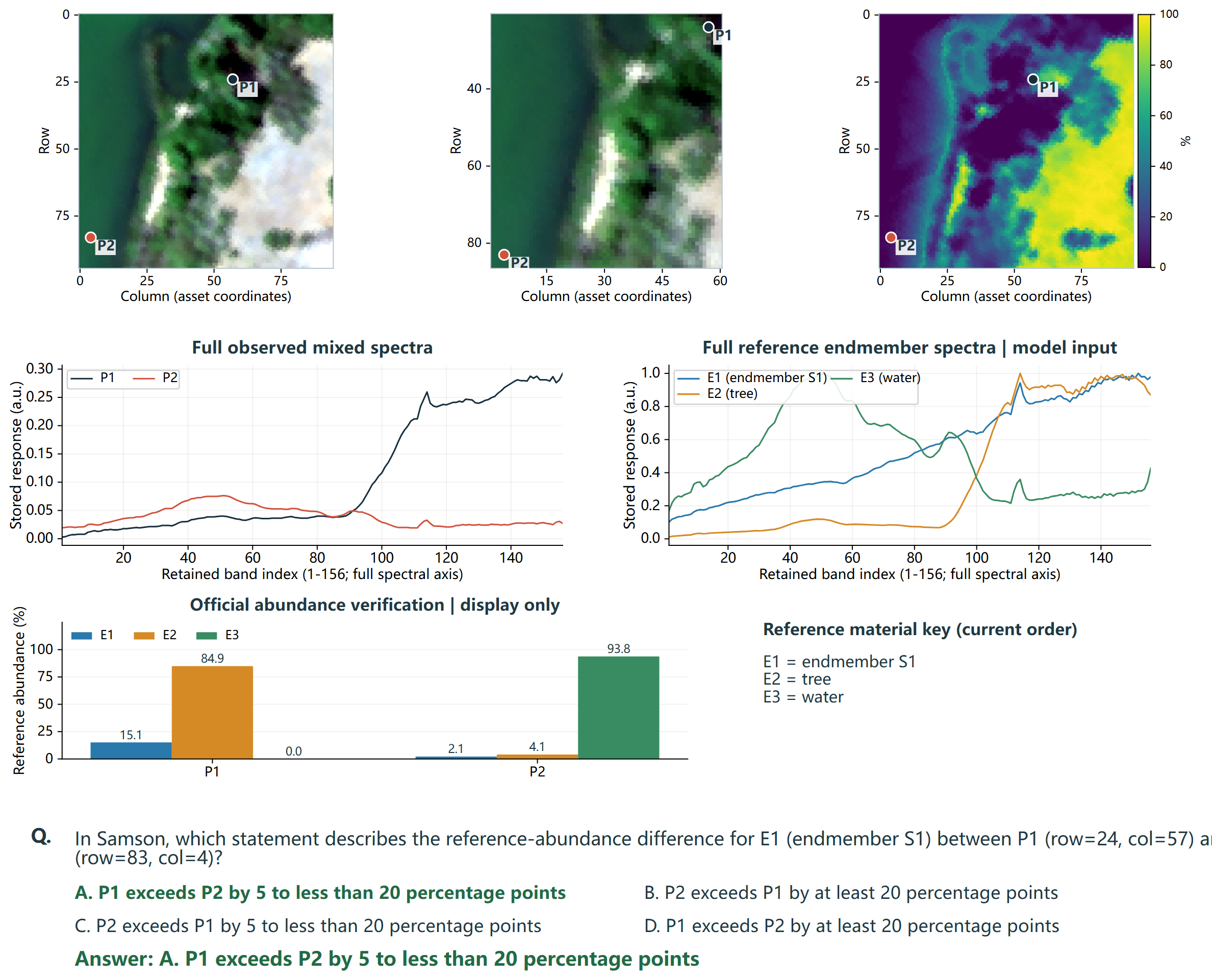}{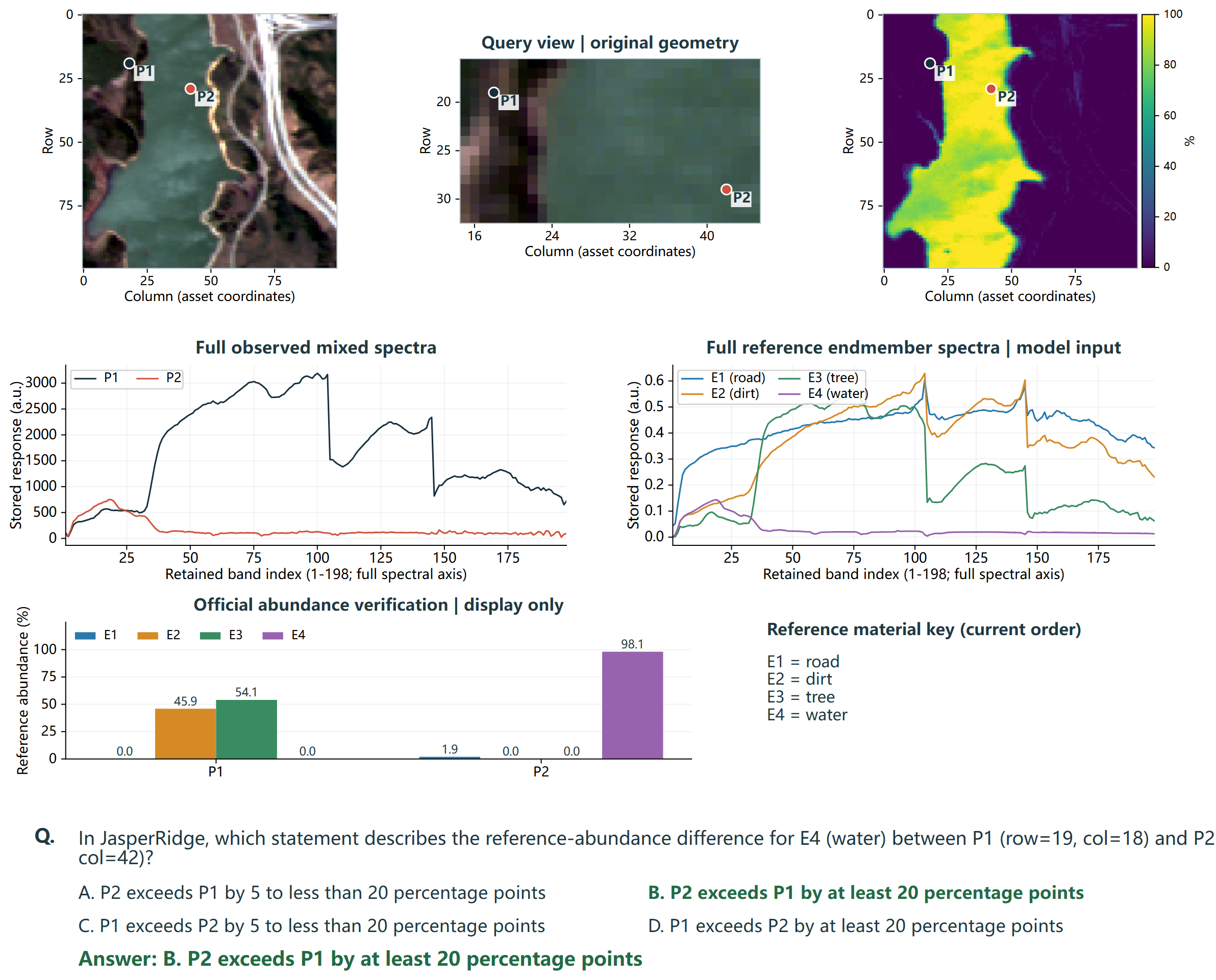}
\clearpage
\HyperGPTQualitativePair{Abundance ranking}{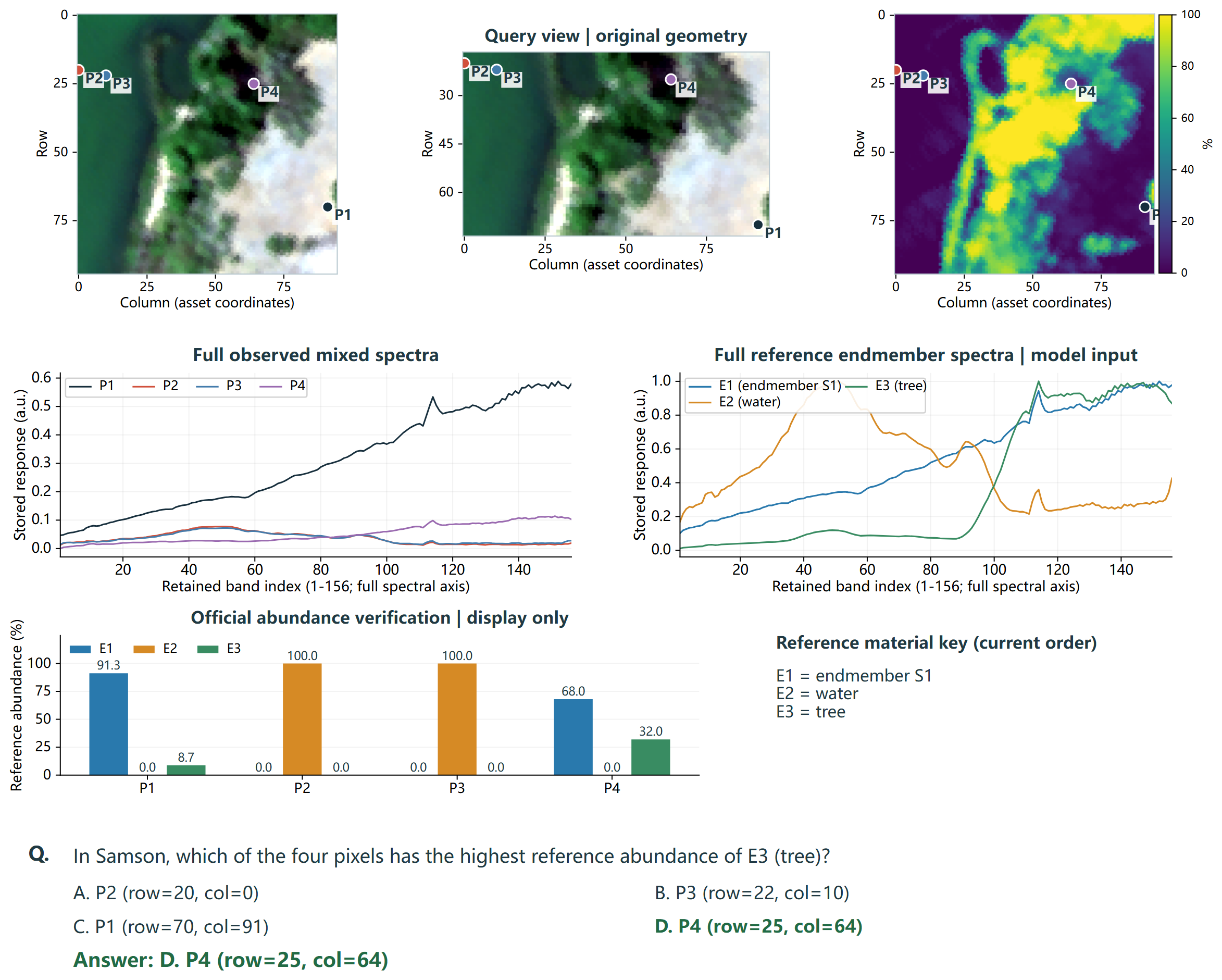}{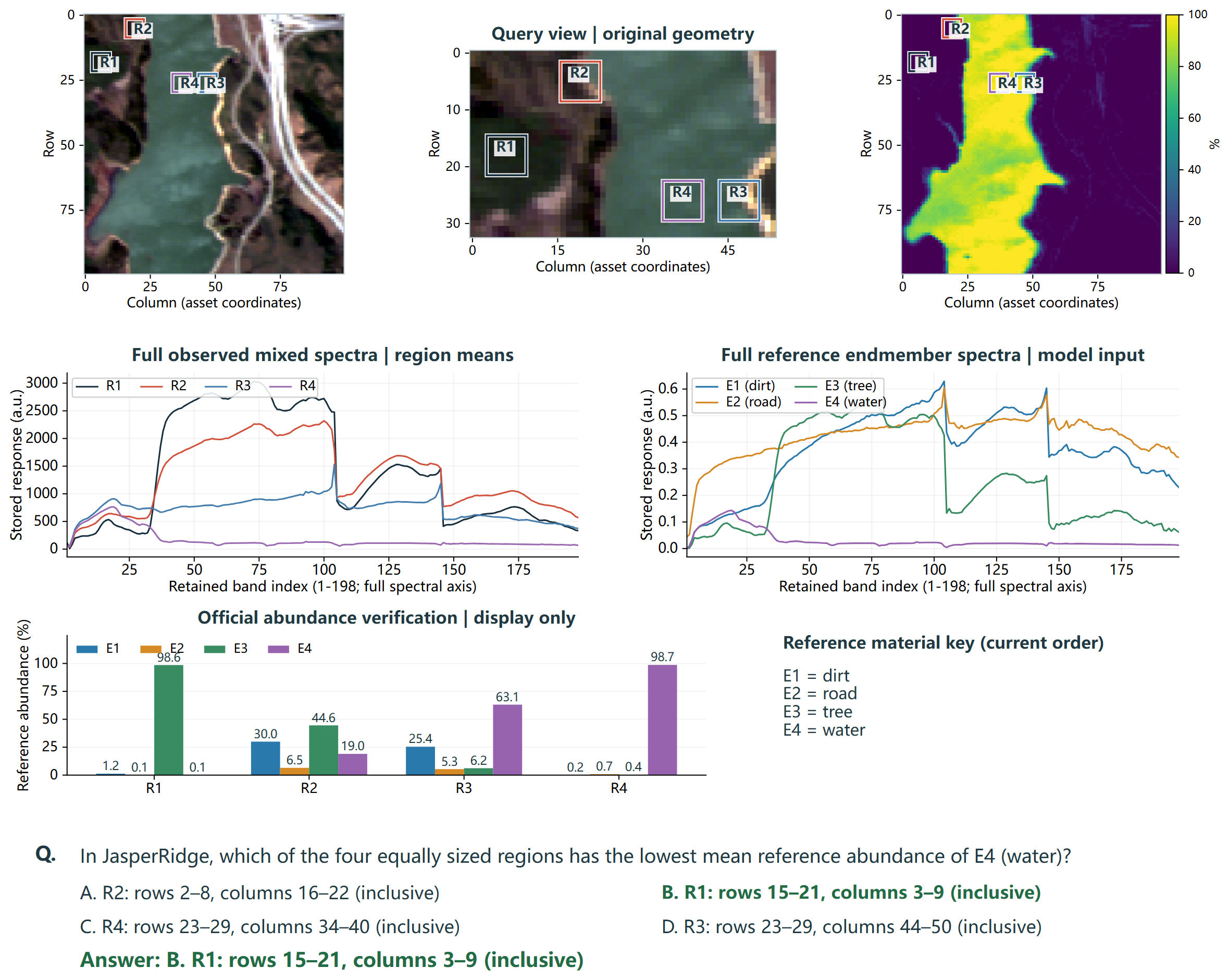}
\clearpage
\HyperGPTQualitativePair{Thresholded composition}{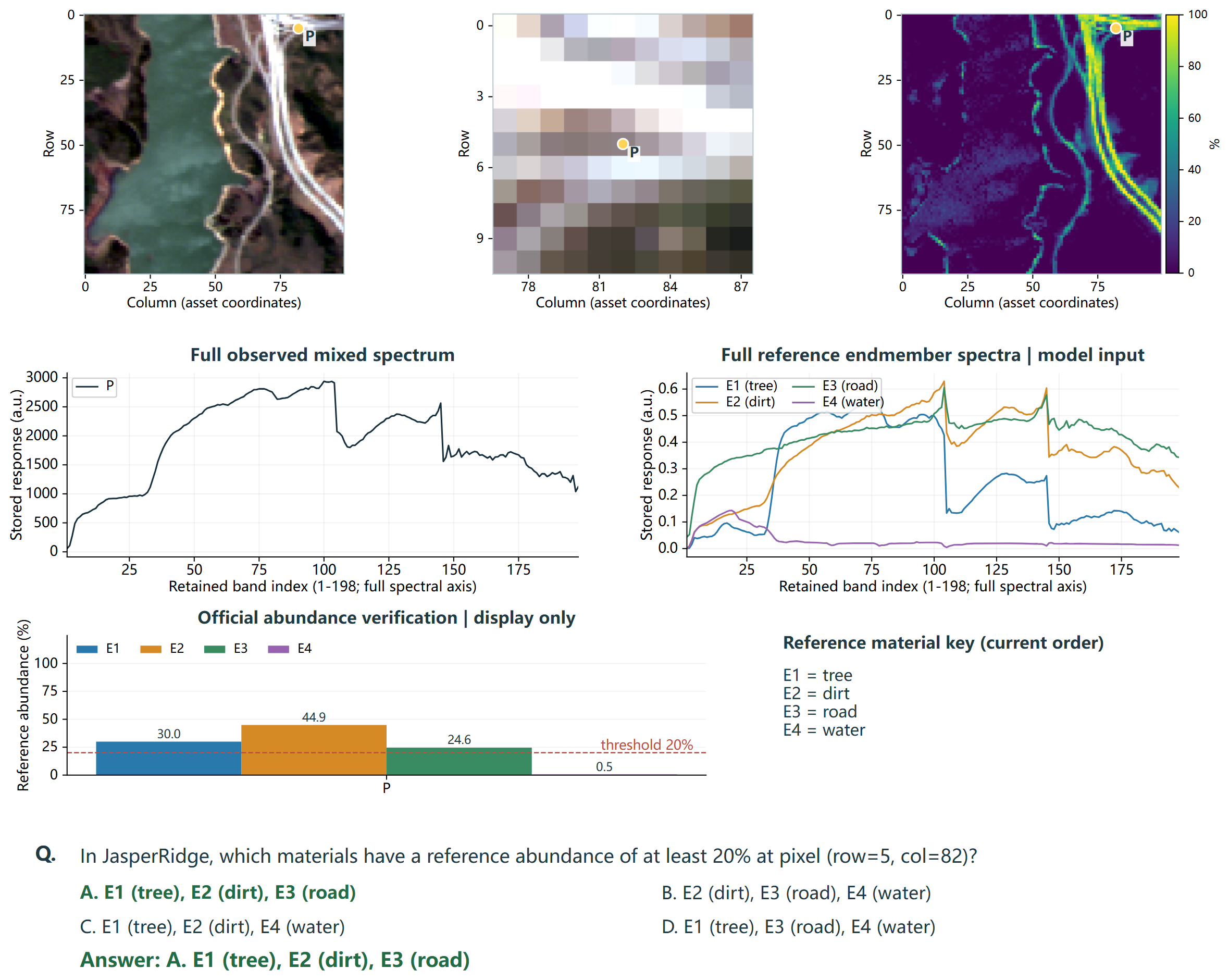}{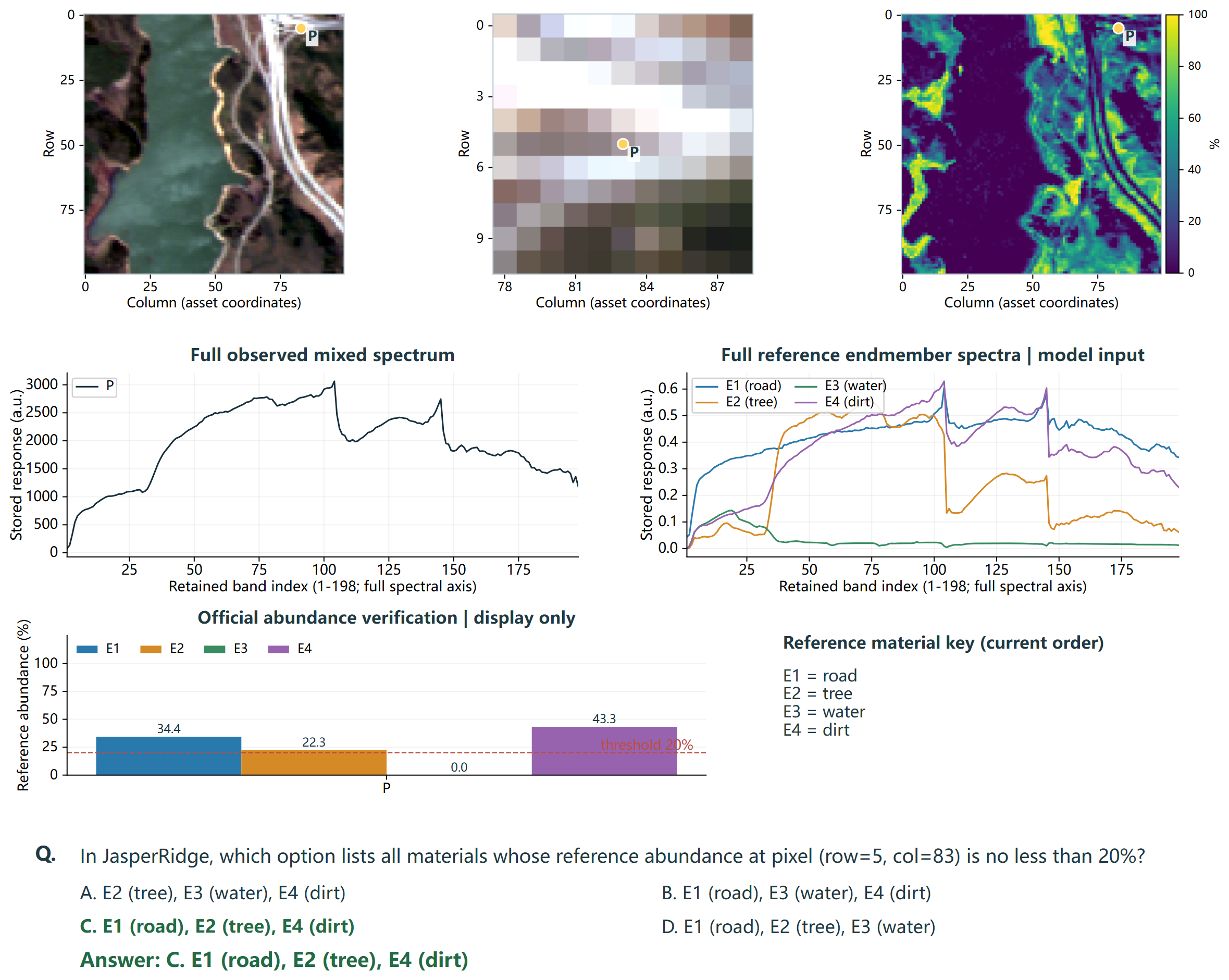}
\clearpage
\HyperGPTQualitativePair{Dominant material}{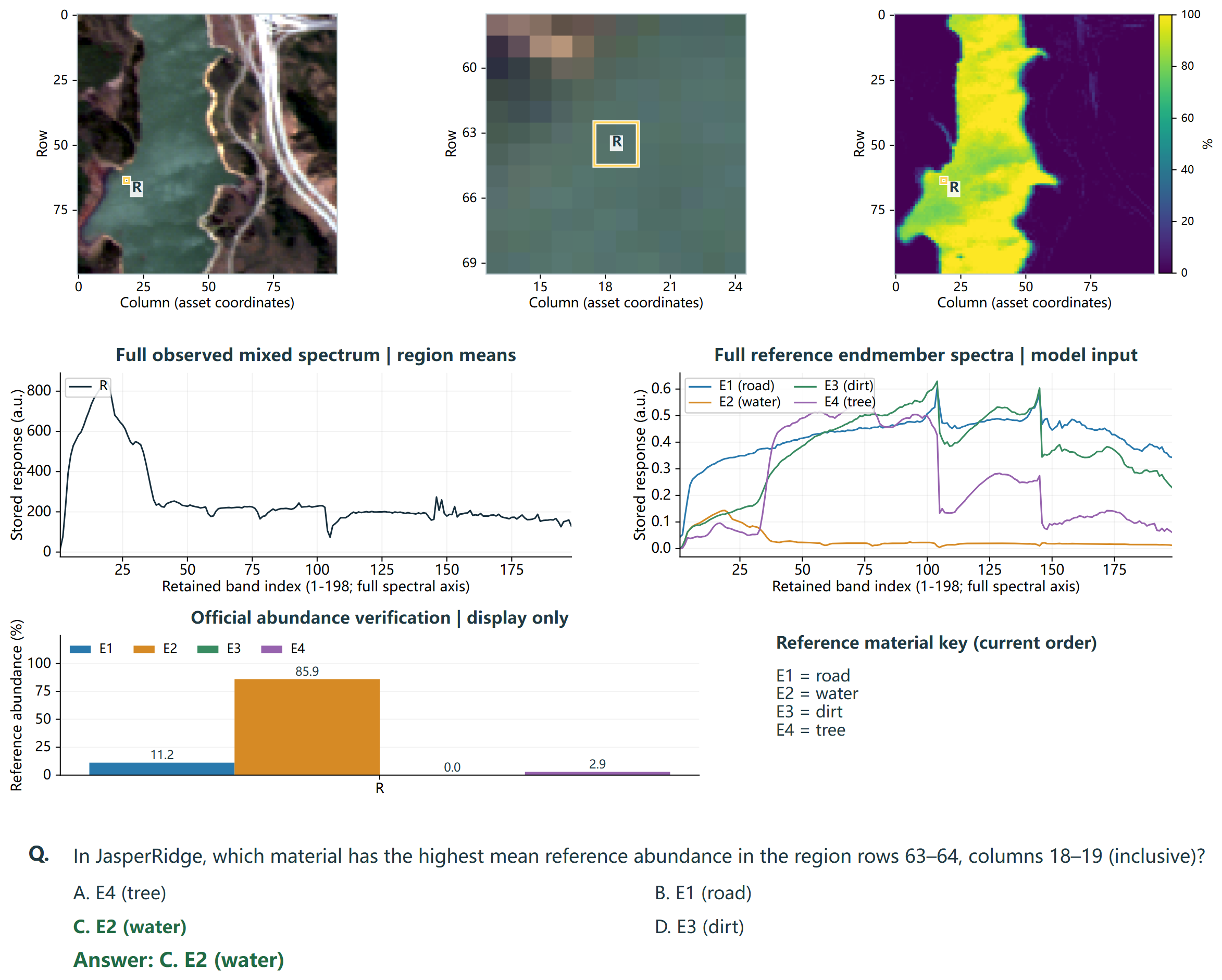}{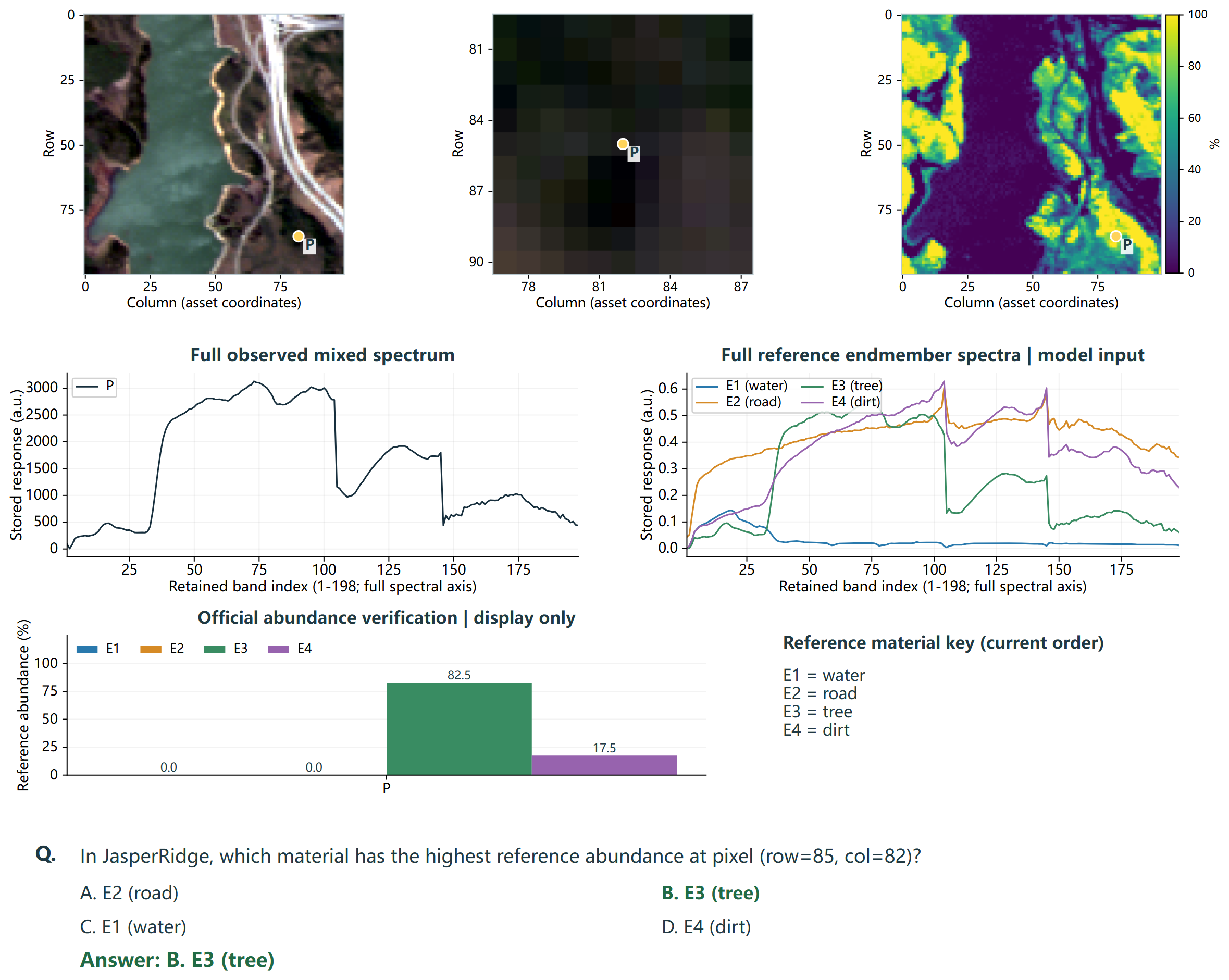}
\clearpage
\HyperGPTQualitativePair{Dominance-based retrieval}{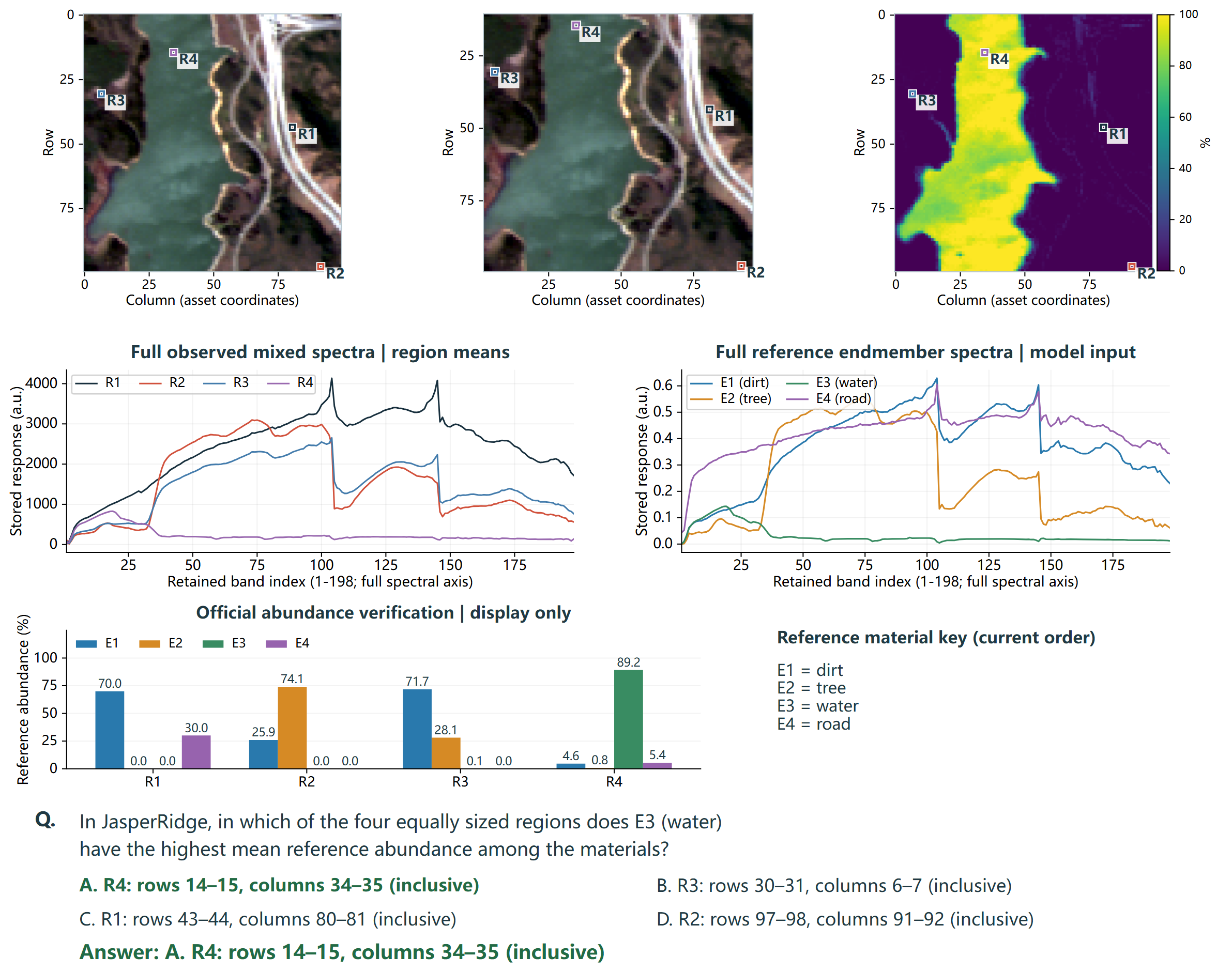}{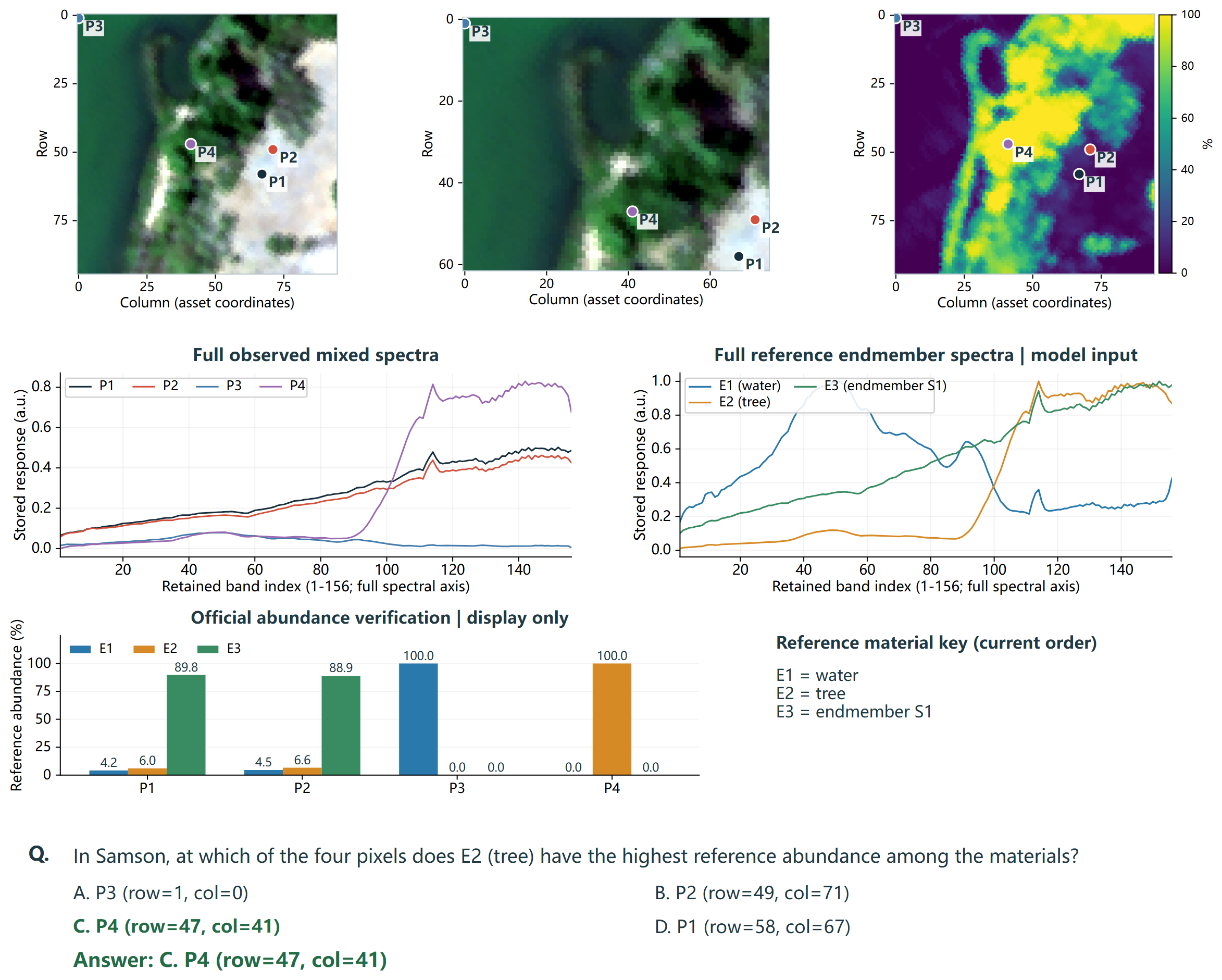}
\clearpage
\endgroup